\documentclass{article} %
\usepackage{iclr2027_conference,times}
\iclrfinalcopy

\usepackage{amsmath,amsfonts,bm}

\def\eqref#1{equation~\ref{#1}}

\def\1{\bm{1}}

\DeclareMathAlphabet{\mathsfit}{\encodingdefault}{\sfdefault}{m}{sl}
\SetMathAlphabet{\mathsfit}{bold}{\encodingdefault}{\sfdefault}{bx}{n}

\usepackage{hyperref}
\usepackage{url}

\usepackage{caption}
\usepackage{subcaption}
\usepackage{graphicx}

\graphicspath{{./images/}}

\usepackage{wrapfig}
\usepackage{longtable}
\usepackage{booktabs}
\usepackage{multirow}
\usepackage{tcolorbox}
\usepackage{listings}
\usepackage[utf8]{inputenc}
\usepackage[T1]{fontenc}
\usepackage[table]{xcolor}

\title{Think Multilingual, Not Harder: A Framework for Analyzing and Teaching Code-Switched Reasoning}

\author{Eleanor M. Lin \& David Jurgens \\%\thanks{ Use footnote for providing further information
Electrical Engineering and Computer Science Department\\
University of Michigan\\
Ann Arbor, MI 48105, USA \\
\texttt{\{elealin,jurgens\}@umich.edu} \\
}

\usepackage{xspace}

\newcommand{\corestyle}[1]{\textsc{#1}}

\newcommand{\core}{\texorpdfstring{\corestyle{CoRe}\xspace}{CoRe}}
\newcommand{\encore}{\texorpdfstring{\corestyle{EnCoRe}\xspace}{EnCoRe}}

\definecolor{resHigh}{HTML}{C6DBEF}  %
\definecolor{resLow}{HTML}{FDD9B5}   %
\definecolor{multiA}{HTML}{E5F5E0}   %
\definecolor{multiB}{HTML}{A1D99B}   %
\definecolor{multiC}{HTML}{5DBB6A}   %
\newcommand{\swatch}[1]{\fcolorbox{black!30}{#1}{\rule{0pt}{4pt}\rule{4pt}{0pt}}}

\begin{document}

\maketitle

\begin{abstract}
Recent developments in reasoning capabilities have enabled large language models (LLMs) to solve increasingly complex tasks. Interestingly, while reasoning models are often trained to generate monolingual text, LLMs have also been observed to code-switch (i.e., mix languages). Prior works have either viewed code-switching as an undesirable error, attempted to control code-switching through modifications to input prompts or the output decoding process, or focus on narrow subsets of languages, domains, tasks, and models. We address these gaps by introducing the first linguistically and behaviorally motivated fine-tuning framework for identifying beneficial code-switched reasoning behaviors in LLMs and teaching these models to code-switch more effectively for reasoning. First, we create the \textbf{\underline{Co}}de-Switched \textbf{\underline{Re}}asoning (\core) corpus, consisting of 14k reasoning traces from 17 models, 23 languages, 12 scripts, and diverse reasoning domains, providing insights into potentially helpful code-switching behaviors. Then, we expand \core into the \textbf{\underline{E}}xte\textbf{\underline{n}}ded \textbf{\underline{Co}}de-Switched \textbf{\underline{Re}}asoning corpus (\encore) with 40 carefully curated datasets for training and evaluating six interventions on code-switching in reasoning across three models and seven languages, totaling 120 fine-tuning conditions. Across 80k+ reasoning traces from both language/culture-agnostic and -specific evaluations, English-dominated reasoning, semantically accurate code-switching, and a more even mix of languages are positively associated with correct answers, whereas dense switching is associated with incorrect answers. Moreover, we are able to elicit positive behaviors through fine-tuning tasks that do not directly demonstrate code-switching. Our work suggests that small but well-curated datasets can change how reasoning models code-switch, allowing us to reap the benefits of reasoning in the many languages that lack large-scale reasoning data. 
\end{abstract}

\section{Introduction}\label{sec:intro}

When human multilinguals mix multiple languages during communication, they are said to \emph{code-switch} \citep{doi:https://doi.org/10.1002/9781405166256.ch13}. Recently, works including \citet{Guo_2025} have observed the parallel phenomenon of code-switching in \textit{reasoning}---the long chain-of-thought texts produced by large language models (LLMs) when solving complex problems. While mechanisms of human code-switching have been studied extensively (e.g., see \citet{doi:https://doi.org/10.1002/9781405166256.ch13} and \citet{cambridge-csw}), in comparison, the emergent behavior of code-switching in LLM reasoning remains relatively poorly understood. Some works \citep[e.g.,][]{Guo_2025,marchisio-etal-2024-understanding} treat code-switching as an error, while others attempt to use code-switching strategically for reasoning tasks in particular \citep{li-etal-2025-impact, li2025impactlanguagemixingbilingual}. In this work, we are particularly interested in code-switched LLM reasoning as a route to improve reasoning capabilities for lower-resource languages. This issue is of critical importance given that currently available reasoning models are mostly limited to high-resource languages like English \citep{ghosh-etal-2025-survey}.

To understand how state-of-the-art reasoning models code-switch and design interventions that improve code-switching quality, we seek answers to the following questions:

\begin{enumerate}
    \item [\textbf{RQ 1}] How do LLMs code-switch during reasoning?
    \item [\textbf{RQ 2}] What types of code-switching are associated with better reasoning task performance?
    \item [\textbf{RQ 3}] What types of fine-tuning interventions can teach reasoning models beneficial forms of code-switching for reasoning?
\end{enumerate}

To address these questions, we first create and analyze a corpus of reasoning examples from diverse models, languages, tasks, and domains (section \ref{sec:taxonomy}). Based on the findings from this initial corpus, we design supervised fine-tuning approaches (including training data and evaluation on parallel and non-parallel language/culture-specific and -agnostic benchmarks) to develop beneficial code-switched reasoning behaviors (section \ref{sec:sft}).

Our contributions are as follows. First, we introduce the \textbf{\underline{Co}}de-Switched \textbf{\underline{Re}}asoning corpus (\core) of 14k reasoning traces from diverse models, languages, tasks, and domains. With this initial corpus, we develop and apply a taxonomy of code-switching behaviors to identify characteristics that benefit reasoning performance. Second, we introduce a linguistically and behaviorally motivated fine-tuning framework for teaching reasoning models to code-switch for better reasoning performance. This framework expands \core into the \textbf{\underline{E}}xte\textbf{\underline{n}}ded \textbf{\underline{Co}}de-Switched \textbf{\underline{Re}}asoning corpus (\encore) with 1M tokens of training data per task per language, across 6 tasks and 7 languages. Third, we systematically analyze code-switching behaviors produced by our framework and their relationship with reasoning performance. We find that fine-tuning for translation tasks most consistently shifts code-switching toward behaviors associated with correct answers. Crucially, this benefit appears to be most pronounced for low resource languages. Overall, we hope that this work will serve as the foundation for further future work on code-switching as a means to improving multilingual LLM reasoning in low-resource settings. We will release all reasoning traces, SFT data, trained adapters, and evaluation scripts needed to reproduce the main figures under open licenses.

\begin{figure}[!htbp]
    \centering
    \includegraphics[clip, trim={0cm 0cm 0cm 0cm}, width=\textwidth]{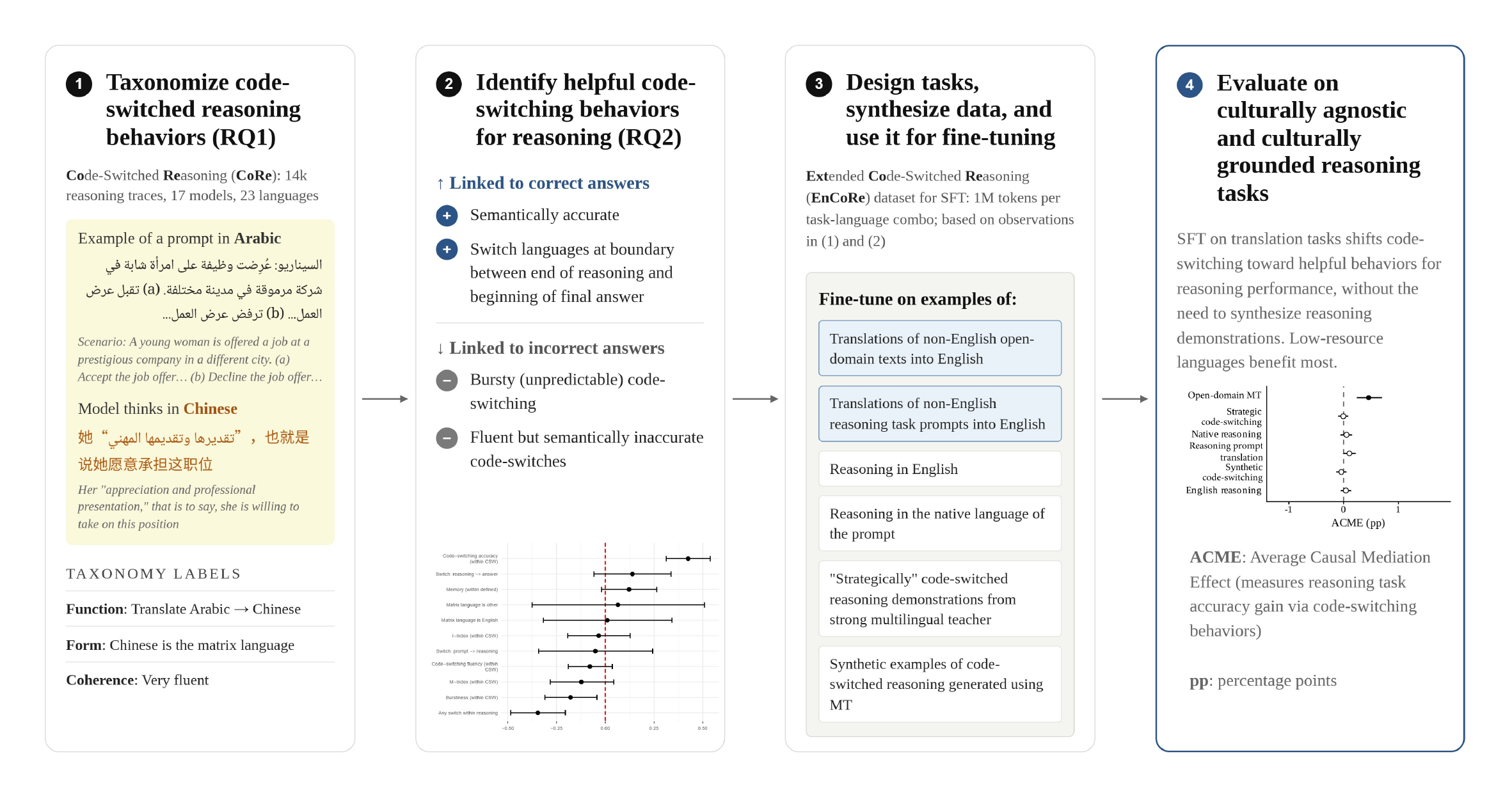} %
    \caption{Our code-switched reasoning framework. (There are 120 instead of 126 conditions shown in (3) due to practical limitations on covering every possible task $\times$ model $\times$ language combination.)}
    \label{fig:ex-csw}
\end{figure}

\section{Related Work}
\label{sec:related}

\textbf{We incorporate prior insights from research on human code-switching to contribute to a growing body of work on controlling the languages that models use during reasoning} \citep[e.g.,][]{tam2025languagemattersmultilingualinput,gao2025thinkingmultilinguallyempowerllm}. Broadly defined, code-switching occurs when a multilingual person mixes multiple languages during communication \citep{doi:https://doi.org/10.1002/9781405166256.ch13}. Human code-switching is driven by a complex array of social and cognitive factors, e.g., ethnolinguistic identity and language proficiency \citep{doi:https://doi.org/10.1002/9781405166256.ch13, doi:10.1177/003368827700800201, cambridge-csw}. While language models---and particularly reasoning models---have likewise been observed to code-switch, the mechanisms behind this behavior remain poorly understood. Some works treat code-switching by language models as an undesirable behavior \citep{marchisio-etal-2024-understanding, Guo_2025}. However, our work more closely aligns with \citet{li-etal-2025-impact}, who attempt to harness code-switching as a useful reasoning behavior. The main difference between this work and \citet{li-etal-2025-impact, li2025impactlanguagemixingbilingual} is that while they target the decoding stage, we modify model behavior through supervised fine-tuning. Additionally, while \citet{winata2026largelanguagemodelsunderstand} observe monolingual English reasoning conditioned on synthetic code-switched prompts, we study spontaneous LLM code-switched reasoning conditioned on monolingual prompts.

\textbf{Prior work on code-switched reasoning lacks our coverage of diverse models, languages, domains, tasks, and reasoning behaviors.} 
\citet{yong2025crosslingualreasoningtesttimescaling} study the English-centric s1 model family using commonsense, factual, cultural, and causal reasoning benchmarks, with prompts in four languages other than English. \citet{li2025impactlanguagemixingbilingual, li-etal-2025-impact} study Chinese/English math reasoning from the Chinese-English QwQ32B-Preview model. \citet{wang2025languagemixingreasoninglanguage, wang-etal-2025-language-mixing} study the Chinese-English DeepSeek-R1 model family and multilingual QwQ-32B, Qwen3, and Gemini 2.0 Flash Thinking models on commonsense, factual, and logical reasoning in 15 languages. \citet{winata2026largelanguagemodelsunderstand} study 9 models from the Gemma, Llama, Qwen, Olmo, and GPT-OSS families; they derive their 13-language dataset from an existing factuality benchmark \citep{haas2025simpleqaverifiedreliablefactuality}. In contrast, in \core and \encore we cover 19 models from 11 families with diverse multilingual capabilities, 26 languages, and reasoning domains/tasks beyond those in prior work, e.g., moral reasoning (taking into consideration the experiments and analyses from both section \ref{sec:taxonomy} and section \ref{sec:sft}). When characterizing code-switching behaviors, \citeauthor{yong2025crosslingualreasoningtesttimescaling} identify the ``quote-and-think'' pattern and apply the pre-existing linguistic concepts of a matrix language and intra-/inter-sentential switching. \citeauthor{wang2025languagemixingreasoninglanguage} quantify relative proportions of reasoning languages. \citeauthor{li2025impactlanguagemixingbilingual} identify four main patterns. In contrast, we introduce a structured taxonomy of code-switching patterns across three dimensions and quantify these patterns through 8 code-switching metrics validated by linguistics and NLP literature.

\textbf{Our work combines top-down, theory-driven and bottom-up, data-driven approaches for classifying reasoning behavior.} \citet{gandhi2025cognitivebehaviorsenableselfimproving} identify reasoning behaviors that both drive performance and parallel human behaviors, while \citet{lee2025cotencyclopediaanalyzingpredicting} characterize reasoning through bottom-up LLM-assisted brainstorming and clustering. We build on Lee et al. but focus on code-switching, add manual curation to ground the taxonomy in linguistic theories of human code-switching, and include smaller multilingual models and domains beyond math, science, and general knowledge.

\textbf{Prior work controls reasoning language at test time; we modify behavior with supervised fine-tuning (SFT).} \citet{yong2025crosslingualreasoningtesttimescaling} and \citet{zhao-etal-2026-comprehensive} steer reasoning language with prompt instructions and by prefilling target-language tokens; \citet{li-etal-2025-impact} train a probe to guide language choice during decoding; and \citet{wang-etal-2025-language-mixing} mask the logits of non-target scripts. We treat these prior inference-time approaches as potentially complementary to our SFT approach, rather than as baselines. Our SFT approach is partially inspired by \citet{muennighoff-etal-2025-s1, muennighoff2025s}, who show that reasoning capabilities can be developed with SFT on a small, carefully curated corpus.

\section{Understanding Code-Switched Reasoning Behaviors} %
\label{sec:taxonomy}
To answer \textbf{RQ 1} on how LLMs code-switch, we create the \textbf{\underline{Co}}de-Switched \textbf{\underline{Re}}asoning corpus (\core) of 14k reasoning traces from 17 models.
\subsection{Approach}\label{sec:taxonomy-approach}
Theory-grounded approaches which look for pre-defined, human-aligned behaviors in model reasoning (e.g., \citeauthor{gandhi2025cognitivebehaviorsenableselfimproving}) may miss novel reasoning behaviors that do not align with humans. Data-driven approaches which derive categories of behaviors from observations of model reasoning (e.g., \citet{lee2025cotencyclopediaanalyzingpredicting}) may fail to surface misalignment with human behaviors that only appear in humans. To address these limitations, we fuse theory-driven and data-driven approaches in our framework.

\textbf{Data and models.} To create \core, we generate reasoning examples from diverse models, languages, tasks, and domains for generalizability (see appendices \ref{sec:taxonomy-models} and \ref{sec:taxonomy-data} for full lists). We generate reasoning examples from 17 models using prompts from 10 datasets in 23 languages (Arabic, Assamese, Azerbaijani, Bengali, Burmese, English, French, German, Greek, Hausa, Hindi, Indonesian, Italian, Japanese, Korean, Mandarin Chinese, Persian, Portuguese, Russian, Spanish, Swahili, Thai, and Yoruba), 12 scripts, and 9 language families \citep{glottolog, unicode}. We include approximately 50 instances from each model/language/dataset combination, yielding 7k instances total. (See appendix \ref{sec:taxonomy-config} for prompting configurations used to generate reasoning examples.)

\textbf{Code-switched reasoning behavior taxonomy.} %
We apply the framework and human evaluation procedure introduced by \citet{lee2025cotencyclopediaanalyzingpredicting} for analyzing LLM reasoning. We prompt Gemini 2.5 Flash to analyze \core for patterns that differentiate code-switching strategies \citep{gemini, geminiteam2025geminifamilyhighlycapable}. 
Then, we apply topic modeling to Gemini's outputs, using BERTopic \citep{grootendorst2022bertopicneuraltopicmodeling}. Finally, we manually group topics into dimensions, categories, and subcategories, guided by relevant literature on code-switching and reasoning in humans and LLMs (see appendices \ref{sec:brainstorm} and \ref{sec:bertopic}). We randomly sample 100 instances for human evaluation (see appendix \ref{sec:human}). We find that Gemini is an effective annotator, generating plausible analyses $85\%$ of the time.

We ground our taxonomy (appendix \ref{sec:full_taxonomy}) in theories from prior literature and our data analysis:

\textit{\underline{1. Function}} describes the purpose served by a code-switch. Common functions include \textit{translating} from one language into another and \textit{quoting} material from the initial user prompt in its original language when the reasoning context is in a different language. Humans also code-switch to quote material in its original language, provide context in another language, and repeat phrases for emphasis and clarity \citep{begum-etal-2016-functions, belani-flanigan-2023-automatic}. %

\textit{\underline{2. Form}} describes the structure of code-switching, e.g., whether a model switches languages for a single word, phrase, sentence, or even the entire reasoning trace. Code-switching may occur with varying frequency/density within the reasoning. Additionally, a particular language may take on the role of the \textit{matrix} (dominant) language \citep{bullock-etal-2018-predicting}.

\textit{\underline{3. Coherence}} describes how \textit{fluent} (natural and understandable) and \textit{accurate} (using semantically appropriate terms) the code-switching is. \citet{kuwanto2024linguisticstheorymeetsllm} also find \textit{fluency} and \textit{accuracy} to be relevant concepts for their human evaluation of automatically generated code-switched text.

\textbf{Code-switching metrics.} 
From the datasets in appendix \ref{sec:taxonomy-data} (excluding instances used to develop the taxonomy), we include about 50 instances from each model/dataset/language combination except English (which does not usually elicit code-switching), yielding another 7k instances in total. We use Qwen3.5-27B to identify code-switched instances, then compute code-switching metrics (see appendix \ref{sec:instance-lid}). We select metrics from the code-switching research community \citep{srivastava-singh-2021-challenges}. Many metrics rely on word-level language identification (LID; see appendix \ref{sec:word-lid}). We validate word- and instance-level LIDs on 140 held-out instances, obtaining 0.933 word-level and 0.942 instance-level accuracy.

We quantify the \textbf{Form} dimension of our taxonomy with six widely used token-based metrics:

\textit{\underline{1. The Code-mixing Index (CMI)}} measures code-switching using the proportion of tokens from the dominant language in the text. A higher CMI indicates more code-switching \citep{das-gamback-2014-identifying, srivastava-singh-2021-challenges}. Given $n$ tokens (including $u$ language-independent tokens) in $N$ languages, if $w_i$ is the number of tokens from the $i^{th}$ language, then 
$\text{CMI} = \frac{\sum_{i=1}^N(w_i) - max\{w_i\}}{n-u}$.

\textit{\underline{2. The Multilingual Index (M-Index)}} measures code-switching by comparing the relative proportions of tokens from each language \citep{automatic, srivastava-singh-2021-challenges}. A higher M-Index indicates more code-switching. Given $n$ tokens in $N$ languages where $w_i$ is the number of tokens from the $i^{th}$ language, $\text{M-Index} = \frac{1 - \sum_{i=1}^{N}(\frac{w_i}{n})^2}{(N-1)(\sum_{i=1}^{N}(\frac{w_i}{n})^2)}$.

\textit{\underline{3. Memory}} correlates consecutive language span lengths \citep{Goh_2008, guzman17_interspeech, srivastava-singh-2021-challenges}. Given a text of $n_r$ language spans (consecutive tokens in the same language), where $\tau_i$ denotes length of the $i^{th}$ language span, $\sigma_1, \mu_1$ denote the standard deviation and mean of language span lengths (except the last), and $\sigma_2, \mu_2$ denote the standard deviation and mean of language spans (except the first), $\text{Memory} = \frac{1}{n_r-1}\sum_{1}^{n_r-1}\frac{(\tau_i - \mu_1)(\tau_{i+1}-\mu_2)}{\sigma_1\sigma_2}$.

\textit{\underline{4. Burstiness}} measures how predictable code-switches are \citep{srivastava-singh-2021-challenges, Goh_2008}. Given the mean and SD of language span lengths ($\mu, \sigma$), $\text{Burstiness} = \frac{\sigma - \mu}{\sigma + \mu}$.

\textit{\underline{5. The Integration-Index (I-Index)}} measures the probability that the language of the next token will differ from the previous token \citep{guzman17_interspeech, srivastava-singh-2021-challenges}. Given $n$ tokens and $S(i,i+1)=1$ if the $i^{th}$ token's language differs from the $i+1^{th}$ token, $\text{I-Index} = \frac{\sum_{i=1}^{n-1}S(i,i+1)}{n-1}$.

\textit{\underline{6. The matrix language}} provides the syntax of a code-switched sentence \citep{myers2001matrix}. Adopting the heuristic from \citet{yong2025crosslingualreasoningtesttimescaling}, we treat the language with the most tokens as the matrix language. The matrix language is (1) same as the prompt, (2) English, or (3) other.

To measure \textit{\underline{code-switching fluency}} and \textit{\underline{code-switching accuracy}} from the \textbf{Coherence} dimension of our taxonomy, we adapt the approach from \citet{kuwanto2024linguisticstheorymeetsllm} for evaluating naturalistic code-switched text generation. We use Qwen/Qwen3.5-27B as our rater (see appendices \ref{sec:acc-eval} and \ref{sec:fl-eval}) because it is among the strongest open models supporting the low-resource languages in our study. We also validate our fluency rating configuration by comparing our ratings on the SwitchLingua dataset of naturalistic, human-evaluated code-switching to the SwitchLingua groundtruth scores \citep{NEURIPS2025_4847447c}. We obtain a moderately strong Spearman's correlation of $r = 0.45\ (p < 0.001)$ between our scores and the SwitchLingua fluency scores, confirming that our judge captures significant signals of fluency (see appendix \ref{sec:switch-lingua} for details).

 \textbf{Statistical modeling (Figure \ref{fig:taxonomy-behaviors}, right).} To estimate how code-switching behaviors relate to reasoning performance, we fit a logistic mixed model. Our mixed model prevents the estimated effects of code-switching from being trivially explained by differences in overall LLM performance, language resource level, or question difficulty. (See appendix \ref{appendix:regression} for details.)

\subsection{Results}
\label{sec:taxonomy-results}
Our framework yields many examples of spontaneously code-switched model reasoning in response to monolingual prompts, which contrasts strongly with \citet{winata2026largelanguagemodelsunderstand}. (They find that ``reasoning traces are consistently in English, even when the inputs are predominantly in non-English languages.'') Next, we use the \textbf{Form}, \textbf{Function}, and \textbf{Coherence} dimensions of our taxonomy to answer \textbf{RQ 1} on how LLMs code-switch and \textbf{RQ 2} on what types of code-switching are associated with better reasoning task performance.

\begin{figure}[t]
    \centering
    \begin{subfigure}[b]{0.5\linewidth}
        \centering
        \includegraphics[clip, trim={0cm 0cm 0cm 0cm}, width=\linewidth]{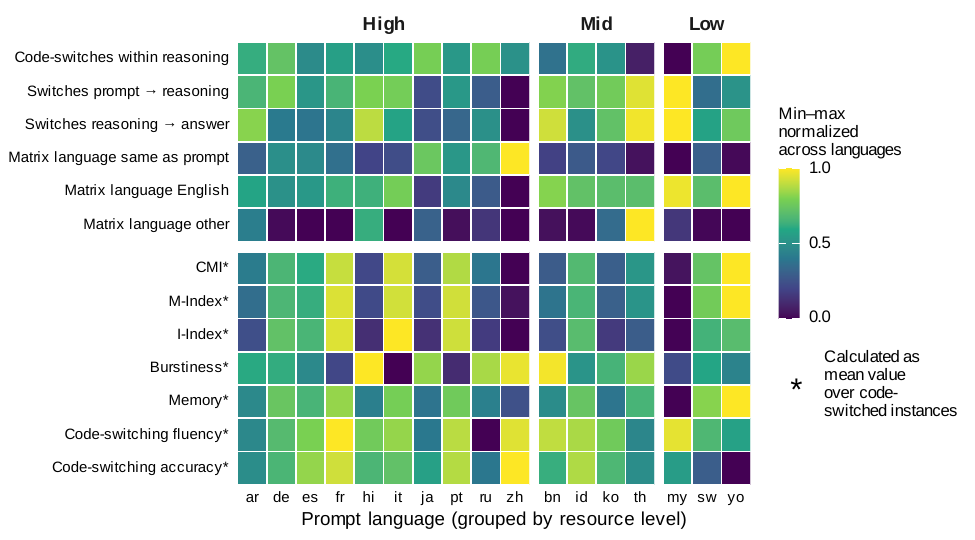}
    \end{subfigure}%
    \begin{subfigure}[b]{0.5\linewidth}
        \centering
        \includegraphics[clip, trim={0cm 0cm 0cm 0cm}, width=\linewidth]{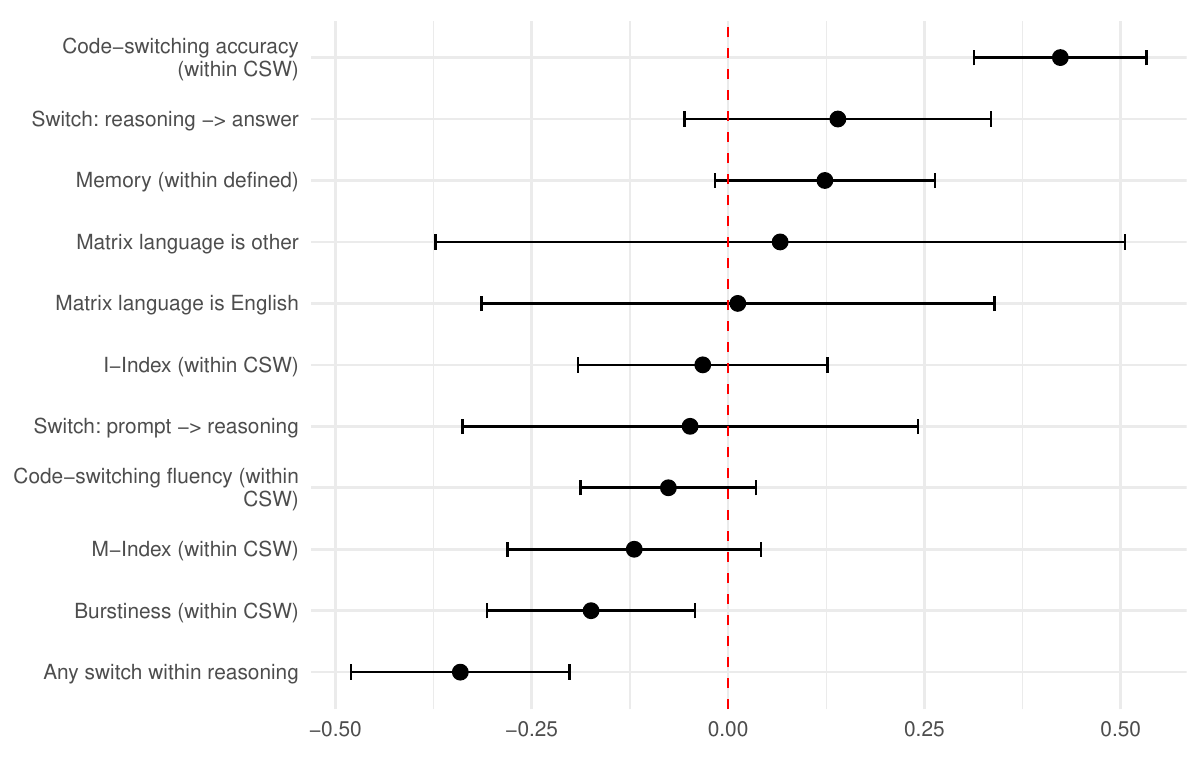}
    \end{subfigure}
    \caption{\textbf{Left:} Despite being written in unrelated languages and scripts, prompts in Arabic, Hindi, and Thai often elicit reasoning dominated by Chinese content. Other languages tend to elicit English-dominated reasoning. As illustrated by Burmese (my), code-switching accuracy and code-switching fluency can diverge substantially for the same language. \textbf{Right:} Code-switching accuracy (but not code-switching fluency) is positively associated with reasoning task performance. (Data is the 7k subset of \core described in section \ref{sec:taxonomy-approach}. ``Within CSW'' refers to metrics that are only defined when there is code-switching within the reasoning part of a model's response. Bars are 95\% CIs.)}
    \label{fig:taxonomy-behaviors}
\end{figure}

\textbf{Analyzing the \textit{form} of code-switched reasoning in response to different prompt languages reveals that models' choices of code-switching languages do not necessarily align with intuitions based on human code-switching or prior work.} In particular, we might expect models to mix closely related, typologically similar, or geographically overlapping languages. Instead, we find that Arabic, Hindi, and Thai prompts (despite being written in unrelated languages and scripts) tend to mix the prompt language with Chinese as the matrix language (Figure \ref{fig:taxonomy-behaviors}, left). Most other prompt languages elicit English as the reasoning matrix language across language resource levels, language families, and scripts. Thus, while we concur with prior works finding that Chinese and English are the most common matrix languages found in code-switched reasoning, we differ from prior findings that models are more likely to code-switch between similar languages \citep{tam2025languagemattersmultilingualinput, Tam2025LanguageMH}.

\textbf{Our analysis of code-switching's \textit{function} in reasoning agrees with \citet{yong2025crosslingualreasoningtesttimescaling} that code-switching commonly serves a ``quote-and-think'' function.} In particular, code-switching often translates prompt content into a higher-resource language or quotes prompt material in its original language (see Figure \ref{fig:ex-csw} for an example in Chinese and Arabic). This model behavior parallels compensatory code-switching in humans, in which multilinguals switch to a more proficient language to compensate for weaker proficiency in another language \citep{language_proficiency_2023}.

\textbf{In contrast with prior work, we find that conditioning answers on high density code-switching is not significantly associated with reasoning task performance (Figure \ref{fig:taxonomy-behaviors}, right).} When evaluating model responses to code-switched prompts, \citet{winata2026largelanguagemodelsunderstand} find that higher code-switching density (measured with Switch-Point Fraction, which is mathematically identical to the I-Index used here) correlates with better task performance. However, we find instead that conditioning answers on higher density code-switching (as measured by the I-Index) in the reasoning generated at test time has no significant effect on reasoning task accuracy.

\textbf{Bursty code-switching is associated with worse reasoning task performance (Figure \ref{fig:taxonomy-behaviors}, right).} Since burstiness is a measure of how unpredictable code-switches are, this result suggests that random code-switching is unhelpful for reasoning.

\textbf{Contrary to intuition, code-switching accuracy and code-switching fluency (taxonomy \textit{coherence} dimension) can diverge substantially.} 
For example (as seen in Figure \ref{fig:taxonomy-behaviors}, left), reasoning in response to Burmese (my) prompts is rated as highly fluent but less accurate under our taxonomy, corresponding to code-switching that is natural and understandable but uses semantically inappropriate terms. This suggests that models can produce smoothly integrated code-switched text that is nonetheless semantically incorrect.

\textbf{{Code-switching accuracy (but not code-switching fluency) is positively associated with reasoning task performance.}} In particular, higher code-switching accuracy is associated with better performance (Figure \ref{fig:taxonomy-behaviors}, right). However, increased fluency of code-switching has no significant association with reasoning task performance. Taken together, these results suggest that the \emph{semantic correctness of} code-switched reasoning is more predictive of performance than its \emph{syntactic integration}.

\section{Producing Code-Switching Behaviors Using Linguistically and Behaviorally Motivated Data Curation}\label{sec:sft}

Based on our observations of code-switched reasoning behaviors, we answer \textbf{RQ 3} by designing and testing the effect of six fine-tuning interventions on code-switching. By eliciting behavioral changes through these controlled interventions, we move beyond the observational approach in section \ref{sec:taxonomy} to induce code-switched reasoning behaviors predictive of better reasoning performance.

\subsection{Approach}\label{sec:approach2}
\textbf{Datasets.}\label{sec:datasets}
We use the Global MMLU, Global PIQA, and No Language Left Behind to build \encore, our reasoning SFT dataset \citep{singh2025globalmmluunderstandingaddressing, chang2026globalpiqaevaluatingcommonsense, nllbteam2022languageleftbehindscaling}.

\emph{\underline{Global MMLU}} extends the Massive Multitask Language Understanding dataset to 42 languages, using human translations and human-validated machine translations \citep{singh2025globalmmluunderstandingaddressing, hendrycks2021measuring}. We select Global MMLU as a source of high-quality reasoning problems from geographically and typologically diverse languages. We train and evaluate on Amharic, Hindi, Igbo, Indonesian, Malay, Swahili, and Yoruba, which represent three scripts, three resource levels, four language families, and over 1 billion speakers \citep{ethnologue}. By selecting for language diversity, we can observe our interventions' interaction with language-specific characteristics. Next, we filter Global MMLU for questions from reasoning-related domains and all questions containing ``='' based on heuristics from \citet{sprague2025to}. (See appendix \ref{sec:reasoning-subs} for our full hand-curated list of reasoning-related subjects.) Finally, we set aside validation and test sets of 500 examples each. 

\textit{\underline{Global PIQA}} evaluates commonsense reasoning in $100+$ languages using questions written by native speakers \citep{chang2026globalpiqaevaluatingcommonsense}. In addition to evaluating on a curated subset of examples from Global MMLU, we evaluate our fine-tuned models on the non-parallel split of Global PIQA, which evaluates culturally specific commonsense reasoning. Global PIQA tests how well our interventions generalize beyond translated benchmarks like Global MMLU to non--English-centric settings. We evaluate in the same seven languages as our Global MMLU subset, using the full 100-instance Global PIQA non-parallel test set for each language.

\emph{\underline{No Language Left Behind (NLLB)}} provides open-source parallel translations between over 200 languages \citep{nllbteam2022languageleftbehindscaling}. We use NLLB for our open-domain machine translation task.%

\textbf{Models.}\label{sec:models}
We apply our fine-tuning interventions to three reasoning models representing different parameter sizes, families, and multilingual capabilities. \emph{Qwen3-8B} supports 119 language varieties, \emph{Phi-4-Reasoning} (14B parameters) only fully supports English, and \emph{DeepSeek-R1-Distill-Llama-8B} has strong capabilities in English and Chinese \citep{Guo_2025}.

\textbf{Data synthesis.} To generate fine-tuning data for those experimental conditions using reasoning demonstrations, we prompt Qwen3-Next-80B-A3B-Thinking using the default recommended sampling parameters \citep{yang2025qwen3technicalreport}. We select Qwen3-Next-80B-A3B-Thinking as the teacher model for its strong multilingual reasoning capabilities. To test the generalizability of our approach to other teachers, we also fine-tune Qwen3-8B in Swahili on demonstrations from gemma-4-26B-A4B-it. (We leave a full-scale teacher generalization test to future work.) For interventions using machine translation (MT) as part of the data synthesis pipeline, we use the SeamlessM4T \citep{communication2023seamlessmultilingualexpressivestreaming}, NLLB-200 \citep{nllbteam2022languageleftbehindscaling}, and 70B Apertus models \citep{apertus2025apertusdemocratizingopencompliant}. We provide details of our MT and data synthesis pipelines in appendices \ref{sec:mt} and \ref{sec:synthesis}.

\textbf{Supervised fine-tuning tasks.}\label{sec:conditions}
The general format for all supervised fine-tuning tasks is $(p, r, a)$: conditioned on prompt $p$, the reasoning model is fine-tuned to generate reasoning $r$ leading to final answer $a$. To ensure a fair comparison across tasks and languages, we set a token budget of 1 million tokens for fine-tuning each model in each language for a given task. Because the same content tokenizes into different numbers of tokens in different languages \citep{ahia-etal-2023-languages}, a fixed token budget yields different numbers of training examples per language. We therefore control for the number of training examples in our downstream analyses. Due to constraints on the amount of high-quality data we are able to generate for Yoruba, we do not release datasets for the native-language reasoning or synthetically code-switched reasoning tasks in this language. Full details of our fine-tuning configuration are in appendix \ref{sec:sft-config}.

Our six SFT tasks are designed to allow direct comparisons between general strategies employed by related frameworks and strategies inspired by section \ref{sec:taxonomy} ($l$, $e$ are the target language and English): 

\textit{\underline{1. Native-language reasoning}} ($p_l, r_l, a_l$) is a monolingual fine-tuning baseline task which we include for comparison with code-switching tasks. We include this baseline to represent the common approach of existing methods that seek to improve language-specific reasoning performance simply by training on demonstrations of reasoning in the target language \citep{chen-etal-2024-breaking}. To synthesize native-language reasoning demonstrations, we generate English reasoning demonstrations from the teacher model and filter for appropriate length and correctness, then translate the reasoning into $l$.

\textit{\underline{2. Open-domain machine translation into English}} $(p_l, \emptyset, a_e)$ trains from sentence-level NLLB parallel translations, with empty reasoning. This machine translation condition is motivated by our observations in section \ref{sec:taxonomy-results} (under the \textbf{Function} taxonomy dimension) that code-switching within reasoning often serves to translate prompt content into a higher-resource language. Thus, we expect improvements in models' general translation abilities to also improve their code-switching during reasoning. Importantly for resource-constrained settings, training on this machine translation task does not require synthesis of reasoning demonstrations.

\textit{\underline{3. Reasoning prompt translation into English}} $(p_l, \emptyset, a_e)$ trains the model to translate Global MMLU's parallel prompts into English. Like the open-domain machine translation task above, the reasoning prompt translation task is motivated by our observations that code-switching often helps models translate and reason about prompt content (section \ref{sec:taxonomy}; taxonomy \textbf{Function} dimension). However (in contrast with the open-domain machine translation task) here the model learns from in-domain parallel translations of reasoning task prompts.

\textit{\underline{4. English-language reasoning}} $(p_l, r_e, a_l)$ demonstrations switch from the prompt language into English for the entire reasoning trace, before switching back to the prompt language to give the final answer. The English-language reasoning task is motivated by analysis of code-switching \textbf{Form} in section \ref{sec:taxonomy}, which suggests that the location and granularity of code-switches (e.g., within v. across boundaries between the prompt, reasoning, and answer) are important characteristics of code-switching behavior. Additionally, our observations of code-switching \textbf{Form} suggest that using English as the main language of reasoning is common across languages other than English. Fine-tuning on English-language reasoning demonstrations allows us to go a step further and observe the causal effects of such data on code-switching behaviors and reasoning task performance.

\textit{\underline{5. Strategically code-switched reasoning}} $(p_l, r_{csw}, a_l)$ demonstrations are synthesized from a strong multilingual teacher model by prompting in the target language with prompt $p_l$, then filtering for reasoning containing multiple languages. Training on these teacher model reasoning demonstrations tests the view that code-switching serves a strategic \textbf{Function} and that a strong multilingual reasoner like our Qwen3-Next-80B-A3B-Thinking teacher model can generate examples of ``good'' code-switches for reasoning \citep{li-etal-2025-impact}. Additionally, we expect the code-switching demonstrations synthesized from a strong multilingual teacher model to be more \textbf{Coherent} than code-switching demonstrations synthesized using post-hoc machine translation.

\textit{\underline{6. Synthetically code-switched reasoning}} $(p_l, r_{csw}, a_l)$ demonstrations are created by first prompting our Qwen3-Next-80B-A3B-Thinking teacher model in English to obtain English-language reasoning demonstration $r_e$. Then we apply machine translation to $r_e$ to obtain reasoning demonstrations $r_l$ in the target language. Finally, we splice together $r_e$ and $r_l$, preserving the original order of reasoning steps from $r_e$ but swapping out some of these randomly selected English-language steps with their translations from $r_l$. We include this task to test the effect of the random (and hence less \textbf{Coherent}) code-switches, in contrast with the strategic code-switching task. Additionally, we expect the synthetically code-switched reasoning demonstrations to be more bursty than strategically code-switched demonstrations, so our results in section \ref{sec:taxonomy-results} suggest that training on the synthetic demonstrations will not be helpful, compared to the strategic ones.

\begin{wrapfigure}[23]{l}{0.5\textwidth}
    \centering
    \includegraphics[clip, trim={0cm 0cm 0cm 0cm}, width=0.48\textwidth]{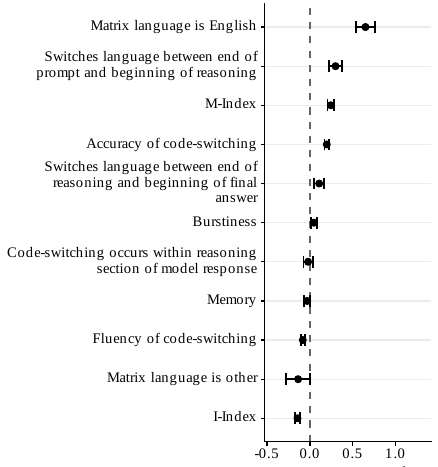}
    \caption{Relationship of code-switching behaviors and change in log odds of correct final answer for models fine-tuned and evaluated on \encore.}
    \label{fig:1panel}
\end{wrapfigure}

\textbf{Statistical modeling of relationship between code-switching and performance (Figure \ref{fig:1panel}).}
On the pooled test-set outputs from Global MMLU and Global PIQA, we fit a mixed model similar to the one in section \ref{sec:taxonomy-approach}. (See appendix \ref{appendix:section4-csw-perf} for details.)

\textbf{Statistical modeling of the effect of SFT tasks on code-switching behaviors (Figure \ref{fig:heatmap-main}).} For behavior $j$, reasoning trace $t$, prompt $p$, language $\ell$, 
$g(\mathbb{E}[b_{j,t}]) = \alpha_{j0} + \sum_s \alpha_{js}\,\mathbb{1}[\text{task}_t = s] + \delta_{j,\text{dataset}} + \delta_{j,\text{model}} + \delta_{j,\text{mt}}\, z(\text{mt}_\ell) + u_{j\ell} + v_{jp}$. $g$ is the logit link for binary behaviors (e.g., ``English is the matrix language'') and the identity link otherwise. $\text{mt}_\ell$ is the MT quality reported by \citet{communication2023seamlessmultilingualexpressivestreaming} or \citet{nllbteam2022languageleftbehindscaling} for $\ell$, $u_{j\ell} \sim \mathcal{N}(0,\sigma^2_{j\ell})$, and $v_{jp} \sim \mathcal{N}(0,\sigma^2_{jp})$.
The baseline (non-fine-tuned) model is the reference level; $\alpha_{js}$ is the change in behavior $j$ associated with SFT task $s$. We drop number of SFT examples from the regression because it is collinear with $s$.

\textbf{Statistical modeling of difference in SFT effects on reasoning task accuracy in low- v. mid/high-resource languages (Figure \ref{fig:acc-gains})}. Coefficients are on the log-odds scale, so we report
average marginal effects (AMEs) on the probability scale. (See appendix \ref{sec:ame} for how reasoning task accuracy models are fit prior to calculating differences in AMEs.)
For condition $c$ and responses $S$, the AME is the mean difference in predicted probability of a correct
answer when every response in $S$ is counterfactually assigned to condition $c$ versus
to the baseline (non-fine-tuned) model: $\mathrm{AME}_c(S) = \frac{1}{|S|} \sum_{i \in S} \bigl[\hat{p}_i(c) - \hat{p}_i(\text{baseline})\bigr]$.
We group languages by resource level: Low
(Amharic, Igbo, Swahili, Yoruba) or Mid/High (Hindi, Indonesian, Malay) \citep{singh2025globalmmluunderstandingaddressing}. Then $\Delta_c = \mathrm{AME}_c(S_{\text{Low}}) - \mathrm{AME}_c(S_{\text{High/Mid}})$. %

\textbf{Mediation analysis (Figure \ref{fig:acc-gains})}. See appendix \ref{appendix:mediation} for details.

\subsection{Results}\label{sec:sft-results}
While the results of section \ref{sec:taxonomy} (\textbf{RQ 2}) inform the design of our SFT interventions, we do not expect identical outcomes in the evaluation of our fine-tuned models, due to differences in the models, languages, and datasets covered, as well as the different random effects structures for modeling the effect of interventions on the code-switching behaviors and the association of code-switching behaviors with accuracy. Therefore, we revisit \textbf{RQ 2} before addressing \textbf{RQ 3}.

\textbf{Code-switching behaviors associated with correct answers (RQ 2).} Figure \ref{fig:1panel} shows the correctness
model from section \ref{sec:approach2} fit to 80,532 test-set responses from the three baseline models and their 120 fine-tuned variants. 
Responses whose matrix language is English
are more likely to be correct than responses dominated by the prompt language ($\beta$ = 0.65). Among code-switched responses, more semantically
accurate code-switching ($\beta$ = 0.20) and a more balanced mix of languages (M-Index, $\beta$ = 0.24)
are associated with correct answers, whereas a high density of code-switching points (I-Index, $\beta$ = -0.15) and higher fluency ($\beta = -0.08$) are associated with incorrect answers. %
Switching language at the reasoning–answer boundary, i.e., reasoning at least partially in one language and answering in another, is also positively associated with correctness ($\beta$ = 0.11). Switching language at the prompt--reasoning boundary is likewise positively associated with correctness ($\beta$ = 0.30). Burstiness ($\beta$ = 0.05) and memory ($\beta = -0.03$) have small but significant associations with reasoning task accuracy. (See appendix \ref{sec:appendix-3panel} for more results.) \textbf{Overall, our analysis suggests that traits such as code-switching accuracy remain influential predictors of reasoning task performance for our fine-tuned models, partially overlapping with our findings from section \ref{sec:taxonomy-results}.}

\begin{figure}[!htbp]
    \centering
    \includegraphics[width=\linewidth,clip, trim={0cm 0cm 0cm 0cm}]{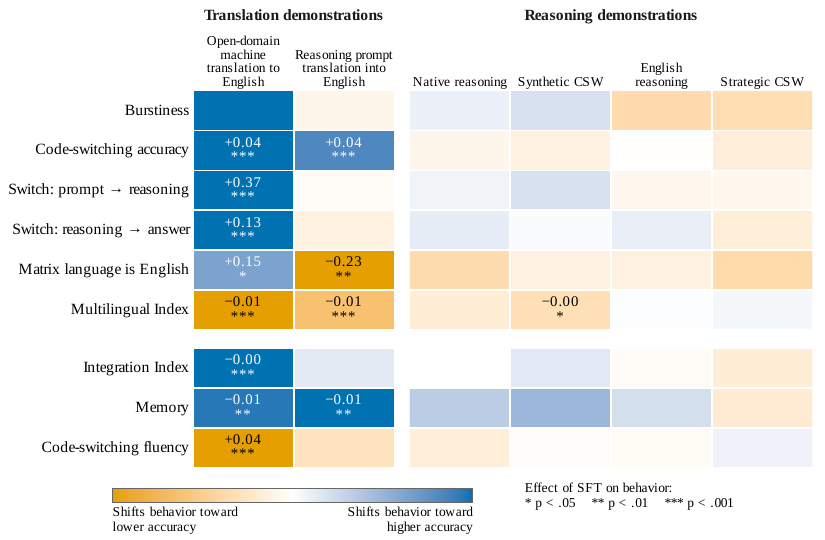}
    \caption{SFT on translation-based tasks has the most significant positive effects on code-switching behaviors associated with final answer accuracy. (Cell labels are effect sizes of interventions; color scale computed within each row.)}
    \label{fig:heatmap-main}
\end{figure}

\textbf{Effects of fine-tuning on code-switching behaviors (RQ 3).} Figure~\ref{fig:heatmap-main} shows the causal effect of each intervention on each behavior that is significantly associated with reasoning task accuracy, relative to the non-fine-tuned baseline models.  
Fine-tuning for open-domain machine translation, which contains no reasoning demonstrations, has the broadest positive effects. Fine-tuning for open-domain MT increases code-switching accuracy ($p<.001$), the odds that the matrix language is English ($p<.05$), and the odds of switching languages at the prompt-reasoning and reasoning-answer boundaries ($p<.001$). All of these behavioral shifts are associated with correct reasoning task answers. Fine-tuning to translate reasoning prompts into English likewise increases code-switching accuracy ($p<.001$). In contrast, fine-tuning on reasoning demonstrations (English, native-language, strategically code-switched, or synthetically code-switched reasoning) produces almost no significant changes in any behavior. \textbf{Demonstrations of translation, rather than reasoning demonstrations, thus appear to be the most effective intervention on model code-switching behaviors, consistent with translation being a central function of code-switching in reasoning (section~\ref{sec:taxonomy-results}).}

\begin{figure}[t]
        \centering
        \includegraphics[clip, trim={0cm 0cm 0cm 0cm},width=\linewidth]{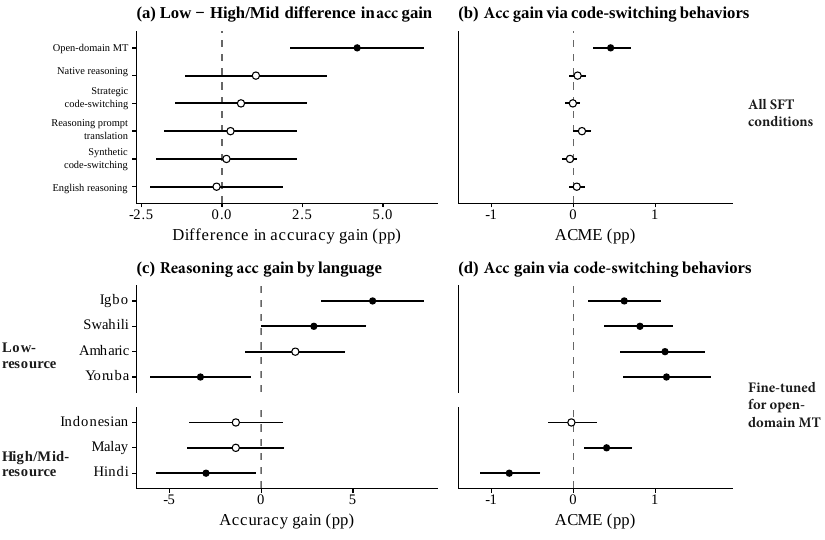}
        \caption{(a) Low-resource languages gain more percentage points (pp) in accuracy (acc) than high- or mid-resource languages under our open-domain MT SFT intervention (p < 0.001). (b) Based on mediation analysis (appendix \ref{appendix:mediation}), the Average Causal Mediation Effects (ACME) on reasoning task accuracy (a measure of how much gains in accuracy are predicted by code-switching behaviors, separate from gains from the SFT tasks alone) are significant and positive for the open-domain MT SFT intervention. (c) When fine-tuned for open-domain MT, models experience reasoning task accuracy gains of 2.26 pp (pooled across the four low-resource languages). (d) Code-switching behaviors predict significant accuracy gains (ACME) in low-resource languages. (Resource levels: \citet{singh2025globalmmluunderstandingaddressing}.)}
        \label{fig:acc-gains}
\end{figure}

\textbf{Fine-tuning on open-domain machine translation increases reasoning task accuracy in low-resource languages (RQ 3; Figure \ref{fig:acc-gains}).} Moreover, low-resource languages are exactly those languages for which this performance improvement is most impactful, with the open-domain machine translation SFT task increasing reasoning task accuracy by 4.20 percentage points (pp) more in low-resource languages than mid/high-resource languages (Figure \ref{fig:acc-gains} (a); p < 0.001). The improvements for low-resource languages are especially striking given the low baseline performance and data scarcity in these languages. Based on mediation analysis (appendix \ref{appendix:mediation}), the Average Causal Mediation Effects (ACME) on reasoning task accuracy (a measure of how much gains in accuracy are predicted by code-switching behaviors, separate from gains from the SFT tasks alone) are significant and positive for the open-domain MT translation SFT intervention (Figure \ref{fig:acc-gains} (b)). When fine-tuned for open-domain MT, models experience reasoning task accuracy gains of 2.26 pp (pooled across the four low-resource languages; Figure \ref{fig:acc-gains} (c)). Code-switching behaviors also predict significant accuracy gains (ACME) in low-resource languages (Figure \ref{fig:acc-gains} (d)). \textbf{Our results suggest that fine-tuning on open-domain machine translation can promote helpful code-switched reasoning behaviors and boost overall reasoning task performance in low-resource languages, requiring model checkpoints with only moderate multilingual capabilities as a starting point.}

\textbf{Student models still benefit from demonstrations generated by Gemma with our framework, showing that benefits generalize across teacher models.} See appendix \ref{sec:gemma} for plots.

\section{Conclusion}
We address the gap in work on how LLMs code-switch during reasoning, types of code-switching helpful for reasoning, and fine-tuning interventions to teach code-switched reasoning, by introducing a framework, data, code, and models for code-switched reasoning. We find code-switching characteristics such as the matrix language, semantic accuracy, and density all predict performance. Our framework significantly increases code-switching behaviors predictive of better reasoning task performance, especially for the lowest-resource languages. Our work establishes code-switching as a means for improving reasoning capabilities in languages with limited data resources.

\subsection*{AI use statement}

In this work, we used generative AI tools for generating synthetic data sets, help developing theoretical models and conceptual frameworks, designing and providing feedback on research methodology and experiments, implementing methods, assisting with translation, cleaning and reformatting datasets, supporting qualitative and thematic data analysis, and interpreting results.
We have not used generative AI tools for formulating mathematical claims, providing critical ingredients for proving mathematical claims, assisting in the writing of proofs, or proposing or refining hypotheses,
and the aforementioned tasks are not applicable to this work.
Additionally, we used generative AI tools for creating and modifying scientific figures and images, suggesting experimental parameters, creating and editing software code, creating artifacts, brainstorming, sourcing/searching for information, editing the research paper to improve readability, identifying relevant literature, and proposing the title and keywords for the research paper. We have reviewed all AI-assisted work. LLM-generated code was verified and tested for correctness, synthesized data was manually reviewed for quality, and cited literature was human-reviewed before incorporation into the paper. We take responsibility for the final content of this work,
including text, claims or artifacts produced with the aid of generative AI.

\subsection*{Reproducibility statement}

To support reproduction of our work, we will release our \core and \encore datasets along with all trained adapters and evaluation scripts needed to reproduce the results and figures in this work under open licenses. Additionally, we have included the LLM prompts we used for data synthesis and analysis in the appendix, and explained details of model hyperparameters/configurations and regression setups in both the main text of the paper and the appendix. We also provide more detailed explanations of our overall data synthesis process in the appendix.

\subsubsection*{Acknowledgments}
This material is based upon work supported by the National Science Foundation under Award No. DGE 2241144 and Award No. 2143529.

\bibliography{iclr2027_conference}

\begin{thebibliography}{74}
\providecommand{\natexlab}[1]{#1}
\providecommand{\url}[1]{\texttt{#1}}
\expandafter\ifx\csname urlstyle\endcsname\relax
  \providecommand{\doi}[1]{doi: #1}\else
  \providecommand{\doi}{doi: \begingroup \urlstyle{rm}\Url}\fi

\bibitem[Ahia et~al.(2023)Ahia, Kumar, Gonen, Kasai, Mortensen, Smith, and Tsvetkov]{ahia-etal-2023-languages}
Orevaoghene Ahia, Sachin Kumar, Hila Gonen, Jungo Kasai, David Mortensen, Noah Smith, and Yulia Tsvetkov.
\newblock Do all languages cost the same? tokenization in the era of commercial language models.
\newblock In Houda Bouamor, Juan Pino, and Kalika Bali (eds.), \emph{Proceedings of the 2023 Conference on Empirical Methods in Natural Language Processing}, pp.\  9904--9923, Singapore, December 2023. Association for Computational Linguistics.
\newblock \doi{10.18653/v1/2023.emnlp-main.614}.
\newblock URL \url{https://aclanthology.org/2023.emnlp-main.614/}.

\bibitem[Anthropic(2025)]{claude2.7sonnet}
Anthropic.
\newblock Claude 3.7 sonnet and claude code, 2025.
\newblock URL \url{https://www.anthropic.com/news/claude-3-7-sonnet}.

\bibitem[Apertus et~al.(2025)Apertus, Hernández-Cano, Hägele, Huang, Romanou, Solergibert, Pasztor, Messmer, Garbaya, Ďurech, Hakimi, Giraldo, Ismayilzada, Foroutan, Moalla, Chen, Sabolčec, Xu, Aerni, AlKhamissi, Mariñas, Amani, Ansaripour, Badanin, Benoit, Boros, Browning, Bösch, Böther, Canova, Challier, Charmillot, Coles, Deriu, Devos, Drescher, Dzenhaliou, Ehrmann, Fan, Fan, Gao, Gila, Grandury, Hashemi, Hoyle, Jiang, Klein, Kucharavy, Kucherenko, Lübeck, Machacek, Manitaras, Marfurt, Matoba, Matrenok, Mendonça, Mohamed, Montariol, Mouchel, Najem-Meyer, Ni, Oliva, Pagliardini, Palme, Panferov, Paoletti, Passerini, Pavlov, Poiroux, Ponkshe, Ranchin, Rando, Sauser, Saydaliev, Sayfiddinov, Schneider, Schuppli, Scialanga, Semenov, Shridhar, Singhal, Sotnikova, Sternfeld, Tarun, Teiletche, Vamvas, Yao, Zhao, Ilic, Klimovic, Krause, Gulcehre, Rosenthal, Ash, Tramèr, VandeVondele, Veraldi, Rajman, Schulthess, Hoefler, Bosselut, Jaggi, and Schlag]{apertus2025apertusdemocratizingopencompliant}
Project Apertus, Alejandro Hernández-Cano, Alexander Hägele, Allen~Hao Huang, Angelika Romanou, Antoni-Joan Solergibert, Barna Pasztor, Bettina Messmer, Dhia Garbaya, Eduard~Frank Ďurech, Ido Hakimi, Juan~García Giraldo, Mete Ismayilzada, Negar Foroutan, Skander Moalla, Tiancheng Chen, Vinko Sabolčec, Yixuan Xu, Michael Aerni, Badr AlKhamissi, Inés~Altemir Mariñas, Mohammad~Hossein Amani, Matin Ansaripour, Ilia Badanin, Harold Benoit, Emanuela Boros, Nicholas Browning, Fabian Bösch, Maximilian Böther, Niklas Canova, Camille Challier, Clement Charmillot, Jonathan Coles, Jan Deriu, Arnout Devos, Lukas Drescher, Daniil Dzenhaliou, Maud Ehrmann, Dongyang Fan, Simin Fan, Silin Gao, Miguel Gila, María Grandury, Diba Hashemi, Alexander Hoyle, Jiaming Jiang, Mark Klein, Andrei Kucharavy, Anastasiia Kucherenko, Frederike Lübeck, Roman Machacek, Theofilos Manitaras, Andreas Marfurt, Kyle Matoba, Simon Matrenok, Henrique Mendonça, Fawzi~Roberto Mohamed, Syrielle Montariol, Luca Mouchel, Sven Najem-Meyer,
  Jingwei Ni, Gennaro Oliva, Matteo Pagliardini, Elia Palme, Andrei Panferov, Léo Paoletti, Marco Passerini, Ivan Pavlov, Auguste Poiroux, Kaustubh Ponkshe, Nathan Ranchin, Javi Rando, Mathieu Sauser, Jakhongir Saydaliev, Muhammad~Ali Sayfiddinov, Marian Schneider, Stefano Schuppli, Marco Scialanga, Andrei Semenov, Kumar Shridhar, Raghav Singhal, Anna Sotnikova, Alexander Sternfeld, Ayush~Kumar Tarun, Paul Teiletche, Jannis Vamvas, Xiaozhe Yao, Hao Zhao, Alexander Ilic, Ana Klimovic, Andreas Krause, Caglar Gulcehre, David Rosenthal, Elliott Ash, Florian Tramèr, Joost VandeVondele, Livio Veraldi, Martin Rajman, Thomas Schulthess, Torsten Hoefler, Antoine Bosselut, Martin Jaggi, and Imanol Schlag.
\newblock Apertus: Democratizing open and compliant llms for global language environments, 2025.
\newblock URL \url{https://arxiv.org/abs/2509.14233}.

\bibitem[Barnett et~al.(2000)Barnett, Codó, Eppler, Forcadell, Gardner-Chloros, van Hout, Moyer, Torras, Turell, Sebba, Starren, and Wensing]{automatic}
Ruthanna Barnett, Eva Codó, Eva Eppler, Montse Forcadell, Penelope Gardner-Chloros, Roeland van Hout, Melissa Moyer, Maria~Carme Torras, Maria~Teresa Turell, Mark Sebba, Marianne Starren, and Sietse Wensing.
\newblock Automatic analysis.
\newblock \emph{International Journal of Bilingualism}, 4\penalty0 (2):\penalty0 198--208, 2000.
\newblock \doi{10.1177/13670069000040020107}.
\newblock URL \url{https://doi.org/10.1177/13670069000040020107}.

\bibitem[Barrault et~al.(2023)Barrault, Chung, Meglioli, Dale, Dong, Duppenthaler, Duquenne, Ellis, Elsahar, Haaheim, Hoffman, Hwang, Inaguma, Klaiber, Kulikov, Li, Licht, Maillard, Mavlyutov, Rakotoarison, Sadagopan, Ramakrishnan, Tran, Wenzek, Yang, Ye, Evtimov, Fernandez, Gao, Hansanti, Kalbassi, Kallet, Kozhevnikov, Gonzalez, Roman, Touret, Wong, Wood, Yu, Andrews, Balioglu, Chen, Costa-jussà, Elbayad, Gong, Guzmán, Heffernan, Jain, Kao, Lee, Ma, Mourachko, Peloquin, Pino, Popuri, Ropers, Saleem, Schwenk, Sun, Tomasello, Wang, Wang, Wang, and Williamson]{communication2023seamlessmultilingualexpressivestreaming}
Loïc Barrault, Yu-An Chung, Mariano~Coria Meglioli, David Dale, Ning Dong, Mark Duppenthaler, Paul-Ambroise Duquenne, Brian Ellis, Hady Elsahar, Justin Haaheim, John Hoffman, Min-Jae Hwang, Hirofumi Inaguma, Christopher Klaiber, Ilia Kulikov, Pengwei Li, Daniel Licht, Jean Maillard, Ruslan Mavlyutov, Alice Rakotoarison, Kaushik~Ram Sadagopan, Abinesh Ramakrishnan, Tuan Tran, Guillaume Wenzek, Yilin Yang, Ethan Ye, Ivan Evtimov, Pierre Fernandez, Cynthia Gao, Prangthip Hansanti, Elahe Kalbassi, Amanda Kallet, Artyom Kozhevnikov, Gabriel~Mejia Gonzalez, Robin~San Roman, Christophe Touret, Corinne Wong, Carleigh Wood, Bokai Yu, Pierre Andrews, Can Balioglu, Peng-Jen Chen, Marta~R. Costa-jussà, Maha Elbayad, Hongyu Gong, Francisco Guzmán, Kevin Heffernan, Somya Jain, Justine Kao, Ann Lee, Xutai Ma, Alex Mourachko, Benjamin Peloquin, Juan Pino, Sravya Popuri, Christophe Ropers, Safiyyah Saleem, Holger Schwenk, Anna Sun, Paden Tomasello, Changhan Wang, Jeff Wang, Skyler Wang, and Mary Williamson.
\newblock Seamless: Multilingual expressive and streaming speech translation, 2023.
\newblock URL \url{https://arxiv.org/abs/2312.05187}.

\bibitem[Begum et~al.(2016)Begum, Bali, Choudhury, Rudra, and Ganguly]{begum-etal-2016-functions}
Rafiya Begum, Kalika Bali, Monojit Choudhury, Koustav Rudra, and Niloy Ganguly.
\newblock Functions of code-switching in tweets: An annotation framework and some initial experiments.
\newblock In Nicoletta Calzolari, Khalid Choukri, Thierry Declerck, Sara Goggi, Marko Grobelnik, Bente Maegaard, Joseph Mariani, Helene Mazo, Asuncion Moreno, Jan Odijk, and Stelios Piperidis (eds.), \emph{Proceedings of the Tenth International Conference on Language Resources and Evaluation ({LREC}'16)}, pp.\  1644--1650, Portoro{\v{z}}, Slovenia, May 2016. European Language Resources Association (ELRA).
\newblock URL \url{https://aclanthology.org/L16-1260/}.

\bibitem[Belani \& Flanigan(2023)Belani and Flanigan]{belani-flanigan-2023-automatic}
Ritu Belani and Jeffrey Flanigan.
\newblock Automatic identification of code-switching functions in speech transcripts.
\newblock In Anna Rogers, Jordan Boyd-Graber, and Naoaki Okazaki (eds.), \emph{Findings of the Association for Computational Linguistics: ACL 2023}, pp.\  7438--7448, Toronto, Canada, July 2023. Association for Computational Linguistics.
\newblock \doi{10.18653/v1/2023.findings-acl.469}.
\newblock URL \url{https://aclanthology.org/2023.findings-acl.469/}.

\bibitem[Bercovich et~al.(2025)Bercovich, Levy, Golan, Dabbah, El-Yaniv, Puny, Galil, Moshe, Ronen, Nabwani, Shahaf, Tropp, Karpas, Zilberstein, Zeng, Singhal, Bukharin, Zhang, Konuk, Shen, Mahabaleshwarkar, Kartal, Suhara, Delalleau, Chen, Wang, Mosallanezhad, Renduchintala, Qian, Rekesh, Jia, Majumdar, Noroozi, Ahmad, Narenthiran, Ficek, Samadi, Huang, Jain, Gitman, Moshkov, Du, Toshniwal, Armstrong, Kisacanin, Novikov, Gitman, Bakhturina, Varshney, Narsimhan, Scowcroft, Kamalu, Su, Kong, Kliegl, Karimi, Lin, Satheesh, Parmar, Gundecha, Norick, Jennings, Prabhumoye, Akter, Patwary, Khattar, Narayanan, Waleffe, Zhang, Su, Huang, Kong, Chadha, Jain, Harvey, Segal, Huang, Kashirsky, McQueen, Putterman, Lam, Venkatesan, Wu, Nguyen, Kilaru, Wang, Warno, Somasamudramath, Bhaskar, Dong, Assaf, Mor, Argov, Junkin, Romanenko, Larroy, Katariya, Rovinelli, Balas, Edelman, Bhiwandiwalla, Subramaniam, Ithape, Ramamoorthy, Wu, Velury, Almog, Daw, Fridman, Galinkin, Evans, Ghosh, Luna, Derczynski, Pope, Long, Schneider,
  Siman, Grzegorzek, Ribalta, Katariya, Alexiuk, Conway, Saar, Guan, Pawelec, Prayaga, Kuchaiev, Ginsburg, Olabiyi, Briski, Cohen, Catanzaro, Alben, Geifman, and Chung]{bercovich2025llamanemotronefficientreasoningmodels}
Akhiad Bercovich, Itay Levy, Izik Golan, Mohammad Dabbah, Ran El-Yaniv, Omri Puny, Ido Galil, Zach Moshe, Tomer Ronen, Najeeb Nabwani, Ido Shahaf, Oren Tropp, Ehud Karpas, Ran Zilberstein, Jiaqi Zeng, Soumye Singhal, Alexander Bukharin, Yian Zhang, Tugrul Konuk, Gerald Shen, Ameya~Sunil Mahabaleshwarkar, Bilal Kartal, Yoshi Suhara, Olivier Delalleau, Zijia Chen, Zhilin Wang, David Mosallanezhad, Adi Renduchintala, Haifeng Qian, Dima Rekesh, Fei Jia, Somshubra Majumdar, Vahid Noroozi, Wasi~Uddin Ahmad, Sean Narenthiran, Aleksander Ficek, Mehrzad Samadi, Jocelyn Huang, Siddhartha Jain, Igor Gitman, Ivan Moshkov, Wei Du, Shubham Toshniwal, George Armstrong, Branislav Kisacanin, Matvei Novikov, Daria Gitman, Evelina Bakhturina, Prasoon Varshney, Makesh Narsimhan, Jane~Polak Scowcroft, John Kamalu, Dan Su, Kezhi Kong, Markus Kliegl, Rabeeh Karimi, Ying Lin, Sanjeev Satheesh, Jupinder Parmar, Pritam Gundecha, Brandon Norick, Joseph Jennings, Shrimai Prabhumoye, Syeda~Nahida Akter, Mostofa Patwary, Abhinav Khattar,
  Deepak Narayanan, Roger Waleffe, Jimmy Zhang, Bor-Yiing Su, Guyue Huang, Terry Kong, Parth Chadha, Sahil Jain, Christine Harvey, Elad Segal, Jining Huang, Sergey Kashirsky, Robert McQueen, Izzy Putterman, George Lam, Arun Venkatesan, Sherry Wu, Vinh Nguyen, Manoj Kilaru, Andrew Wang, Anna Warno, Abhilash Somasamudramath, Sandip Bhaskar, Maka Dong, Nave Assaf, Shahar Mor, Omer~Ullman Argov, Scot Junkin, Oleksandr Romanenko, Pedro Larroy, Monika Katariya, Marco Rovinelli, Viji Balas, Nicholas Edelman, Anahita Bhiwandiwalla, Muthu Subramaniam, Smita Ithape, Karthik Ramamoorthy, Yuting Wu, Suguna~Varshini Velury, Omri Almog, Joyjit Daw, Denys Fridman, Erick Galinkin, Michael Evans, Shaona Ghosh, Katherine Luna, Leon Derczynski, Nikki Pope, Eileen Long, Seth Schneider, Guillermo Siman, Tomasz Grzegorzek, Pablo Ribalta, Monika Katariya, Chris Alexiuk, Joey Conway, Trisha Saar, Ann Guan, Krzysztof Pawelec, Shyamala Prayaga, Oleksii Kuchaiev, Boris Ginsburg, Oluwatobi Olabiyi, Kari Briski, Jonathan Cohen, Bryan
  Catanzaro, Jonah Alben, Yonatan Geifman, and Eric Chung.
\newblock Llama-nemotron: Efficient reasoning models, 2025.
\newblock URL \url{https://arxiv.org/abs/2505.00949}.

\bibitem[Bullock et~al.(2018)Bullock, Guzm{\'a}n, Serigos, Sharath, and Toribio]{bullock-etal-2018-predicting}
Barbara Bullock, Wally Guzm{\'a}n, Jacqueline Serigos, Vivek Sharath, and Almeida~Jacqueline Toribio.
\newblock Predicting the presence of a matrix language in code-switching.
\newblock In Gustavo Aguilar, Fahad AlGhamdi, Victor Soto, Thamar Solorio, Mona Diab, and Julia Hirschberg (eds.), \emph{Proceedings of the Third Workshop on Computational Approaches to Linguistic Code-Switching}, pp.\  68--75, Melbourne, Australia, July 2018. Association for Computational Linguistics.
\newblock \doi{10.18653/v1/W18-3208}.
\newblock URL \url{https://aclanthology.org/W18-3208/}.

\bibitem[Bullock \& Toribio(2009)Bullock and Toribio]{cambridge-csw}
Barbara~E. Bullock and Almeida~Jacqueline Toribio.
\newblock \emph{The Cambridge Handbook of Linguistic Code-switching}.
\newblock Cambridge University Press, 2009.

\bibitem[Chang et~al.(2026)Chang, Arnett, Sadallah, Eldesokey, Kashar, Daud, Olanihun, Mohammed, Praise, Sharma, Gupta, Merin, Bremang, Iyigun, Simplício, Essouaied, Chorana, Eppa, Oladipo, Kuri, Ramesh, Dorkin, Kondoro, Aji, Çetintaş, Hanbury, Dembele, Niksarli, Álvaro Arroyo, Bajand, Khanna, Chkhaidze, Condez, Hartl, Mkhonto, Hoblitzell, Tran, Poulis, Majumder, Chaudhary, Vacalopoulou, Wong, Simonsen, Kovalev, Nayak, S, Lana, Purwarianti, Alhafni, Busole, Ghanem, Nathani, Đurić, Ogundipe, Agbonile, Bergsson, Fischer, Tutar, Çınar, Kane, Udomcharoenchaikit, Helwe, Nerella, Liu, Nwokolo, Homan, Sampebgo, España-Bonet, Amol, Lee, Smart, Arad, Dzenhaliou, Choi, Liu, Semedo, Anugraha, Popoola, Mataciunas, Nyaboke, Owusu, Kumar, Tavares, Glória-Silva, Goyal, Lee, Buchanan, Anajemba, Grace, Mickel, Herranen, Acharya, Nisar, Anand, Habumuremyi, Ajiboye, Yulianrifat, Adenuga, Rudnicka, Itiola, Butt, Sheikh, Thekkekara, Haouari, Nsengiyumva, Ilasariya, Tjiaranata, Laakom, Grasso, Periti, Orabona, Solomon,
  Winata, Ngo, Udhedhe-oze, Vinagre, Challagolla, Urbizu-Garmendia, Vadithya, Son, Abdykadyrova, Mohapatra, Ullah, Einarsson, Hu, Saffari, Zaidi, Zhang, Shairah, Vuong, Kuulmets, Patel, Bouamor, Yu, Debess, İbrahim Ethem~Deveci, Hanif, Cho, Vieira, Calvo, Manzi, Salifou, Daud, Yusuf, Itzhak, Zhelyazkov, Belashkin, Spada, Brinton, Isbarov, Čibej, Kocoń, Cuhel, Krito, Purbey, Za, Mickel, Kunz, Ratovondranto, Varsha, Jeong, Dávalos, Lee, Magalhães, Yi, Kim, Chataignon, Imperial, Thevakumar, Land, Alekseenko, Jiang, Kim, Sirts, R, V, Tshinu, Kukk, Ponkshe, Huseynova, He, Enevoldsen, Alvarez, Zaman, Mrini, Kyars, Gour, Lainitha, Kruusmaa, Mukherjee, Chouhan, Castro, Porrino-Moscoso, Nzambi, Choshen, Sencan, Øvrelid, Alazraki, Jones, Ehimen-Ugbede, Thevakumar, Thavarasa, Malik, Keita, Jangid, Santis, Garcia, Šuppa, D'Ciofalo, Ojastu, Attaullah, Sikander, Narayan, Skandalis, Mehak, İlteriş Bozkurt, Bayu, Velayuthan, Vizo, Leventhal, Marcińczuk, Almasi, Potočnjak, Bangera, Shafiei, Ansari, Sharma, Indoria,
  Rehman, Habibi, Kolić, Kınay, Galant, Rathore, Permpredanun, Maugin, Norman, Corrêa, Ljubešić, Thomas, de~Silva, Joshi, Ponkshe, Habash, Udeze, Thomas, Ligeti-Nagy, Coulibaly, Ogundepo, Buliaminu, Fejiro, God'spraise, Samuel, Oluwaseun, Akindejoye, Snissarenko, Chiemezie, Kınay, Tursun, Joshua, Fiyinfoluwa, Rodríguez, Gamallo, Arora, Valente, Rupnik, Ekiugbo, Agarwal, Sahoo, Prokopidis, Niau-Puhipau, Yahya, Mignone, Singhal, Raja, Kadiyala, Merx, Larsen, Rajalakshmi, Ghosh, Oji, Solis, Guerra, Zawar, Bashir, Alzaabi, Sandeep, Batchu, Kantareddy, Muzammil, Pranida, Buchanan, Rutunda, Land, Sulollari, Ali, Sapkota, Kengatharaiyer, Tautvaisas, Sen, Banerjee, Diarra, Afolayan, M, Lee, Shah, Venkitachalam, Djurabaeva, Ibejih, Dutta, Gupta, Suárez, Ahmadi, Sukumar, Song, A, Sofianopoulos, Simon, Benčina, Gvasalia, More, Dragazis, Milosavljević, Kaufhold, S, Alrashed, Ranathunga, Someya, Pungeršek, Haklay, Jibril, Aoyama, Abashidze, Cruz, Blevins, Nikas, Idoko, Do, Chubakov, Munda, Owoeye, Gargiani,
  Rathore, Johannesen, Ugwu, Putra, Kumar, Arzt, Konovalov, Nedumpozhimana, Ondrejova, Horbik, Kummitha, Dinić, Sewunetie, Wu, Zhao, Diarra, Nikankin, Mathur, Bagla, Bangera, Chen, Li, Xavier, Belinkov, Alyafeai, Batozargalova, Shan, Tam, Tang, Nadova, Abbasi, Biderman, Stap, Ataman, Schmidt, Gonen, Wang, and Adelani]{chang2026globalpiqaevaluatingcommonsense}
Tyler~A. Chang, Catherine Arnett, Abdelrahman Sadallah, Abdelrahman Eldesokey, Abeer Kashar, Abolade Daud, Abosede~Grace Olanihun, Adamu~Labaran Mohammed, Adeyemi Praise, Adhikarimayum~Meerajita Sharma, Aditi Gupta, Adril~Putra Merin, Adwoa Bremang, Afitab Iyigun, Afonso Simplício, Ahmed Essouaied, Aicha Chorana, Akhil Eppa, Akintunde Oladipo, Akriti Kuri, Akshay Ramesh, Aleksei Dorkin, Alfred~Malengo Kondoro, Alham~Fikri Aji, Ali~Eren Çetintaş, Allan Hanbury, Alou Dembele, Alp Niksarli, Álvaro Arroyo, Amin Bajand, Amol Khanna, Ana Chkhaidze, Ana~Carolina Condez, Anamaria-Roberta Hartl, Andiswa Mkhonto, Andrew Hoblitzell, Andrew Tran, Angelos Poulis, Anirban Majumder, Anjali Chaudhary, Anna Vacalopoulou, Annette Kuuipolani~Kanahele Wong, Annika Simonsen, Anton Kovalev, Anupam Nayak, Ashvanth S, Ayodeji Lana, Ayu Purwarianti, Bashar Alhafni, Benedict Busole, Bernard Ghanem, Bharti Nathani, Biljana~Stojanovska Đurić, Blessing Ogundipe, Bolaotan Agbonile, Bragi Bergsson, Bruce~Torres Fischer, Burak Tutar,
  Burcu Çınar, Cade Kane, Can Udomcharoenchaikit, Chadi Helwe, Chaithra~Reddy Nerella, Chen~Cecilia Liu, Chiamaka Nwokolo, Christopher Homan, Clément Sampebgo, Cristina España-Bonet, Cynthia Amol, Daeyoep Lee, Dan~Saattrup Smart, Dana Arad, Daniil Dzenhaliou, Dasol Choi, David Liu, David Semedo, David Anugraha, Deborah Popoola, Deividas Mataciunas, Delphine Nyaboke, Dennis Owusu, Dhyuthy~Krishna Kumar, Diogo Tavares, Diogo Glória-Silva, Divyanshu Goyal, DongGeon Lee, E.~Kelly Buchanan, Ebele~Nwamaka Anajemba, Egonu~Ngozi Grace, Elena Mickel, Elias Herranen, Eliza Acharya, Eman Nisar, Emile Anand, Emmanuel Habumuremyi, Emuobonuvie~Maria Ajiboye, Eryawan~Presma Yulianrifat, Esther Adenuga, Ewa Rudnicka, Faith Itiola, Faran~Taimoor Butt, Fareeha~Fayyaz Sheikh, Fathima Thekkekara, Fatima Haouari, Faustin Nsengiyumva, Fenal~Ashokbhai Ilasariya, Filbert~Aurelian Tjiaranata, Firas Laakom, Francesca Grasso, Francesco Periti, Francesco Orabona, Gbenga~Kayode Solomon, Genta~Indra Winata, Gia~Nghia Ngo, Gloria
  Udhedhe-oze, Gonçalo Vinagre, Gopi Naga Sai~Ram Challagolla, Gorka Urbizu-Garmendia, Gouthami Vadithya, Guijin Son, Gulnaz Abdykadyrova, Gyan~Swaroop Mohapatra, Hafeez Ullah, Hafsteinn Einarsson, Hai Hu, Hamidreza Saffari, Hamza Zaidi, Haopeng Zhang, Harethah~Abu Shairah, Harry Vuong, Hele-Andra Kuulmets, Hitesh~Laxmichand Patel, Houda Bouamor, Hwanjo Yu, Iben~Nyholm Debess, İbrahim Ethem~Deveci, Ikhlasul~Akmal Hanif, Ikhyun Cho, Inês Vieira, Inês Calvo, Isaac Manzi, Ismael~Illa Salifou, Ismail Daud, Ismail Yusuf, Itay Itzhak, Ivan Zhelyazkov, Ivan Belashkin, Ivan Spada, Jacob Brinton, Jafar Isbarov, Jaka Čibej, Jan Kocoń, Jan Cuhel, Jauza Krito, Jebish Purbey, Jennifer Za, Jennifer Mickel, Jenny Kunz, Jessica Ratovondranto, Jeyarajalingam Varsha, Jihae Jeong, Jimena~Tena Dávalos, Jinu Lee, João Magalhães, John Seon~Keun Yi, Jongin Kim, Joseph Chataignon, Joseph~Marvin Imperial, Jubeerathan Thevakumar, Judith Land, Julia Alekseenko, Junchen Jiang, Jungwhan Kim, Kairit Sirts, Kamesh R, Kamesh V,
  Kanda Tshinu, Kätriin Kukk, Kaustubh Ponkshe, Kavsar Huseynova, Ke~He, Kenneth Enevoldsen, Kent~Joshua Alvarez, Kerem Zaman, Khalil Mrini, Kian Kyars, Komal Gour, Krishnakumar Lainitha, Krister Kruusmaa, Kunal Mukherjee, Kusum Chouhan, Laura Castro, Laura~M. Porrino-Moscoso, Lenny Sivi~Za Nzambi, Leshem Choshen, Levent Sencan, Lilja Øvrelid, Lisa Alazraki, Loretta~Oma Jones, Lovina Ehimen-Ugbede, Luheerathan Thevakumar, Luxshan Thavarasa, Mahnoor Malik, Mamadou~K. Keita, Mansi Jangid, Marco~De Santis, Marcos Garcia, Marek Šuppa, Mariam D'Ciofalo, Marii Ojastu, Marium Attaullah, Maryam Sikander, Mausami Narayan, Maximos Skandalis, Mehak Mehak, Mehmet İlteriş Bozkurt, Melaku Bayu, Menan Velayuthan, Mhasilenuo Vizo, Michael Leventhal, Michał Marcińczuk, Mina Almasi, Mirna Potočnjak, Mithil Bangera, Mohammadamin Shafiei, Mohiba Ansari, Mridul Sharma, Mrityunjaya Indoria, Mughees~Ur Rehman, Muhammad Ravi~Shulthan Habibi, Murat Kolić, Murat~Barkın Kınay, Nada Galant, Naina~Singh Rathore, Naphat
  Permpredanun, Narada Maugin, Nathalie Norman, Nicholas~Kluge Corrêa, Nikola Ljubešić, Nirmal Thomas, Nisansa de~Silva, Nisheeth Joshi, Nitish Ponkshe, Nizar Habash, Nneoma Udeze, Noel Thomas, Noémi Ligeti-Nagy, Nouhoum Coulibaly, Odunayo Ogundepo, Odunayo~Kareemat Buliaminu, Oghojafor~Godswill Fejiro, Okechukwu God'spraise, Olanrewaju Samuel, Olaoye~Deborah Oluwaseun, Olasoji Akindejoye, Olga Snissarenko, Onyinye~Anulika Chiemezie, Orkun Kınay, Osman Tursun, Oyelade~Oluwafemi Joshua, Oyesanmi Fiyinfoluwa, Pablo Rodríguez, Pablo Gamallo, Palak Arora, Pedro Valente, Peter Rupnik, Philip~Oghenesuowho Ekiugbo, Prakhar Agarwal, Pramit Sahoo, Prokopis Prokopidis, Pua Niau-Puhipau, Quadri Yahya, Rachele Mignone, Raghav Singhal, Rahul Raja, Ram Mohan~Rao Kadiyala, Raphael Merx, Rasmus Larsen, Ratnavel Rajalakshmi, Rishav Ghosh, Romina Oji, Ron~Kekeha Solis, Rui Guerra, Rushikesh Zawar, Sa'ad~Nasir Bashir, Saeed Alzaabi, Sahil Sandeep, Sai~Pavan Batchu, Sai~Sandeep Kantareddy, Saleha Muzammil,
  Salsabila~Zahirah Pranida, Sam Buchanan, Samuel Rutunda, Sander Land, Sarah Sulollari, Sardar Ali, Saroj Sapkota, Sarveswaran Kengatharaiyer, Saulius Tautvaisas, Sayambhu Sen, Sayantani Banerjee, Sebastien Diarra, Segun Afolayan, Senthilnathan M, Sewoong Lee, Shaan Shah, Shankar Venkitachalam, Sharifa Djurabaeva, Sharon Ibejih, Shivanya~Shomir Dutta, Siddhant Gupta, Silvia~Paniagua Suárez, Sina Ahmadi, Sivasuthan Sukumar, Siyuan Song, Snegha A, Sokratis Sofianopoulos, Sona~Elza Simon, Sonja Benčina, Sophie Gvasalia, Sphurti More, Spyros Dragazis, Stefan Milosavljević, Stephan~P. Kaufhold, Suba S, Sultan Alrashed, Surangika Ranathunga, Taiga Someya, Taja~Kuzman Pungeršek, Tal Haklay, Tasi'u Jibril, Tatsuya Aoyama, Tea Abashidze, Terenz Jomar~Dela Cruz, Terra Blevins, Themistoklis Nikas, Theresa Idoko, Thu~Mai Do, Tilek Chubakov, Tina Munda, Tobiloba Owoeye, Tommaso Gargiani, Uma Rathore, Uni Johannesen, Uwuma Ugwu, Vallerie~Alexandra Putra, Vanya~Bannihatti Kumar, Varvara Arzt, Vasily Konovalov,
  Vasudevan Nedumpozhimana, Viktoria Ondrejova, Viktoryia Horbik, Vishnu Vardhan~Reddy Kummitha, Vuk Dinić, Walelign Sewunetie, Winston Wu, Xiaojing Zhao, Yacouba Diarra, Yaniv Nikankin, Yash Mathur, Yash Bagla, Yeshil Bangera, Yixi Chen, Yiyuan Li, Yolanda Xavier, Yonatan Belinkov, Zaid Alyafeai, Zhargal Batozargalova, Zhengyang Shan, Zhi~Rui Tam, Zilu Tang, Zuzana Nadova, Baber Abbasi, Stella Biderman, David Stap, Duygu Ataman, Fabian Schmidt, Hila Gonen, Jiayi Wang, and David~Ifeoluwa Adelani.
\newblock Global piqa: Evaluating commonsense reasoning across 100+ languages and cultures, 2026.
\newblock URL \url{https://arxiv.org/abs/2510.24081}.

\bibitem[Chen et~al.(2024{\natexlab{a}})Chen, Zheng, Wu, Gong, Zhang, and Li]{chen-etal-2024-breaking}
Nuo Chen, Zinan Zheng, Ning Wu, Ming Gong, Dongmei Zhang, and Jia Li.
\newblock Breaking language barriers in multilingual mathematical reasoning: Insights and observations.
\newblock In Yaser Al-Onaizan, Mohit Bansal, and Yun-Nung Chen (eds.), \emph{Findings of the {A}ssociation for {C}omputational {L}inguistics: {EMNLP} 2024}, pp.\  7001--7016, Miami, Florida, USA, November 2024{\natexlab{a}}. Association for Computational Linguistics.
\newblock \doi{10.18653/v1/2024.findings-emnlp.411}.
\newblock URL \url{https://aclanthology.org/2024.findings-emnlp.411/}.

\bibitem[Chen et~al.(2024{\natexlab{b}})Chen, Guo, Haddow, and Heafield]{chen-etal-2024-iterative}
Pinzhen Chen, Zhicheng Guo, Barry Haddow, and Kenneth Heafield.
\newblock Iterative translation refinement with large language models.
\newblock In Carolina Scarton, Charlotte Prescott, Chris Bayliss, Chris Oakley, Joanna Wright, Stuart Wrigley, Xingyi Song, Edward Gow-Smith, Rachel Bawden, V{\'i}ctor~M S{\'a}nchez-Cartagena, Patrick Cadwell, Ekaterina Lapshinova-Koltunski, Vera Cabarr{\~a}o, Konstantinos Chatzitheodorou, Mary Nurminen, Diptesh Kanojia, and Helena Moniz (eds.), \emph{Proceedings of the 25th Annual Conference of the European Association for Machine Translation (Volume 1)}, pp.\  181--190, Sheffield, UK, June 2024{\natexlab{b}}. European Association for Machine Translation (EAMT).
\newblock URL \url{https://aclanthology.org/2024.eamt-1.17/}.

\bibitem[Chen et~al.(2025)Chen, Zhang, Hong, Kundu, and Wang]{chen2025sealsteerablereasoningcalibration}
Runjin Chen, Zhenyu Zhang, Junyuan Hong, Souvik Kundu, and Zhangyang Wang.
\newblock Seal: Steerable reasoning calibration of large language models for free, 2025.
\newblock URL \url{https://arxiv.org/abs/2504.07986}.

\bibitem[Chiu et~al.(2025{\natexlab{a}})Chiu, Jiang, Lin, Park, Li, Ravi, Bhatia, Antoniak, Tsvetkov, Shwartz, and Choi]{chiu-etal-2025-culturalbench}
Yu~Ying Chiu, Liwei Jiang, Bill~Yuchen Lin, Chan~Young Park, Shuyue~Stella Li, Sahithya Ravi, Mehar Bhatia, Maria Antoniak, Yulia Tsvetkov, Vered Shwartz, and Yejin Choi.
\newblock {C}ultural{B}ench: A robust, diverse and challenging benchmark for measuring {LM}s' cultural knowledge through human-{AI} red-teaming.
\newblock In Wanxiang Che, Joyce Nabende, Ekaterina Shutova, and Mohammad~Taher Pilehvar (eds.), \emph{Proceedings of the 63rd Annual Meeting of the {A}ssociation for {C}omputational {L}inguistics (Volume 1: Long Papers)}, pp.\  25663--25701, Vienna, Austria, July 2025{\natexlab{a}}. Association for Computational Linguistics.
\newblock ISBN 979-8-89176-251-0.
\newblock \doi{10.18653/v1/2025.acl-long.1247}.
\newblock URL \url{https://aclanthology.org/2025.acl-long.1247/}.

\bibitem[Chiu et~al.(2025{\natexlab{b}})Chiu, Jiang, Lin, Park, Li, Ravi, Bhatia, Antoniak, Tsvetkov, Shwartz, and Choi]{chiu2025culturalbenchrobustdiversechallenging}
Yu~Ying Chiu, Liwei Jiang, Bill~Yuchen Lin, Chan~Young Park, Shuyue~Stella Li, Sahithya Ravi, Mehar Bhatia, Maria Antoniak, Yulia Tsvetkov, Vered Shwartz, and Yejin Choi.
\newblock Culturalbench: A robust, diverse, and challenging cultural benchmark by human-ai culturalteaming, 2025{\natexlab{b}}.
\newblock URL \url{https://arxiv.org/abs/2410.02677}.

\bibitem[Costa-jussà et~al.(2022)Costa-jussà, Cross, Çelebi, Elbayad, Heafield, Heffernan, Kalbassi, Lam, Licht, Maillard, Sun, Wang, Wenzek, Youngblood, Akula, Barrault, Gonzalez, Hansanti, Hoffman, Jarrett, Sadagopan, Rowe, Spruit, Tran, Andrews, Ayan, Bhosale, Edunov, Fan, Gao, Goswami, Guzmán, Koehn, Mourachko, Ropers, Saleem, Schwenk, and Wang]{nllbteam2022languageleftbehindscaling}
Marta~R. Costa-jussà, James Cross, Onur Çelebi, Maha Elbayad, Kenneth Heafield, Kevin Heffernan, Elahe Kalbassi, Janice Lam, Daniel Licht, Jean Maillard, Anna Sun, Skyler Wang, Guillaume Wenzek, Al~Youngblood, Bapi Akula, Loic Barrault, Gabriel~Mejia Gonzalez, Prangthip Hansanti, John Hoffman, Semarley Jarrett, Kaushik~Ram Sadagopan, Dirk Rowe, Shannon Spruit, Chau Tran, Pierre Andrews, Necip~Fazil Ayan, Shruti Bhosale, Sergey Edunov, Angela Fan, Cynthia Gao, Vedanuj Goswami, Francisco Guzmán, Philipp Koehn, Alexandre Mourachko, Christophe Ropers, Safiyyah Saleem, Holger Schwenk, and Jeff Wang.
\newblock No language left behind: Scaling human-centered machine translation, 2022.
\newblock URL \url{https://arxiv.org/abs/2207.04672}.

\bibitem[Das \& Gamb{\"a}ck(2014)Das and Gamb{\"a}ck]{das-gamback-2014-identifying}
Amitava Das and Bj{\"o}rn Gamb{\"a}ck.
\newblock Identifying languages at the word level in code-mixed {I}ndian social media text.
\newblock In Dipti~Misra Sharma, Rajeev Sangal, and Jyoti~D. Pawar (eds.), \emph{Proceedings of the 11th International Conference on Natural Language Processing}, pp.\  378--387, Goa, India, December 2014. NLP Association of India.
\newblock URL \url{https://aclanthology.org/W14-5152/}.

\bibitem[Eberhard et~al.(2026)Eberhard, Simons, and Robinson]{ethnologue}
David~M. Eberhard, Gary~F. Simons, and Alison~J. Robinson.
\newblock Ethnologue: Languages of the world.
\newblock \url{https://www.ethnologue.com/}, 2026.
\newblock Accessed: 2026-03-18.

\bibitem[Gandhi et~al.(2025)Gandhi, Chakravarthy, Singh, Lile, and Goodman]{gandhi2025cognitivebehaviorsenableselfimproving}
Kanishk Gandhi, Ayush Chakravarthy, Anikait Singh, Nathan Lile, and Noah~D. Goodman.
\newblock Cognitive behaviors that enable self-improving reasoners, or, four habits of highly effective stars, 2025.
\newblock URL \url{https://arxiv.org/abs/2503.01307}.

\bibitem[Gao et~al.(2025)Gao, Huang, Zhu, Huang, Li, and Yuan]{gao2025thinkingmultilinguallyempowerllm}
Changjiang Gao, Xu~Huang, Wenhao Zhu, Shujian Huang, Lei Li, and Fei Yuan.
\newblock Could thinking multilingually empower llm reasoning?, 2025.
\newblock URL \url{https://arxiv.org/abs/2504.11833}.

\bibitem[{Gemini Team} et~al.(2025){Gemini Team}, Anil, Borgeaud, Alayrac, Yu, Soricut, Schalkwyk, Dai, Hauth, Millican, Silver, Johnson, Antonoglou, Schrittwieser, Glaese, Chen, Pitler, Lillicrap, Lazaridou, Firat, Molloy, Isard, Barham, Hennigan, Lee, Viola, Reynolds, Xu, Doherty, Collins, Meyer, Rutherford, Moreira, Ayoub, Goel, Krawczyk, Du, Chi, Cheng, Ni, Shah, Kane, Chan, Faruqui, Severyn, Lin, Li, Cheng, Ittycheriah, Mahdieh, Chen, Sun, Tran, Bagri, Lakshminarayanan, Liu, Orban, Güra, Zhou, Song, Boffy, Ganapathy, Zheng, Choe, Ágoston Weisz, Zhu, Lu, Gopal, Kahn, Kula, Pitman, Shah, Taropa, Merey, Baeuml, Chen, Shafey, Zhang, Sercinoglu, Tucker, Piqueras, Krikun, Barr, Savinov, Danihelka, Roelofs, White, Andreassen, von Glehn, Yagati, Kazemi, Gonzalez, Khalman, Sygnowski, Frechette, Smith, Culp, Proleev, Luan, Chen, Lottes, Schucher, Lebron, Rrustemi, Clay, Crone, Kocisky, Zhao, Perz, Yu, Howard, Bloniarz, Rae, Lu, Sifre, Maggioni, Alcober, Garrette, Barnes, Thakoor, Austin, Barth-Maron, Wong,
  Joshi, Chaabouni, Fatiha, Ahuja, Tomar, Senter, Chadwick, Kornakov, Attaluri, Iturrate, Liu, Li, Cogan, Chen, Jia, Gu, Zhang, Grimstad, Hartman, Garcia, Pillai, Devlin, Laskin, de~Las~Casas, Valter, Tao, Blanco, Badia, Reitter, Chen, Brennan, Rivera, Brin, Iqbal, Surita, Labanowski, Rao, Winkler, Parisotto, Gu, Olszewska, Addanki, Miech, Louis, Teplyashin, Brown, Catt, Balaguer, Xiang, Wang, Ashwood, Briukhov, Webson, Ganapathy, Sanghavi, Kannan, Chang, Stjerngren, Djolonga, Sun, Bapna, Aitchison, Pejman, Michalewski, Yu, Wang, Love, Ahn, Bloxwich, Han, Humphreys, Sellam, Bradbury, Godbole, Samangooei, Damoc, Kaskasoli, Arnold, Vasudevan, Agrawal, Riesa, Lepikhin, Tanburn, Srinivasan, Lim, Hodkinson, Shyam, Ferret, Hand, Garg, Paine, Li, Li, Giang, Neitz, Abbas, York, Reid, Cole, Chowdhery, Das, Rogozińska, Nikolaev, Sprechmann, Nado, Zilka, Prost, He, Monteiro, Mishra, Welty, Newlan, Jia, Allamanis, Hu, de~Liedekerke, Gilmer, Saroufim, Rijhwani, Hou, Shrivastava, Baddepudi, Goldin, Ozturel, Cassirer, Xu,
  Sohn, Sachan, Amplayo, Swanson, Petrova, Narayan, Guez, Brahma, Landon, Patel, Zhao, Villela, Wang, Jia, Rahtz, Giménez, Yeung, Keeling, Georgiev, Mincu, Wu, Haykal, Saputro, Vodrahalli, Qin, Cankara, Sharma, Fernando, Hawkins, Neyshabur, Kim, Hutter, Agrawal, Castro-Ros, van~den Driessche, Wang, Yang, yiin Chang, Komarek, McIlroy, Lučić, Zhang, Farhan, Sharman, Natsev, Michel, Bansal, Qiao, Cao, Shakeri, Butterfield, Chung, Rubenstein, Agrawal, Mensch, Soparkar, Lenc, Chung, Pope, Maggiore, Kay, Jhakra, Wang, Maynez, Phuong, Tobin, Tacchetti, Trebacz, Robinson, Katariya, Riedel, Bailey, Xiao, Ghelani, Aroyo, Slone, Houlsby, Xiong, Yang, Gribovskaya, Adler, Wirth, Lee, Li, Kagohara, Pavagadhi, Bridgers, Bortsova, Ghemawat, Ahmed, Liu, Powell, Bolina, Iinuma, Zablotskaia, Besley, Chung, Dozat, Comanescu, Si, Greer, Su, Polacek, Kaufman, Tokumine, Hu, Buchatskaya, Miao, Elhawaty, Siddhant, Tomasev, Xing, Greer, Miller, Ashraf, Roy, Zhang, Ma, Filos, Besta, Blevins, Klimenko, Yeh, Changpinyo, Mu, Chang,
  Pajarskas, Muir, Cohen, Lan, Haridasan, Marathe, Hansen, Douglas, Samuel, Wang, Austin, Lan, Jiang, Chiu, Lorenzo, Sjösund, Cevey, Gleicher, Avrahami, Boral, Srinivasan, Selo, May, Aisopos, Hussenot, Soares, Baumli, Chang, Recasens, Caine, Pritzel, Pavetic, Pardo, Gergely, Frye, Ramasesh, Horgan, Badola, Kassner, Roy, Dyer, Campos, Tomala, Tang, Badawy, White, Mustafa, Lang, Jindal, Vikram, Gong, Caelles, Hemsley, Thornton, Feng, Stokowiec, Zheng, Thacker, Çağlar Ünlü, Zhang, Saleh, Svensson, Bileschi, Patil, Anand, Ring, Tsihlas, Vezer, Selvi, Shevlane, Rodriguez, Kwiatkowski, Daruki, Rong, Dafoe, FitzGerald, Gu-Lemberg, Khan, Hendricks, Pellat, Feinberg, Cobon-Kerr, Sainath, Rauh, Hashemi, Ives, Hasson, Noland, Cao, Byrd, Hou, Wang, Sottiaux, Paganini, Lespiau, Moufarek, Hassan, Shivakumar, van Amersfoort, Mandhane, Joshi, Goyal, Tung, Brock, Sheahan, Misra, Li, Rakićević, Dehghani, Liu, Mittal, Oh, Noury, Sezener, Huot, Lamm, Cao, Chen, Mudgal, Stella, Brooks, Vasudevan, Liu, Chain, Melinkeri,
  Cohen, Wang, Seymore, Zubkov, Goel, Yue, Krishnakumaran, Albert, Hurley, Sano, Mohananey, Joughin, Filonov, Kępa, Eldawy, Lim, Rishi, Badiezadegan, Bos, Chang, Jain, Padmanabhan, Puttagunta, Krishna, Baker, Kalb, Bedapudi, Kurzrok, Lei, Yu, Litvin, Zhou, Wu, Sobell, Siciliano, Papir, Neale, Bragagnolo, Toor, Chen, Anklin, Wang, Feng, Gholami, Ling, Liu, Walter, Moghaddam, Kishore, Adamek, Mercado, Mallinson, Wandekar, Cagle, Ofek, Garrido, Lombriser, Mukha, Sun, Mohammad, Matak, Qian, Peswani, Janus, Yuan, Schelin, David, Garg, He, Duzhyi, Älgmyr, Lottaz, Li, Yadav, Xu, Chinien, Shivanna, Chuklin, Li, Spadine, Wolfe, Mohamed, Das, Dai, He, von Dincklage, Upadhyay, Maurya, Chi, Krause, Salama, Rabinovitch, M, Selvan, Dektiarev, Ghiasi, Guven, Gupta, Liu, Sharma, Shtacher, Paul, Akerlund, Aubet, Huang, Zhu, Zhu, Teixeira, Fritze, Bertolini, Marinescu, Bölle, Paulus, Gupta, Latkar, Chang, Sanders, Wilson, Wu, Tan, Thiet, Doshi, Lall, Mishra, Chen, Luong, Benjamin, Lee, Andrejczuk, Rabiej, Ranjan, Styrc,
  Yin, Simon, Harriott, Bansal, Robsky, Bacon, Greene, Mirylenka, Zhou, Sarvana, Goyal, Andermatt, Siegler, Horn, Israel, Pongetti, Chen, Selvatici, Silva, Wang, Tolins, Guu, Yogev, Cai, Agostini, Shah, Nguyen, Donnaile, Pereira, Friso, Stambler, Kurzrok, Kuang, Romanikhin, Geller, Yan, Jang, Lee, Fica, Malmi, Tan, Banica, Balle, Pham, Huang, Avram, Shi, Singh, Hidey, Ahuja, Saxena, Dooley, Potharaju, O'Neill, Gokulchandran, Foley, Zhao, Dusenberry, Liu, Mehta, Kotikalapudi, Safranek-Shrader, Goodman, Kessinger, Globen, Kolhar, Gorgolewski, Ibrahim, Song, Eichenbaum, Brovelli, Potluri, Lahoti, Baetu, Ghorbani, Chen, Crawford, Pal, Sridhar, Gurita, Mujika, Petrovski, Cedoz, Li, Chen, Santo, Goyal, Punjabi, Kappaganthu, Kwak, LV, Velury, Choudhury, Hall, Shah, Figueira, Thomas, Lu, Zhou, Kumar, Jurdi, Chikkerur, Ma, Yu, Kwak, Ähdel, Rajayogam, Choma, Liu, Barua, Ji, Park, Hellendoorn, Bailey, Bilal, Zhou, Khatir, Sutton, Rzadkowski, Macintosh, Vij, Shagin, Medina, Liang, Zhou, Shah, Bi, Dankovics, Banga,
  Lehmann, Bredesen, Lin, Hoffmann, Lai, Chung, Yang, Balani, Bražinskas, Sozanschi, Hayes, Alcalde, Makarov, Chen, Stella, Snijders, Mandl, Kärrman, Nowak, Wu, Dyck, Vaidyanathan, R, Mallet, Rudominer, Johnston, Mittal, Udathu, Christensen, Verma, Irving, Santucci, Elsayed, Davoodi, Georgiev, Tenney, Hua, Cideron, Leurent, Alnahlawi, Georgescu, Wei, Zheng, Scandinaro, Jiang, Snoek, Sundararajan, Wang, Ontiveros, Karo, Cole, Rajashekhar, Tumeh, Ben-David, Jain, Uesato, Datta, Bunyan, Wu, Zhang, Stanczyk, Zhang, Steiner, Naskar, Azzam, Johnson, Paszke, Chiu, Elias, Mohiuddin, Muhammad, Miao, Lee, Vieillard, Park, Zhang, Stanway, Garmon, Karmarkar, Dong, Lee, Kumar, Zhou, Evens, Isaac, Irving, Loper, Fink, Arkatkar, Chen, Shafran, Petrychenko, Chen, Jia, Levskaya, Zhu, Grabowski, Mao, Magni, Yao, Snaider, Casagrande, Palmer, Suganthan, Castaño, Giannoumis, Kim, Rybiński, Sreevatsa, Prendki, Soergel, Goedeckemeyer, Gierke, Jafari, Gaba, Wiesner, Wright, Wei, Vashisht, Kulizhskaya, Hoover, Le, Li, Iwuanyanwu,
  Liu, Ramirez, Khorlin, Cui, LIN, Wu, Aguilar, Pallo, Chakladar, Perng, Abellan, Zhang, Dasgupta, Kushman, Penchev, Repina, Wu, van~der Weide, Ponnapalli, Kaplan, Simsa, Li, Dousse, Yang, Piper, Ie, Pasumarthi, Lintz, Vijayakumar, Andor, Valenzuela, Lui, Paduraru, Peng, Lee, Zhang, Greene, Nguyen, Kurylowicz, Hardin, Dixon, Janzer, Choo, Feng, Zhang, Singhal, Du, McKinnon, Antropova, Bolukbasi, Keller, Reid, Finchelstein, Raad, Crocker, Hawkins, Dadashi, Gaffney, Franko, Bulanova, Leblond, Chung, Askham, Cobo, Xu, Fischer, Xu, Sorokin, Alberti, Lin, Evans, Dimitriev, Forbes, Banarse, Tung, Omernick, Bishop, Sterneck, Jain, Xia, Amid, Piccinno, Wang, Banzal, Mankowitz, Polozov, Krakovna, Brown, Bateni, Duan, Firoiu, Thotakuri, Natan, Geist, tan Girgin, Li, Ye, Roval, Tojo, Kwong, Lee-Thorp, Yew, Sinopalnikov, Ramos, Mellor, Sharma, Wu, Miller, Sonnerat, Vnukov, Greig, Beattie, Caveness, Bai, Eisenschlos, Korchemniy, Tsai, Jasarevic, Kong, Dao, Zheng, Liu, Yang, Zhu, Teh, Sanmiya, Gladchenko, Trdin, Toyama,
  Rosen, Tavakkol, Xue, Elkind, Woodman, Carpenter, Papamakarios, Kemp, Kafle, Grunina, Sinha, Talbert, Wu, Owusu-Afriyie, Du, Thornton, Pont-Tuset, Narayana, Li, Fatehi, Wieting, Ajmeri, Uria, Ko, Knight, Héliou, Niu, Gu, Pang, Li, Levine, Stolovich, Santamaria-Fernandez, Goenka, Yustalim, Strudel, Elqursh, Deck, Lee, Li, Levin, Hoffmann, Holtmann-Rice, Bachem, Arora, Koh, Yeganeh, Põder, Tariq, Sun, Ionita, Seyedhosseini, Tafti, Liu, Gulati, Liu, Ye, Chrzaszcz, Wang, Sethi, Li, Brown, Singh, Fan, Parisi, Stanton, Koverkathu, Choquette-Choo, Li, Lu, Ittycheriah, Shroff, Varadarajan, Bahargam, Willoughby, Gaddy, Desjardins, Cornero, Robenek, Mittal, Albrecht, Shenoy, Moiseev, Jacobsson, Ghaffarkhah, Rivière, Walton, Crepy, Parrish, Zhou, Farabet, Radebaugh, Srinivasan, van~der Salm, Fidjeland, Scellato, Latorre-Chimoto, Klimczak-Plucińska, Bridson, de~Cesare, Hudson, Mendolicchio, Walker, Morris, Mauger, Guseynov, Reid, Odoom, Loher, Cotruta, Yenugula, Grewe, Petrushkina, Duerig, Sanchez, Yadlowsky, Shen,
  Globerson, Webb, Dua, Li, Bhupatiraju, Hurt, Qureshi, Agarwal, Shani, Eyal, Khare, Belle, Wang, Tekur, Kale, Wei, Sang, Saeta, Liechty, Sun, Zhao, Lee, Nayak, Fritz, Vuyyuru, Aslanides, Vyas, Wicke, Ma, Eltyshev, Martin, Cate, Manyika, Amiri, Kim, Xiong, Kang, Luisier, Tripuraneni, Madras, Guo, Waters, Wang, Ainslie, Baldridge, Zhang, Pruthi, Bauer, Yang, Mansour, Gelman, Xu, Polovets, Liu, Cai, Chen, Sheng, Xue, Ozair, Angermueller, Li, Sinha, Wang, Wiesinger, Koukoumidis, Tian, Iyer, Gurumurthy, Goldenson, Shah, Blake, Yu, Urbanowicz, Palomaki, Fernando, Durden, Mehta, Momchev, Rahimtoroghi, Georgaki, Raul, Ruder, Redshaw, Lee, Zhou, Jalan, Li, Hechtman, Schuh, Nasr, Milan, Mikulik, Franco, Green, Nguyen, Kelley, Mahendru, Hu, Howland, Vargas, Hui, Bansal, Rao, Ghiya, Wang, Ye, Sarr, Preston, Elish, Li, Kaku, Gupta, Pasupat, Juan, Someswar, M., Chen, Amini, Fabrikant, Chu, Dong, Muthal, Buthpitiya, Jauhari, Hua, Khandelwal, Hitron, Ren, Rinaldi, Drath, Dabush, Jiang, Godhia, Sachs, Chen, Fan, Taitelbaum,
  Noga, Dai, Wang, Liang, Hamer, Ferng, Elkind, Atias, Lee, Listík, Carlen, van~de Kerkhof, Pikus, Zaher, Müller, Zykova, Stefanec, Gatsko, Hirnschall, Sethi, Xu, Ahuja, Tsai, Stefanoiu, Feng, Dhandhania, Katyal, Gupta, Parulekar, Pitta, Zhao, Bhatia, Bhavnani, Alhadlaq, Li, Danenberg, Tu, Pine, Filippova, Ghosh, Limonchik, Urala, Lanka, Clive, Sun, Li, Wu, Hongtongsak, Li, Thakkar, Omarov, Majmundar, Alverson, Kucharski, Patel, Jain, Zabelin, Pelagatti, Kohli, Kumar, Kim, Sankar, Shah, Ramachandruni, Zeng, Bariach, Weidinger, Vu, Andreev, He, Hui, Kashem, Subramanya, Hsiao, Hassabis, Kavukcuoglu, Sadovsky, Le, Strohman, Wu, Petrov, Dean, and Vinyals]{geminiteam2025geminifamilyhighlycapable}
{Gemini Team}, Rohan Anil, Sebastian Borgeaud, Jean-Baptiste Alayrac, Jiahui Yu, Radu Soricut, Johan Schalkwyk, Andrew~M. Dai, Anja Hauth, Katie Millican, David Silver, Melvin Johnson, Ioannis Antonoglou, Julian Schrittwieser, Amelia Glaese, Jilin Chen, Emily Pitler, Timothy Lillicrap, Angeliki Lazaridou, Orhan Firat, James Molloy, Michael Isard, Paul~R. Barham, Tom Hennigan, Benjamin Lee, Fabio Viola, Malcolm Reynolds, Yuanzhong Xu, Ryan Doherty, Eli Collins, Clemens Meyer, Eliza Rutherford, Erica Moreira, Kareem Ayoub, Megha Goel, Jack Krawczyk, Cosmo Du, Ed~Chi, Heng-Tze Cheng, Eric Ni, Purvi Shah, Patrick Kane, Betty Chan, Manaal Faruqui, Aliaksei Severyn, Hanzhao Lin, YaGuang Li, Yong Cheng, Abe Ittycheriah, Mahdis Mahdieh, Mia Chen, Pei Sun, Dustin Tran, Sumit Bagri, Balaji Lakshminarayanan, Jeremiah Liu, Andras Orban, Fabian Güra, Hao Zhou, Xinying Song, Aurelien Boffy, Harish Ganapathy, Steven Zheng, HyunJeong Choe, Ágoston Weisz, Tao Zhu, Yifeng Lu, Siddharth Gopal, Jarrod Kahn, Maciej Kula, Jeff
  Pitman, Rushin Shah, Emanuel Taropa, Majd~Al Merey, Martin Baeuml, Zhifeng Chen, Laurent~El Shafey, Yujing Zhang, Olcan Sercinoglu, George Tucker, Enrique Piqueras, Maxim Krikun, Iain Barr, Nikolay Savinov, Ivo Danihelka, Becca Roelofs, Anaïs White, Anders Andreassen, Tamara von Glehn, Lakshman Yagati, Mehran Kazemi, Lucas Gonzalez, Misha Khalman, Jakub Sygnowski, Alexandre Frechette, Charlotte Smith, Laura Culp, Lev Proleev, Yi~Luan, Xi~Chen, James Lottes, Nathan Schucher, Federico Lebron, Alban Rrustemi, Natalie Clay, Phil Crone, Tomas Kocisky, Jeffrey Zhao, Bartek Perz, Dian Yu, Heidi Howard, Adam Bloniarz, Jack~W. Rae, Han Lu, Laurent Sifre, Marcello Maggioni, Fred Alcober, Dan Garrette, Megan Barnes, Shantanu Thakoor, Jacob Austin, Gabriel Barth-Maron, William Wong, Rishabh Joshi, Rahma Chaabouni, Deeni Fatiha, Arun Ahuja, Gaurav~Singh Tomar, Evan Senter, Martin Chadwick, Ilya Kornakov, Nithya Attaluri, Iñaki Iturrate, Ruibo Liu, Yunxuan Li, Sarah Cogan, Jeremy Chen, Chao Jia, Chenjie Gu, Qiao Zhang,
  Jordan Grimstad, Ale~Jakse Hartman, Xavier Garcia, Thanumalayan~Sankaranarayana Pillai, Jacob Devlin, Michael Laskin, Diego de~Las~Casas, Dasha Valter, Connie Tao, Lorenzo Blanco, Adrià~Puigdomènech Badia, David Reitter, Mianna Chen, Jenny Brennan, Clara Rivera, Sergey Brin, Shariq Iqbal, Gabriela Surita, Jane Labanowski, Abhi Rao, Stephanie Winkler, Emilio Parisotto, Yiming Gu, Kate Olszewska, Ravi Addanki, Antoine Miech, Annie Louis, Denis Teplyashin, Geoff Brown, Elliot Catt, Jan Balaguer, Jackie Xiang, Pidong Wang, Zoe Ashwood, Anton Briukhov, Albert Webson, Sanjay Ganapathy, Smit Sanghavi, Ajay Kannan, Ming-Wei Chang, Axel Stjerngren, Josip Djolonga, Yuting Sun, Ankur Bapna, Matthew Aitchison, Pedram Pejman, Henryk Michalewski, Tianhe Yu, Cindy Wang, Juliette Love, Junwhan Ahn, Dawn Bloxwich, Kehang Han, Peter Humphreys, Thibault Sellam, James Bradbury, Varun Godbole, Sina Samangooei, Bogdan Damoc, Alex Kaskasoli, Sébastien M.~R. Arnold, Vijay Vasudevan, Shubham Agrawal, Jason Riesa, Dmitry
  Lepikhin, Richard Tanburn, Srivatsan Srinivasan, Hyeontaek Lim, Sarah Hodkinson, Pranav Shyam, Johan Ferret, Steven Hand, Ankush Garg, Tom~Le Paine, Jian Li, Yujia Li, Minh Giang, Alexander Neitz, Zaheer Abbas, Sarah York, Machel Reid, Elizabeth Cole, Aakanksha Chowdhery, Dipanjan Das, Dominika Rogozińska, Vitaliy Nikolaev, Pablo Sprechmann, Zachary Nado, Lukas Zilka, Flavien Prost, Luheng He, Marianne Monteiro, Gaurav Mishra, Chris Welty, Josh Newlan, Dawei Jia, Miltiadis Allamanis, Clara~Huiyi Hu, Raoul de~Liedekerke, Justin Gilmer, Carl Saroufim, Shruti Rijhwani, Shaobo Hou, Disha Shrivastava, Anirudh Baddepudi, Alex Goldin, Adnan Ozturel, Albin Cassirer, Yunhan Xu, Daniel Sohn, Devendra Sachan, Reinald~Kim Amplayo, Craig Swanson, Dessie Petrova, Shashi Narayan, Arthur Guez, Siddhartha Brahma, Jessica Landon, Miteyan Patel, Ruizhe Zhao, Kevin Villela, Luyu Wang, Wenhao Jia, Matthew Rahtz, Mai Giménez, Legg Yeung, James Keeling, Petko Georgiev, Diana Mincu, Boxi Wu, Salem Haykal, Rachel Saputro, Kiran
  Vodrahalli, James Qin, Zeynep Cankara, Abhanshu Sharma, Nick Fernando, Will Hawkins, Behnam Neyshabur, Solomon Kim, Adrian Hutter, Priyanka Agrawal, Alex Castro-Ros, George van~den Driessche, Tao Wang, Fan Yang, Shuo yiin Chang, Paul Komarek, Ross McIlroy, Mario Lučić, Guodong Zhang, Wael Farhan, Michael Sharman, Paul Natsev, Paul Michel, Yamini Bansal, Siyuan Qiao, Kris Cao, Siamak Shakeri, Christina Butterfield, Justin Chung, Paul~Kishan Rubenstein, Shivani Agrawal, Arthur Mensch, Kedar Soparkar, Karel Lenc, Timothy Chung, Aedan Pope, Loren Maggiore, Jackie Kay, Priya Jhakra, Shibo Wang, Joshua Maynez, Mary Phuong, Taylor Tobin, Andrea Tacchetti, Maja Trebacz, Kevin Robinson, Yash Katariya, Sebastian Riedel, Paige Bailey, Kefan Xiao, Nimesh Ghelani, Lora Aroyo, Ambrose Slone, Neil Houlsby, Xuehan Xiong, Zhen Yang, Elena Gribovskaya, Jonas Adler, Mateo Wirth, Lisa Lee, Music Li, Thais Kagohara, Jay Pavagadhi, Sophie Bridgers, Anna Bortsova, Sanjay Ghemawat, Zafarali Ahmed, Tianqi Liu, Richard Powell,
  Vijay Bolina, Mariko Iinuma, Polina Zablotskaia, James Besley, Da-Woon Chung, Timothy Dozat, Ramona Comanescu, Xiance Si, Jeremy Greer, Guolong Su, Martin Polacek, Raphaël~Lopez Kaufman, Simon Tokumine, Hexiang Hu, Elena Buchatskaya, Yingjie Miao, Mohamed Elhawaty, Aditya Siddhant, Nenad Tomasev, Jinwei Xing, Christina Greer, Helen Miller, Shereen Ashraf, Aurko Roy, Zizhao Zhang, Ada Ma, Angelos Filos, Milos Besta, Rory Blevins, Ted Klimenko, Chih-Kuan Yeh, Soravit Changpinyo, Jiaqi Mu, Oscar Chang, Mantas Pajarskas, Carrie Muir, Vered Cohen, Charline~Le Lan, Krishna Haridasan, Amit Marathe, Steven Hansen, Sholto Douglas, Rajkumar Samuel, Mingqiu Wang, Sophia Austin, Chang Lan, Jiepu Jiang, Justin Chiu, Jaime~Alonso Lorenzo, Lars~Lowe Sjösund, Sébastien Cevey, Zach Gleicher, Thi Avrahami, Anudhyan Boral, Hansa Srinivasan, Vittorio Selo, Rhys May, Konstantinos Aisopos, Léonard Hussenot, Livio~Baldini Soares, Kate Baumli, Michael~B. Chang, Adrià Recasens, Ben Caine, Alexander Pritzel, Filip Pavetic,
  Fabio Pardo, Anita Gergely, Justin Frye, Vinay Ramasesh, Dan Horgan, Kartikeya Badola, Nora Kassner, Subhrajit Roy, Ethan Dyer, Víctor~Campos Campos, Alex Tomala, Yunhao Tang, Dalia~El Badawy, Elspeth White, Basil Mustafa, Oran Lang, Abhishek Jindal, Sharad Vikram, Zhitao Gong, Sergi Caelles, Ross Hemsley, Gregory Thornton, Fangxiaoyu Feng, Wojciech Stokowiec, Ce~Zheng, Phoebe Thacker, Çağlar Ünlü, Zhishuai Zhang, Mohammad Saleh, James Svensson, Max Bileschi, Piyush Patil, Ankesh Anand, Roman Ring, Katerina Tsihlas, Arpi Vezer, Marco Selvi, Toby Shevlane, Mikel Rodriguez, Tom Kwiatkowski, Samira Daruki, Keran Rong, Allan Dafoe, Nicholas FitzGerald, Keren Gu-Lemberg, Mina Khan, Lisa~Anne Hendricks, Marie Pellat, Vladimir Feinberg, James Cobon-Kerr, Tara Sainath, Maribeth Rauh, Sayed~Hadi Hashemi, Richard Ives, Yana Hasson, Eric Noland, Yuan Cao, Nathan Byrd, Le~Hou, Qingze Wang, Thibault Sottiaux, Michela Paganini, Jean-Baptiste Lespiau, Alexandre Moufarek, Samer Hassan, Kaushik Shivakumar, Joost van
  Amersfoort, Amol Mandhane, Pratik Joshi, Anirudh Goyal, Matthew Tung, Andrew Brock, Hannah Sheahan, Vedant Misra, Cheng Li, Nemanja Rakićević, Mostafa Dehghani, Fangyu Liu, Sid Mittal, Junhyuk Oh, Seb Noury, Eren Sezener, Fantine Huot, Matthew Lamm, Nicola~De Cao, Charlie Chen, Sidharth Mudgal, Romina Stella, Kevin Brooks, Gautam Vasudevan, Chenxi Liu, Mainak Chain, Nivedita Melinkeri, Aaron Cohen, Venus Wang, Kristie Seymore, Sergey Zubkov, Rahul Goel, Summer Yue, Sai Krishnakumaran, Brian Albert, Nate Hurley, Motoki Sano, Anhad Mohananey, Jonah Joughin, Egor Filonov, Tomasz Kępa, Yomna Eldawy, Jiawern Lim, Rahul Rishi, Shirin Badiezadegan, Taylor Bos, Jerry Chang, Sanil Jain, Sri Gayatri~Sundara Padmanabhan, Subha Puttagunta, Kalpesh Krishna, Leslie Baker, Norbert Kalb, Vamsi Bedapudi, Adam Kurzrok, Shuntong Lei, Anthony Yu, Oren Litvin, Xiang Zhou, Zhichun Wu, Sam Sobell, Andrea Siciliano, Alan Papir, Robby Neale, Jonas Bragagnolo, Tej Toor, Tina Chen, Valentin Anklin, Feiran Wang, Richie Feng, Milad
  Gholami, Kevin Ling, Lijuan Liu, Jules Walter, Hamid Moghaddam, Arun Kishore, Jakub Adamek, Tyler Mercado, Jonathan Mallinson, Siddhinita Wandekar, Stephen Cagle, Eran Ofek, Guillermo Garrido, Clemens Lombriser, Maksim Mukha, Botu Sun, Hafeezul~Rahman Mohammad, Josip Matak, Yadi Qian, Vikas Peswani, Pawel Janus, Quan Yuan, Leif Schelin, Oana David, Ankur Garg, Yifan He, Oleksii Duzhyi, Anton Älgmyr, Timothée Lottaz, Qi~Li, Vikas Yadav, Luyao Xu, Alex Chinien, Rakesh Shivanna, Aleksandr Chuklin, Josie Li, Carrie Spadine, Travis Wolfe, Kareem Mohamed, Subhabrata Das, Zihang Dai, Kyle He, Daniel von Dincklage, Shyam Upadhyay, Akanksha Maurya, Luyan Chi, Sebastian Krause, Khalid Salama, Pam~G Rabinovitch, Pavan Kumar~Reddy M, Aarush Selvan, Mikhail Dektiarev, Golnaz Ghiasi, Erdem Guven, Himanshu Gupta, Boyi Liu, Deepak Sharma, Idan~Heimlich Shtacher, Shachi Paul, Oscar Akerlund, François-Xavier Aubet, Terry Huang, Chen Zhu, Eric Zhu, Elico Teixeira, Matthew Fritze, Francesco Bertolini, Liana-Eleonora
  Marinescu, Martin Bölle, Dominik Paulus, Khyatti Gupta, Tejasi Latkar, Max Chang, Jason Sanders, Roopa Wilson, Xuewei Wu, Yi-Xuan Tan, Lam~Nguyen Thiet, Tulsee Doshi, Sid Lall, Swaroop Mishra, Wanming Chen, Thang Luong, Seth Benjamin, Jasmine Lee, Ewa Andrejczuk, Dominik Rabiej, Vipul Ranjan, Krzysztof Styrc, Pengcheng Yin, Jon Simon, Malcolm~Rose Harriott, Mudit Bansal, Alexei Robsky, Geoff Bacon, David Greene, Daniil Mirylenka, Chen Zhou, Obaid Sarvana, Abhimanyu Goyal, Samuel Andermatt, Patrick Siegler, Ben Horn, Assaf Israel, Francesco Pongetti, Chih-Wei~"Louis" Chen, Marco Selvatici, Pedro Silva, Kathie Wang, Jackson Tolins, Kelvin Guu, Roey Yogev, Xiaochen Cai, Alessandro Agostini, Maulik Shah, Hung Nguyen, Noah~\'{O} Donnaile, Sébastien Pereira, Linda Friso, Adam Stambler, Adam Kurzrok, Chenkai Kuang, Yan Romanikhin, Mark Geller, ZJ~Yan, Kane Jang, Cheng-Chun Lee, Wojciech Fica, Eric Malmi, Qijun Tan, Dan Banica, Daniel Balle, Ryan Pham, Yanping Huang, Diana Avram, Hongzhi Shi, Jasjot Singh, Chris
  Hidey, Niharika Ahuja, Pranab Saxena, Dan Dooley, Srividya~Pranavi Potharaju, Eileen O'Neill, Anand Gokulchandran, Ryan Foley, Kai Zhao, Mike Dusenberry, Yuan Liu, Pulkit Mehta, Ragha Kotikalapudi, Chalence Safranek-Shrader, Andrew Goodman, Joshua Kessinger, Eran Globen, Prateek Kolhar, Chris Gorgolewski, Ali Ibrahim, Yang Song, Ali Eichenbaum, Thomas Brovelli, Sahitya Potluri, Preethi Lahoti, Cip Baetu, Ali Ghorbani, Charles Chen, Andy Crawford, Shalini Pal, Mukund Sridhar, Petru Gurita, Asier Mujika, Igor Petrovski, Pierre-Louis Cedoz, Chenmei Li, Shiyuan Chen, Niccolò~Dal Santo, Siddharth Goyal, Jitesh Punjabi, Karthik Kappaganthu, Chester Kwak, Pallavi LV, Sarmishta Velury, Himadri Choudhury, Jamie Hall, Premal Shah, Ricardo Figueira, Matt Thomas, Minjie Lu, Ting Zhou, Chintu Kumar, Thomas Jurdi, Sharat Chikkerur, Yenai Ma, Adams Yu, Soo Kwak, Victor Ähdel, Sujeevan Rajayogam, Travis Choma, Fei Liu, Aditya Barua, Colin Ji, Ji~Ho Park, Vincent Hellendoorn, Alex Bailey, Taylan Bilal, Huanjie Zhou,
  Mehrdad Khatir, Charles Sutton, Wojciech Rzadkowski, Fiona Macintosh, Roopali Vij, Konstantin Shagin, Paul Medina, Chen Liang, Jinjing Zhou, Pararth Shah, Yingying Bi, Attila Dankovics, Shipra Banga, Sabine Lehmann, Marissa Bredesen, Zifan Lin, John~Eric Hoffmann, Jonathan Lai, Raynald Chung, Kai Yang, Nihal Balani, Arthur Bražinskas, Andrei Sozanschi, Matthew Hayes, Héctor~Fernández Alcalde, Peter Makarov, Will Chen, Antonio Stella, Liselotte Snijders, Michael Mandl, Ante Kärrman, Paweł Nowak, Xinyi Wu, Alex Dyck, Krishnan Vaidyanathan, Raghavender R, Jessica Mallet, Mitch Rudominer, Eric Johnston, Sushil Mittal, Akhil Udathu, Janara Christensen, Vishal Verma, Zach Irving, Andreas Santucci, Gamaleldin Elsayed, Elnaz Davoodi, Marin Georgiev, Ian Tenney, Nan Hua, Geoffrey Cideron, Edouard Leurent, Mahmoud Alnahlawi, Ionut Georgescu, Nan Wei, Ivy Zheng, Dylan Scandinaro, Heinrich Jiang, Jasper Snoek, Mukund Sundararajan, Xuezhi Wang, Zack Ontiveros, Itay Karo, Jeremy Cole, Vinu Rajashekhar, Lara Tumeh,
  Eyal Ben-David, Rishub Jain, Jonathan Uesato, Romina Datta, Oskar Bunyan, Shimu Wu, John Zhang, Piotr Stanczyk, Ye~Zhang, David Steiner, Subhajit Naskar, Michael Azzam, Matthew Johnson, Adam Paszke, Chung-Cheng Chiu, Jaume~Sanchez Elias, Afroz Mohiuddin, Faizan Muhammad, Jin Miao, Andrew Lee, Nino Vieillard, Jane Park, Jiageng Zhang, Jeff Stanway, Drew Garmon, Abhijit Karmarkar, Zhe Dong, Jong Lee, Aviral Kumar, Luowei Zhou, Jonathan Evens, William Isaac, Geoffrey Irving, Edward Loper, Michael Fink, Isha Arkatkar, Nanxin Chen, Izhak Shafran, Ivan Petrychenko, Zhe Chen, Johnson Jia, Anselm Levskaya, Zhenkai Zhu, Peter Grabowski, Yu~Mao, Alberto Magni, Kaisheng Yao, Javier Snaider, Norman Casagrande, Evan Palmer, Paul Suganthan, Alfonso Castaño, Irene Giannoumis, Wooyeol Kim, Mikołaj Rybiński, Ashwin Sreevatsa, Jennifer Prendki, David Soergel, Adrian Goedeckemeyer, Willi Gierke, Mohsen Jafari, Meenu Gaba, Jeremy Wiesner, Diana~Gage Wright, Yawen Wei, Harsha Vashisht, Yana Kulizhskaya, Jay Hoover, Maigo Le,
  Lu~Li, Chimezie Iwuanyanwu, Lu~Liu, Kevin Ramirez, Andrey Khorlin, Albert Cui, Tian LIN, Marcus Wu, Ricardo Aguilar, Keith Pallo, Abhishek Chakladar, Ginger Perng, Elena~Allica Abellan, Mingyang Zhang, Ishita Dasgupta, Nate Kushman, Ivo Penchev, Alena Repina, Xihui Wu, Tom van~der Weide, Priya Ponnapalli, Caroline Kaplan, Jiri Simsa, Shuangfeng Li, Olivier Dousse, Fan Yang, Jeff Piper, Nathan Ie, Rama Pasumarthi, Nathan Lintz, Anitha Vijayakumar, Daniel Andor, Pedro Valenzuela, Minnie Lui, Cosmin Paduraru, Daiyi Peng, Katherine Lee, Shuyuan Zhang, Somer Greene, Duc~Dung Nguyen, Paula Kurylowicz, Cassidy Hardin, Lucas Dixon, Lili Janzer, Kiam Choo, Ziqiang Feng, Biao Zhang, Achintya Singhal, Dayou Du, Dan McKinnon, Natasha Antropova, Tolga Bolukbasi, Orgad Keller, David Reid, Daniel Finchelstein, Maria~Abi Raad, Remi Crocker, Peter Hawkins, Robert Dadashi, Colin Gaffney, Ken Franko, Anna Bulanova, Rémi Leblond, Shirley Chung, Harry Askham, Luis~C. Cobo, Kelvin Xu, Felix Fischer, Jun Xu, Christina Sorokin,
  Chris Alberti, Chu-Cheng Lin, Colin Evans, Alek Dimitriev, Hannah Forbes, Dylan Banarse, Zora Tung, Mark Omernick, Colton Bishop, Rachel Sterneck, Rohan Jain, Jiawei Xia, Ehsan Amid, Francesco Piccinno, Xingyu Wang, Praseem Banzal, Daniel~J. Mankowitz, Alex Polozov, Victoria Krakovna, Sasha Brown, MohammadHossein Bateni, Dennis Duan, Vlad Firoiu, Meghana Thotakuri, Tom Natan, Matthieu Geist, Ser tan Girgin, Hui Li, Jiayu Ye, Ofir Roval, Reiko Tojo, Michael Kwong, James Lee-Thorp, Christopher Yew, Danila Sinopalnikov, Sabela Ramos, John Mellor, Abhishek Sharma, Kathy Wu, David Miller, Nicolas Sonnerat, Denis Vnukov, Rory Greig, Jennifer Beattie, Emily Caveness, Libin Bai, Julian Eisenschlos, Alex Korchemniy, Tomy Tsai, Mimi Jasarevic, Weize Kong, Phuong Dao, Zeyu Zheng, Frederick Liu, Fan Yang, Rui Zhu, Tian~Huey Teh, Jason Sanmiya, Evgeny Gladchenko, Nejc Trdin, Daniel Toyama, Evan Rosen, Sasan Tavakkol, Linting Xue, Chen Elkind, Oliver Woodman, John Carpenter, George Papamakarios, Rupert Kemp, Sushant
  Kafle, Tanya Grunina, Rishika Sinha, Alice Talbert, Diane Wu, Denese Owusu-Afriyie, Cosmo Du, Chloe Thornton, Jordi Pont-Tuset, Pradyumna Narayana, Jing Li, Saaber Fatehi, John Wieting, Omar Ajmeri, Benigno Uria, Yeongil Ko, Laura Knight, Amélie Héliou, Ning Niu, Shane Gu, Chenxi Pang, Yeqing Li, Nir Levine, Ariel Stolovich, Rebeca Santamaria-Fernandez, Sonam Goenka, Wenny Yustalim, Robin Strudel, Ali Elqursh, Charlie Deck, Hyo Lee, Zonglin Li, Kyle Levin, Raphael Hoffmann, Dan Holtmann-Rice, Olivier Bachem, Sho Arora, Christy Koh, Soheil~Hassas Yeganeh, Siim Põder, Mukarram Tariq, Yanhua Sun, Lucian Ionita, Mojtaba Seyedhosseini, Pouya Tafti, Zhiyu Liu, Anmol Gulati, Jasmine Liu, Xinyu Ye, Bart Chrzaszcz, Lily Wang, Nikhil Sethi, Tianrun Li, Ben Brown, Shreya Singh, Wei Fan, Aaron Parisi, Joe Stanton, Vinod Koverkathu, Christopher~A. Choquette-Choo, Yunjie Li, TJ~Lu, Abe Ittycheriah, Prakash Shroff, Mani Varadarajan, Sanaz Bahargam, Rob Willoughby, David Gaddy, Guillaume Desjardins, Marco Cornero, Brona
  Robenek, Bhavishya Mittal, Ben Albrecht, Ashish Shenoy, Fedor Moiseev, Henrik Jacobsson, Alireza Ghaffarkhah, Morgane Rivière, Alanna Walton, Clément Crepy, Alicia Parrish, Zongwei Zhou, Clement Farabet, Carey Radebaugh, Praveen Srinivasan, Claudia van~der Salm, Andreas Fidjeland, Salvatore Scellato, Eri Latorre-Chimoto, Hanna Klimczak-Plucińska, David Bridson, Dario de~Cesare, Tom Hudson, Piermaria Mendolicchio, Lexi Walker, Alex Morris, Matthew Mauger, Alexey Guseynov, Alison Reid, Seth Odoom, Lucia Loher, Victor Cotruta, Madhavi Yenugula, Dominik Grewe, Anastasia Petrushkina, Tom Duerig, Antonio Sanchez, Steve Yadlowsky, Amy Shen, Amir Globerson, Lynette Webb, Sahil Dua, Dong Li, Surya Bhupatiraju, Dan Hurt, Haroon Qureshi, Ananth Agarwal, Tomer Shani, Matan Eyal, Anuj Khare, Shreyas~Rammohan Belle, Lei Wang, Chetan Tekur, Mihir~Sanjay Kale, Jinliang Wei, Ruoxin Sang, Brennan Saeta, Tyler Liechty, Yi~Sun, Yao Zhao, Stephan Lee, Pandu Nayak, Doug Fritz, Manish~Reddy Vuyyuru, John Aslanides, Nidhi Vyas,
  Martin Wicke, Xiao Ma, Evgenii Eltyshev, Nina Martin, Hardie Cate, James Manyika, Keyvan Amiri, Yelin Kim, Xi~Xiong, Kai Kang, Florian Luisier, Nilesh Tripuraneni, David Madras, Mandy Guo, Austin Waters, Oliver Wang, Joshua Ainslie, Jason Baldridge, Han Zhang, Garima Pruthi, Jakob Bauer, Feng Yang, Riham Mansour, Jason Gelman, Yang Xu, George Polovets, Ji~Liu, Honglong Cai, Warren Chen, XiangHai Sheng, Emily Xue, Sherjil Ozair, Christof Angermueller, Xiaowei Li, Anoop Sinha, Weiren Wang, Julia Wiesinger, Emmanouil Koukoumidis, Yuan Tian, Anand Iyer, Madhu Gurumurthy, Mark Goldenson, Parashar Shah, MK~Blake, Hongkun Yu, Anthony Urbanowicz, Jennimaria Palomaki, Chrisantha Fernando, Ken Durden, Harsh Mehta, Nikola Momchev, Elahe Rahimtoroghi, Maria Georgaki, Amit Raul, Sebastian Ruder, Morgan Redshaw, Jinhyuk Lee, Denny Zhou, Komal Jalan, Dinghua Li, Blake Hechtman, Parker Schuh, Milad Nasr, Kieran Milan, Vladimir Mikulik, Juliana Franco, Tim Green, Nam Nguyen, Joe Kelley, Aroma Mahendru, Andrea Hu, Joshua
  Howland, Ben Vargas, Jeffrey Hui, Kshitij Bansal, Vikram Rao, Rakesh Ghiya, Emma Wang, Ke~Ye, Jean~Michel Sarr, Melanie~Moranski Preston, Madeleine Elish, Steve Li, Aakash Kaku, Jigar Gupta, Ice Pasupat, Da-Cheng Juan, Milan Someswar, Tejvi M., Xinyun Chen, Aida Amini, Alex Fabrikant, Eric Chu, Xuanyi Dong, Amruta Muthal, Senaka Buthpitiya, Sarthak Jauhari, Nan Hua, Urvashi Khandelwal, Ayal Hitron, Jie Ren, Larissa Rinaldi, Shahar Drath, Avigail Dabush, Nan-Jiang Jiang, Harshal Godhia, Uli Sachs, Anthony Chen, Yicheng Fan, Hagai Taitelbaum, Hila Noga, Zhuyun Dai, James Wang, Chen Liang, Jenny Hamer, Chun-Sung Ferng, Chenel Elkind, Aviel Atias, Paulina Lee, Vít Listík, Mathias Carlen, Jan van~de Kerkhof, Marcin Pikus, Krunoslav Zaher, Paul Müller, Sasha Zykova, Richard Stefanec, Vitaly Gatsko, Christoph Hirnschall, Ashwin Sethi, Xingyu~Federico Xu, Chetan Ahuja, Beth Tsai, Anca Stefanoiu, Bo~Feng, Keshav Dhandhania, Manish Katyal, Akshay Gupta, Atharva Parulekar, Divya Pitta, Jing Zhao, Vivaan Bhatia,
  Yashodha Bhavnani, Omar Alhadlaq, Xiaolin Li, Peter Danenberg, Dennis Tu, Alex Pine, Vera Filippova, Abhipso Ghosh, Ben Limonchik, Bhargava Urala, Chaitanya~Krishna Lanka, Derik Clive, Yi~Sun, Edward Li, Hao Wu, Kevin Hongtongsak, Ianna Li, Kalind Thakkar, Kuanysh Omarov, Kushal Majmundar, Michael Alverson, Michael Kucharski, Mohak Patel, Mudit Jain, Maksim Zabelin, Paolo Pelagatti, Rohan Kohli, Saurabh Kumar, Joseph Kim, Swetha Sankar, Vineet Shah, Lakshmi Ramachandruni, Xiangkai Zeng, Ben Bariach, Laura Weidinger, Tu~Vu, Alek Andreev, Antoine He, Kevin Hui, Sheleem Kashem, Amar Subramanya, Sissie Hsiao, Demis Hassabis, Koray Kavukcuoglu, Adam Sadovsky, Quoc Le, Trevor Strohman, Yonghui Wu, Slav Petrov, Jeffrey Dean, and Oriol Vinyals.
\newblock Gemini: A family of highly capable multimodal models, 2025.
\newblock URL \url{https://arxiv.org/abs/2312.11805}.

\bibitem[{Gemma Team} et~al.(2025){Gemma Team}, Kamath, Ferret, Pathak, Vieillard, Merhej, Perrin, Matejovicova, Ramé, Rivière, Rouillard, Mesnard, Cideron, bastien Grill, Ramos, Yvinec, Casbon, Pot, Penchev, Liu, Visin, Kenealy, Beyer, Zhai, Tsitsulin, Busa-Fekete, Feng, Sachdeva, Coleman, Gao, Mustafa, Barr, Parisotto, Tian, Eyal, Cherry, Peter, Sinopalnikov, Bhupatiraju, Agarwal, Kazemi, Malkin, Kumar, Vilar, Brusilovsky, Luo, Steiner, Friesen, Sharma, Sharma, Gilady, Goedeckemeyer, Saade, Feng, Kolesnikov, Bendebury, Abdagic, Vadi, György, Pinto, Das, Bapna, Miech, Yang, Paterson, Shenoy, Chakrabarti, Piot, Wu, Shahriari, Petrini, Chen, Lan, Choquette-Choo, Carey, Brick, Deutsch, Eisenbud, Cattle, Cheng, Paparas, Sreepathihalli, Reid, Tran, Zelle, Noland, Huizenga, Kharitonov, Liu, Amirkhanyan, Cameron, Hashemi, Klimczak-Plucińska, Singh, Mehta, Lehri, Hazimeh, Ballantyne, Szpektor, Nardini, Pouget-Abadie, Chan, Stanton, Wieting, Lai, Orbay, Fernandez, Newlan, yeong Ji, Singh, Black, Yu, Hui,
  Vodrahalli, Greff, Qiu, Valentine, Coelho, Ritter, Hoffman, Watson, Chaturvedi, Moynihan, Ma, Babar, Noy, Byrd, Roy, Momchev, Chauhan, Sachdeva, Bunyan, Botarda, Caron, Rubenstein, Culliton, Schmid, Sessa, Xu, Stanczyk, Tafti, Shivanna, Wu, Pan, Rokni, Willoughby, Vallu, Mullins, Jerome, Smoot, Girgin, Iqbal, Reddy, Sheth, Põder, Bhatnagar, Panyam, Eiger, Zhang, Liu, Yacovone, Liechty, Kalra, Evci, Misra, Roseberry, Feinberg, Kolesnikov, Han, Kwon, Chen, Chow, Zhu, Wei, Egyed, Cotruta, Giang, Kirk, Rao, Black, Babar, Lo, Moreira, Martins, Sanseviero, Gonzalez, Gleicher, Warkentin, Mirrokni, Senter, Collins, Barral, Ghahramani, Hadsell, Matias, Sculley, Petrov, Fiedel, Shazeer, Vinyals, Dean, Hassabis, Kavukcuoglu, Farabet, Buchatskaya, Alayrac, Anil, Dmitry, Lepikhin, Borgeaud, Bachem, Joulin, Andreev, Hardin, Dadashi, and Hussenot]{gemmateam2025gemma3technicalreport}
{Gemma Team}, Aishwarya Kamath, Johan Ferret, Shreya Pathak, Nino Vieillard, Ramona Merhej, Sarah Perrin, Tatiana Matejovicova, Alexandre Ramé, Morgane Rivière, Louis Rouillard, Thomas Mesnard, Geoffrey Cideron, Jean bastien Grill, Sabela Ramos, Edouard Yvinec, Michelle Casbon, Etienne Pot, Ivo Penchev, Gaël Liu, Francesco Visin, Kathleen Kenealy, Lucas Beyer, Xiaohai Zhai, Anton Tsitsulin, Robert Busa-Fekete, Alex Feng, Noveen Sachdeva, Benjamin Coleman, Yi~Gao, Basil Mustafa, Iain Barr, Emilio Parisotto, David Tian, Matan Eyal, Colin Cherry, Jan-Thorsten Peter, Danila Sinopalnikov, Surya Bhupatiraju, Rishabh Agarwal, Mehran Kazemi, Dan Malkin, Ravin Kumar, David Vilar, Idan Brusilovsky, Jiaming Luo, Andreas Steiner, Abe Friesen, Abhanshu Sharma, Abheesht Sharma, Adi~Mayrav Gilady, Adrian Goedeckemeyer, Alaa Saade, Alex Feng, Alexander Kolesnikov, Alexei Bendebury, Alvin Abdagic, Amit Vadi, András György, André~Susano Pinto, Anil Das, Ankur Bapna, Antoine Miech, Antoine Yang, Antonia Paterson, Ashish
  Shenoy, Ayan Chakrabarti, Bilal Piot, Bo~Wu, Bobak Shahriari, Bryce Petrini, Charlie Chen, Charline~Le Lan, Christopher~A. Choquette-Choo, CJ~Carey, Cormac Brick, Daniel Deutsch, Danielle Eisenbud, Dee Cattle, Derek Cheng, Dimitris Paparas, Divyashree~Shivakumar Sreepathihalli, Doug Reid, Dustin Tran, Dustin Zelle, Eric Noland, Erwin Huizenga, Eugene Kharitonov, Frederick Liu, Gagik Amirkhanyan, Glenn Cameron, Hadi Hashemi, Hanna Klimczak-Plucińska, Harman Singh, Harsh Mehta, Harshal~Tushar Lehri, Hussein Hazimeh, Ian Ballantyne, Idan Szpektor, Ivan Nardini, Jean Pouget-Abadie, Jetha Chan, Joe Stanton, John Wieting, Jonathan Lai, Jordi Orbay, Joseph Fernandez, Josh Newlan, Ju~yeong Ji, Jyotinder Singh, Kat Black, Kathy Yu, Kevin Hui, Kiran Vodrahalli, Klaus Greff, Linhai Qiu, Marcella Valentine, Marina Coelho, Marvin Ritter, Matt Hoffman, Matthew Watson, Mayank Chaturvedi, Michael Moynihan, Min Ma, Nabila Babar, Natasha Noy, Nathan Byrd, Nick Roy, Nikola Momchev, Nilay Chauhan, Noveen Sachdeva, Oskar
  Bunyan, Pankil Botarda, Paul Caron, Paul~Kishan Rubenstein, Phil Culliton, Philipp Schmid, Pier~Giuseppe Sessa, Pingmei Xu, Piotr Stanczyk, Pouya Tafti, Rakesh Shivanna, Renjie Wu, Renke Pan, Reza Rokni, Rob Willoughby, Rohith Vallu, Ryan Mullins, Sammy Jerome, Sara Smoot, Sertan Girgin, Shariq Iqbal, Shashir Reddy, Shruti Sheth, Siim Põder, Sijal Bhatnagar, Sindhu~Raghuram Panyam, Sivan Eiger, Susan Zhang, Tianqi Liu, Trevor Yacovone, Tyler Liechty, Uday Kalra, Utku Evci, Vedant Misra, Vincent Roseberry, Vlad Feinberg, Vlad Kolesnikov, Woohyun Han, Woosuk Kwon, Xi~Chen, Yinlam Chow, Yuvein Zhu, Zichuan Wei, Zoltan Egyed, Victor Cotruta, Minh Giang, Phoebe Kirk, Anand Rao, Kat Black, Nabila Babar, Jessica Lo, Erica Moreira, Luiz~Gustavo Martins, Omar Sanseviero, Lucas Gonzalez, Zach Gleicher, Tris Warkentin, Vahab Mirrokni, Evan Senter, Eli Collins, Joelle Barral, Zoubin Ghahramani, Raia Hadsell, Yossi Matias, D.~Sculley, Slav Petrov, Noah Fiedel, Noam Shazeer, Oriol Vinyals, Jeff Dean, Demis Hassabis,
  Koray Kavukcuoglu, Clement Farabet, Elena Buchatskaya, Jean-Baptiste Alayrac, Rohan Anil, Dmitry, Lepikhin, Sebastian Borgeaud, Olivier Bachem, Armand Joulin, Alek Andreev, Cassidy Hardin, Robert Dadashi, and Léonard Hussenot.
\newblock Gemma 3 technical report, 2025.
\newblock URL \url{https://arxiv.org/abs/2503.19786}.

\bibitem[Ghosh et~al.(2025)Ghosh, Datta, Saha, and Agarwal]{ghosh-etal-2025-survey}
Akash Ghosh, Debayan Datta, Sriparna Saha, and Chirag Agarwal.
\newblock A survey of multilingual reasoning in language models.
\newblock In Christos Christodoulopoulos, Tanmoy Chakraborty, Carolyn Rose, and Violet Peng (eds.), \emph{Findings of the Association for Computational Linguistics: EMNLP 2025}, pp.\  8920--8936, Suzhou, China, November 2025. Association for Computational Linguistics.
\newblock ISBN 979-8-89176-335-7.
\newblock \doi{10.18653/v1/2025.findings-emnlp.474}.
\newblock URL \url{https://aclanthology.org/2025.findings-emnlp.474/}.

\bibitem[Goh \& Barabási(2008)Goh and Barabási]{Goh_2008}
K.-I. Goh and A.-L. Barabási.
\newblock Burstiness and memory in complex systems.
\newblock \emph{Europhysics Letters}, 81\penalty0 (4):\penalty0 48002, jan 2008.
\newblock \doi{10.1209/0295-5075/81/48002}.
\newblock URL \url{https://doi.org/10.1209/0295-5075/81/48002}.

\bibitem[Google(2025{\natexlab{a}})]{gemini}
Google.
\newblock Gemini models, 2025{\natexlab{a}}.
\newblock URL \url{https://ai.google.dev/gemini-api/docs/models/}.

\bibitem[Google(2025{\natexlab{b}})]{gemma-3-4b-it}
Google.
\newblock google/gemma-3-4b-it, 2025{\natexlab{b}}.
\newblock URL \url{https://huggingface.co/google/gemma-3-4b-it}.

\bibitem[Grootendorst(2022)]{grootendorst2022bertopicneuraltopicmodeling}
Maarten Grootendorst.
\newblock Bertopic: Neural topic modeling with a class-based tf-idf procedure, 2022.
\newblock URL \url{https://arxiv.org/abs/2203.05794}.

\bibitem[Gumperz(1977)]{doi:10.1177/003368827700800201}
John~J. Gumperz.
\newblock The sociolinguistic significance of conversational code-switching.
\newblock \emph{RELC Journal}, 8\penalty0 (2):\penalty0 1--34, 1977.
\newblock \doi{10.1177/003368827700800201}.
\newblock URL \url{https://doi.org/10.1177/003368827700800201}.

\bibitem[Guo et~al.(2025)Guo, Yang, Zhang, Song, Wang, Zhu, Xu, Zhang, Ma, Bi, Zhang, Yu, Wu, Wu, Gou, Shao, Li, Gao, Liu, Xue, Wang, Wu, Feng, Lu, Zhao, Deng, Ruan, Dai, Chen, Ji, Li, Lin, Dai, Luo, Hao, Chen, Li, Zhang, Xu, Ding, Gao, Qu, Li, Guo, Li, Chen, Yuan, Tu, Qiu, Li, Cai, Ni, Liang, Chen, Dong, Hu, You, Gao, Guan, Huang, Yu, Wang, Zhang, Zhao, Wang, Zhang, Xu, Xia, Zhang, Zhang, Tang, Zhou, Li, Wang, Li, Tian, Huang, Zhang, Wang, Chen, Du, Ge, Zhang, Pan, Wang, Chen, Jin, Chen, Lu, Zhou, Chen, Ye, Wang, Yu, Zhou, Pan, Li, Zhou, Wu, Yun, Pei, Sun, Wang, Zeng, Liu, Liang, Gao, Yu, Zhang, Xiao, An, Liu, Wang, Chen, Nie, Cheng, Liu, Xie, Liu, Yang, Li, Su, Lin, Li, Jin, Shen, Chen, Sun, Wang, Song, Zhou, Wang, Shan, Li, Wang, Wei, Zhang, Xu, Li, Zhao, Sun, Wang, Yu, Zhang, Shi, Xiong, He, Piao, Wang, Tan, Ma, Liu, Guo, Ou, Wang, Gong, Zou, He, Xiong, Luo, You, Liu, Zhou, Zhu, Huang, Li, Zheng, Zhu, Ma, Tang, Zha, Yan, Ren, Ren, Sha, Fu, Xu, Xie, Zhang, Hao, Ma, Yan, Wu, Gu, Zhu, Liu, Li, Xie, Song,
  Pan, Huang, Xu, Zhang, and Zhang]{Guo_2025}
Daya Guo, Dejian Yang, Haowei Zhang, Junxiao Song, Peiyi Wang, Qihao Zhu, Runxin Xu, Ruoyu Zhang, Shirong Ma, Xiao Bi, Xiaokang Zhang, Xingkai Yu, Yu~Wu, Z.~F. Wu, Zhibin Gou, Zhihong Shao, Zhuoshu Li, Ziyi Gao, Aixin Liu, Bing Xue, Bingxuan Wang, Bochao Wu, Bei Feng, Chengda Lu, Chenggang Zhao, Chengqi Deng, Chong Ruan, Damai Dai, Deli Chen, Dongjie Ji, Erhang Li, Fangyun Lin, Fucong Dai, Fuli Luo, Guangbo Hao, Guanting Chen, Guowei Li, H.~Zhang, Hanwei Xu, Honghui Ding, Huazuo Gao, Hui Qu, Hui Li, Jianzhong Guo, Jiashi Li, Jingchang Chen, Jingyang Yuan, Jinhao Tu, Junjie Qiu, Junlong Li, J.~L. Cai, Jiaqi Ni, Jian Liang, Jin Chen, Kai Dong, Kai Hu, Kaichao You, Kaige Gao, Kang Guan, Kexin Huang, Kuai Yu, Lean Wang, Lecong Zhang, Liang Zhao, Litong Wang, Liyue Zhang, Lei Xu, Leyi Xia, Mingchuan Zhang, Minghua Zhang, Minghui Tang, Mingxu Zhou, Meng Li, Miaojun Wang, Mingming Li, Ning Tian, Panpan Huang, Peng Zhang, Qiancheng Wang, Qinyu Chen, Qiushi Du, Ruiqi Ge, Ruisong Zhang, Ruizhe Pan, Runji Wang, R.~J.
  Chen, R.~L. Jin, Ruyi Chen, Shanghao Lu, Shangyan Zhou, Shanhuang Chen, Shengfeng Ye, Shiyu Wang, Shuiping Yu, Shunfeng Zhou, Shuting Pan, S.~S. Li, Shuang Zhou, Shaoqing Wu, Tao Yun, Tian Pei, Tianyu Sun, T.~Wang, Wangding Zeng, Wen Liu, Wenfeng Liang, Wenjun Gao, Wenqin Yu, Wentao Zhang, W.~L. Xiao, Wei An, Xiaodong Liu, Xiaohan Wang, Xiaokang Chen, Xiaotao Nie, Xin Cheng, Xin Liu, Xin Xie, Xingchao Liu, Xinyu Yang, Xinyuan Li, Xuecheng Su, Xuheng Lin, X.~Q. Li, Xiangyue Jin, Xiaojin Shen, Xiaosha Chen, Xiaowen Sun, Xiaoxiang Wang, Xinnan Song, Xinyi Zhou, Xianzu Wang, Xinxia Shan, Y.~K. Li, Y.~Q. Wang, Y.~X. Wei, Yang Zhang, Yanhong Xu, Yao Li, Yao Zhao, Yaofeng Sun, Yaohui Wang, Yi~Yu, Yichao Zhang, Yifan Shi, Yiliang Xiong, Ying He, Yishi Piao, Yisong Wang, Yixuan Tan, Yiyang Ma, Yiyuan Liu, Yongqiang Guo, Yuan Ou, Yuduan Wang, Yue Gong, Yuheng Zou, Yujia He, Yunfan Xiong, Yuxiang Luo, Yuxiang You, Yuxuan Liu, Yuyang Zhou, Y.~X. Zhu, Yanping Huang, Yaohui Li, Yi~Zheng, Yuchen Zhu, Yunxian Ma, Ying
  Tang, Yukun Zha, Yuting Yan, Z.~Z. Ren, Zehui Ren, Zhangli Sha, Zhe Fu, Zhean Xu, Zhenda Xie, Zhengyan Zhang, Zhewen Hao, Zhicheng Ma, Zhigang Yan, Zhiyu Wu, Zihui Gu, Zijia Zhu, Zijun Liu, Zilin Li, Ziwei Xie, Ziyang Song, Zizheng Pan, Zhen Huang, Zhipeng Xu, Zhongyu Zhang, and Zhen Zhang.
\newblock Deepseek-r1 incentivizes reasoning in llms through reinforcement learning.
\newblock \emph{Nature}, 645\penalty0 (8081):\penalty0 633–638, September 2025.
\newblock ISSN 1476-4687.
\newblock \doi{10.1038/s41586-025-09422-z}.
\newblock URL \url{http://dx.doi.org/10.1038/s41586-025-09422-z}.

\bibitem[Guzmán et~al.(2017)Guzmán, Ricard, Serigos, Bullock, and Toribio]{guzman17_interspeech}
Gualberto Guzmán, Joseph Ricard, Jacqueline Serigos, Barbara~E. Bullock, and Almeida~Jacqueline Toribio.
\newblock {Metrics for Modeling Code-Switching Across Corpora}.
\newblock In \emph{{Interspeech 2017}}, pp.\  67--71, 2017.
\newblock \doi{10.21437/Interspeech.2017-1429}.

\bibitem[Haas et~al.(2025)Haas, Yona, D'Antonio, Goldshtein, and Das]{haas2025simpleqaverifiedreliablefactuality}
Lukas Haas, Gal Yona, Giovanni D'Antonio, Sasha Goldshtein, and Dipanjan Das.
\newblock Simpleqa verified: A reliable factuality benchmark to measure parametric knowledge, 2025.
\newblock URL \url{https://arxiv.org/abs/2509.07968}.

\bibitem[Hammarström et~al.(2025)Hammarström, Forkel, Haspelmath, and Bank]{glottolog}
Harald Hammarström, Robert Forkel, Martin Haspelmath, and Sebastian Bank.
\newblock Glottolog 5.2, 2025.
\newblock URL \url{https://doi.org/10.5281/zenodo.15525265}.

\bibitem[Hendrycks et~al.(2021)Hendrycks, Burns, Basart, Zou, Mazeika, Song, and Steinhardt]{hendrycks2021measuring}
Dan Hendrycks, Collin Burns, Steven Basart, Andy Zou, Mantas Mazeika, Dawn Song, and Jacob Steinhardt.
\newblock Measuring massive multitask language understanding.
\newblock In \emph{International Conference on Learning Representations}, 2021.
\newblock URL \url{https://openreview.net/forum?id=d7KBjmI3GmQ}.

\bibitem[Kargaran et~al.(2024)Kargaran, Yvon, and Sch{\"u}tze]{kargaran-etal-2024-glotscript}
Amir~Hossein Kargaran, Fran{\c{c}}ois Yvon, and Hinrich Sch{\"u}tze.
\newblock {G}lot{S}cript: A resource and tool for low resource writing system identification.
\newblock In Nicoletta Calzolari, Min-Yen Kan, Veronique Hoste, Alessandro Lenci, Sakriani Sakti, and Nianwen Xue (eds.), \emph{Proceedings of the 2024 Joint International Conference on Computational Linguistics, Language Resources and Evaluation (LREC-COLING 2024)}, pp.\  7774--7784, Torino, Italia, May 2024. ELRA and ICCL.
\newblock URL \url{https://aclanthology.org/2024.lrec-main.687/}.

\bibitem[Kazemi et~al.(2025)Kazemi, Fatemi, Bansal, Palowitch, Anastasiou, Mehta, Jain, Aglietti, Jindal, Chen, Dikkala, Tyen, Liu, Shalit, Chiappa, Olszewska, Tay, Tran, Le, and Firat]{kazemi2025bigbenchextrahard}
Mehran Kazemi, Bahare Fatemi, Hritik Bansal, John Palowitch, Chrysovalantis Anastasiou, Sanket~Vaibhav Mehta, Lalit~K. Jain, Virginia Aglietti, Disha Jindal, Peter Chen, Nishanth Dikkala, Gladys Tyen, Xin Liu, Uri Shalit, Silvia Chiappa, Kate Olszewska, Yi~Tay, Vinh~Q. Tran, Quoc~V. Le, and Orhan Firat.
\newblock Big-bench extra hard, 2025.
\newblock URL \url{https://arxiv.org/abs/2502.19187}.

\bibitem[Kumar \& Jurgens(2025)Kumar and Jurgens]{kumar2025rulesmeantbrokenunderstanding}
Shivani Kumar and David Jurgens.
\newblock Are rules meant to be broken? understanding multilingual moral reasoning as a computational pipeline with {U}ni{M}oral.
\newblock In Wanxiang Che, Joyce Nabende, Ekaterina Shutova, and Mohammad~Taher Pilehvar (eds.), \emph{Proceedings of the 63rd Annual Meeting of the Association for Computational Linguistics (Volume 1: Long Papers)}, pp.\  5890--5912, Vienna, Austria, July 2025. Association for Computational Linguistics.
\newblock ISBN 979-8-89176-251-0.
\newblock \doi{10.18653/v1/2025.acl-long.294}.
\newblock URL \url{https://aclanthology.org/2025.acl-long.294/}.

\bibitem[Kuwanto et~al.(2024)Kuwanto, Agarwal, Winata, and Wijaya]{kuwanto2024linguisticstheorymeetsllm}
Garry Kuwanto, Chaitanya Agarwal, Genta~Indra Winata, and Derry~Tanti Wijaya.
\newblock Linguistics theory meets llm: Code-switched text generation via equivalence constrained large language models, 2024.
\newblock URL \url{https://arxiv.org/abs/2410.22660}.

\bibitem[Lee et~al.(2025)Lee, Kim, Seo, Jo, Go, Hwang, Park, Yue, Welleck, Neubig, Lee, and Seo]{lee2025cotencyclopediaanalyzingpredicting}
Seongyun Lee, Seungone Kim, Minju Seo, Yongrae Jo, Dongyoung Go, Hyeonbin Hwang, Jinho Park, Xiang Yue, Sean Welleck, Graham Neubig, Moontae Lee, and Minjoon Seo.
\newblock The cot encyclopedia: Analyzing, predicting, and controlling how a reasoning model will think, 2025.
\newblock URL \url{https://arxiv.org/abs/2505.10185}.

\bibitem[Li et~al.(2025{\natexlab{a}})Li, Xin, Miao, Long, and Ungar]{li-etal-2025-impact}
Yihao Li, Jiayi Xin, Miranda~Muqing Miao, Qi~Long, and Lyle Ungar.
\newblock The impact of language mixing on bilingual {LLM} reasoning.
\newblock In Christos Christodoulopoulos, Tanmoy Chakraborty, Carolyn Rose, and Violet Peng (eds.), \emph{Proceedings of the 2025 Conference on Empirical Methods in Natural Language Processing}, pp.\  32531--32548, Suzhou, China, November 2025{\natexlab{a}}. Association for Computational Linguistics.
\newblock ISBN 979-8-89176-332-6.
\newblock \doi{10.18653/v1/2025.emnlp-main.1654}.
\newblock URL \url{https://aclanthology.org/2025.emnlp-main.1654/}.

\bibitem[Li et~al.(2025{\natexlab{b}})Li, Xin, Miao, Long, and Ungar]{li2025impactlanguagemixingbilingual}
Yihao Li, Jiayi Xin, Miranda~Muqing Miao, Qi~Long, and Lyle Ungar.
\newblock The impact of language mixing on bilingual llm reasoning, 2025{\natexlab{b}}.
\newblock URL \url{https://arxiv.org/abs/2507.15849}.

\bibitem[Marchisio et~al.(2024)Marchisio, Ko, Berard, Dehaze, and Ruder]{marchisio-etal-2024-understanding}
Kelly Marchisio, Wei-Yin Ko, Alexandre Berard, Th{\'e}o Dehaze, and Sebastian Ruder.
\newblock Understanding and mitigating language confusion in {LLM}s.
\newblock In Yaser Al-Onaizan, Mohit Bansal, and Yun-Nung Chen (eds.), \emph{Proceedings of the 2024 Conference on Empirical Methods in Natural Language Processing}, pp.\  6653--6677, Miami, Florida, USA, November 2024. Association for Computational Linguistics.
\newblock \doi{10.18653/v1/2024.emnlp-main.380}.
\newblock URL \url{https://aclanthology.org/2024.emnlp-main.380/}.

\bibitem[Mistral-AI et~al.(2025)Mistral-AI, Rastogi, Jiang, Lo, Berrada, Lample, Rute, Barmentlo, Yadav, Khandelwal, Chandu, Blier, Saulnier, Dinot, Darrin, Gupta, Soletskyi, Vaze, Scao, Wang, Yang, Liu, Sablayrolles, Héliou, Martin, Ehrenberg, Agarwal, Roux, Darcet, Mensch, Bout, Rozière, Monicault, Bamford, Wallenwein, Renaudin, Lanfranchi, Dabert, Mizelle, de~las Casas, Chane-Sane, Fugier, Hanna, Delerce, Guinet, Novikov, Martin, Jaju, Ludziejewski, Chabran, Delignon, Studnia, Amar, Roberts, Denize, Saxena, Jain, Zhao, Martin, Gao, Lavaud, Pellat, Guillaumin, Felardos, Augustin, Seznec, Raghuraman, Duchenne, Wang, von Platen, Saffer, Jacob, Wambergue, Kurylowicz, Muddireddy, Chagniot, Stock, Agrawal, Sauvestre, Delacourt, Gandhi, Subramanian, Dalal, Gandhi, Ghosh, Mishra, Aithal, Antoniak, Schueller, Lavril, Robert, Wang, Lacroix, Nemychnikova, Paltz, Richard, Li, Marshall, Zhang, and Tang]{mistralai2025magistral}
Mistral-AI, Abhinav Rastogi, Albert~Q. Jiang, Andy Lo, Gabrielle Berrada, Guillaume Lample, Jason Rute, Joep Barmentlo, Karmesh Yadav, Kartik Khandelwal, Khyathi~Raghavi Chandu, Léonard Blier, Lucile Saulnier, Matthieu Dinot, Maxime Darrin, Neha Gupta, Roman Soletskyi, Sagar Vaze, Teven~Le Scao, Yihan Wang, Adam Yang, Alexander~H. Liu, Alexandre Sablayrolles, Amélie Héliou, Amélie Martin, Andy Ehrenberg, Anmol Agarwal, Antoine Roux, Arthur Darcet, Arthur Mensch, Baptiste Bout, Baptiste Rozière, Baudouin~De Monicault, Chris Bamford, Christian Wallenwein, Christophe Renaudin, Clémence Lanfranchi, Darius Dabert, Devon Mizelle, Diego de~las Casas, Elliot Chane-Sane, Emilien Fugier, Emma~Bou Hanna, Gauthier Delerce, Gauthier Guinet, Georgii Novikov, Guillaume Martin, Himanshu Jaju, Jan Ludziejewski, Jean-Hadrien Chabran, Jean-Malo Delignon, Joachim Studnia, Jonas Amar, Josselin~Somerville Roberts, Julien Denize, Karan Saxena, Kush Jain, Lingxiao Zhao, Louis Martin, Luyu Gao, Lélio~Renard Lavaud, Marie
  Pellat, Mathilde Guillaumin, Mathis Felardos, Maximilian Augustin, Mickaël Seznec, Nikhil Raghuraman, Olivier Duchenne, Patricia Wang, Patrick von Platen, Patryk Saffer, Paul Jacob, Paul Wambergue, Paula Kurylowicz, Pavankumar~Reddy Muddireddy, Philomène Chagniot, Pierre Stock, Pravesh Agrawal, Romain Sauvestre, Rémi Delacourt, Sanchit Gandhi, Sandeep Subramanian, Shashwat Dalal, Siddharth Gandhi, Soham Ghosh, Srijan Mishra, Sumukh Aithal, Szymon Antoniak, Thibault Schueller, Thibaut Lavril, Thomas Robert, Thomas Wang, Timothée Lacroix, Valeriia Nemychnikova, Victor Paltz, Virgile Richard, Wen-Ding Li, William Marshall, Xuanyu Zhang, and Yunhao Tang.
\newblock Magistral, 2025.
\newblock URL \url{https://arxiv.org/abs/2506.10910}.

\bibitem[Moshkov et~al.(2025)Moshkov, Hanley, Sorokin, Toshniwal, Henkel, Schifferer, Du, and Gitman]{moshkov2025aimo2winningsolutionbuilding}
Ivan Moshkov, Darragh Hanley, Ivan Sorokin, Shubham Toshniwal, Christof Henkel, Benedikt Schifferer, Wei Du, and Igor Gitman.
\newblock Aimo-2 winning solution: Building state-of-the-art mathematical reasoning models with openmathreasoning dataset, 2025.
\newblock URL \url{https://arxiv.org/abs/2504.16891}.

\bibitem[Muennighoff et~al.(2025{\natexlab{a}})Muennighoff, Yang, Shi, Li, Fei-Fei, Hajishirzi, Zettlemoyer, Liang, Candes, and Hashimoto]{muennighoff-etal-2025-s1}
Niklas Muennighoff, Zitong Yang, Weijia Shi, Xiang~Lisa Li, Li~Fei-Fei, Hannaneh Hajishirzi, Luke Zettlemoyer, Percy Liang, Emmanuel Candes, and Tatsunori Hashimoto.
\newblock s1: Simple test-time scaling.
\newblock In Christos Christodoulopoulos, Tanmoy Chakraborty, Carolyn Rose, and Violet Peng (eds.), \emph{Proceedings of the 2025 Conference on Empirical Methods in Natural Language Processing}, pp.\  20275--20321, Suzhou, China, November 2025{\natexlab{a}}. Association for Computational Linguistics.
\newblock ISBN 979-8-89176-332-6.
\newblock \doi{10.18653/v1/2025.emnlp-main.1025}.
\newblock URL \url{https://aclanthology.org/2025.emnlp-main.1025/}.

\bibitem[Muennighoff et~al.(2025{\natexlab{b}})Muennighoff, Yang, Shi, Li, Fei-Fei, Hajishirzi, Zettlemoyer, Liang, Candes, and Hashimoto]{muennighoff2025s}
Niklas Muennighoff, Zitong Yang, Weijia Shi, Xiang~Lisa Li, Li~Fei-Fei, Hannaneh Hajishirzi, Luke Zettlemoyer, Percy Liang, Emmanuel Candes, and Tatsunori Hashimoto.
\newblock s1: Simple test-time scaling.
\newblock In \emph{Workshop on Reasoning and Planning for Large Language Models}, 2025{\natexlab{b}}.
\newblock URL \url{https://openreview.net/forum?id=LdH0vrgAHm}.

\bibitem[Myers-Scotton(2001)]{myers2001matrix}
Carol Myers-Scotton.
\newblock \emph{Codeswitching Worldwide II}, chapter The matrix language frame model: Developments and responses, pp.\  23--58.
\newblock Mouton de Gruyter, Berlin, 2001.

\bibitem[Myers-Scotton(2017)]{doi:https://doi.org/10.1002/9781405166256.ch13}
Carol Myers-Scotton.
\newblock \emph{Code-Switching}, chapter~13, pp.\  217--237.
\newblock John Wiley \& Sons, Ltd, 2017.
\newblock ISBN 9781405166256.
\newblock \doi{https://doi.org/10.1002/9781405166256.ch13}.
\newblock URL \url{https://onlinelibrary.wiley.com/doi/abs/10.1002/9781405166256.ch13}.

\bibitem[Phatthiyaphaibun et~al.(2024)Phatthiyaphaibun, Chaovavanich, Polpanumas, Suriyawongkul, Lowphansirikul, and Chormai]{pythainlp}
Wannaphong Phatthiyaphaibun, Korakot Chaovavanich, Charin Polpanumas, Arthit Suriyawongkul, Lalita Lowphansirikul, and Pattarawat Chormai.
\newblock {P}y{T}hai{NLP}: {T}hai natural language processing in {P}ython, June 2024.
\newblock URL \url{https://github.com/PyThaiNLP/pythainlp/}.

\bibitem[Schächinger~Tenés et~al.(2023)Schächinger~Tenés, Weiner-Bühler, Volpin, Grob, Skoruppa, and Segerer]{language_proficiency_2023}
L.~T. Schächinger~Tenés, J.~C. Weiner-Bühler, L.~Volpin, A.~Grob, K.~Skoruppa, and R.~K. Segerer.
\newblock Language proficiency predictors of code-switching behavior in dual-language-learning children.
\newblock \emph{Bilingualism: Language and Cognition}, 26\penalty0 (5):\penalty0 942–958, 2023.
\newblock \doi{10.1017/S1366728923000081}.

\bibitem[Singh et~al.(2025)Singh, Romanou, Fourrier, Adelani, Ngui, Vila-Suero, Limkonchotiwat, Marchisio, Leong, Susanto, Ng, Longpre, Ruder, Ko, Bosselut, Oh, Martins, Choshen, Ippolito, Ferrante, Fadaee, Ermis, and Hooker]{singh2025globalmmluunderstandingaddressing}
Shivalika Singh, Angelika Romanou, Cl{\'e}mentine Fourrier, David~Ifeoluwa Adelani, Jian~Gang Ngui, Daniel Vila-Suero, Peerat Limkonchotiwat, Kelly Marchisio, Wei~Qi Leong, Yosephine Susanto, Raymond Ng, Shayne Longpre, Sebastian Ruder, Wei-Yin Ko, Antoine Bosselut, Alice Oh, Andre Martins, Leshem Choshen, Daphne Ippolito, Enzo Ferrante, Marzieh Fadaee, Beyza Ermis, and Sara Hooker.
\newblock Global {MMLU}: Understanding and addressing cultural and linguistic biases in multilingual evaluation.
\newblock In Wanxiang Che, Joyce Nabende, Ekaterina Shutova, and Mohammad~Taher Pilehvar (eds.), \emph{Proceedings of the 63rd Annual Meeting of the Association for Computational Linguistics (Volume 1: Long Papers)}, pp.\  18761--18799, Vienna, Austria, July 2025. Association for Computational Linguistics.
\newblock ISBN 979-8-89176-251-0.
\newblock \doi{10.18653/v1/2025.acl-long.919}.
\newblock URL \url{https://aclanthology.org/2025.acl-long.919/}.

\bibitem[Sprague et~al.(2025)Sprague, Yin, Rodriguez, Jiang, Wadhwa, Singhal, Zhao, Ye, Mahowald, and Durrett]{sprague2025to}
Zayne~Rea Sprague, Fangcong Yin, Juan~Diego Rodriguez, Dongwei Jiang, Manya Wadhwa, Prasann Singhal, Xinyu Zhao, Xi~Ye, Kyle Mahowald, and Greg Durrett.
\newblock To cot or not to cot? chain-of-thought helps mainly on math and symbolic reasoning.
\newblock In \emph{The Thirteenth International Conference on Learning Representations}, 2025.
\newblock URL \url{https://openreview.net/forum?id=w6nlcS8Kkn}.

\bibitem[Srivastava \& Singh(2021)Srivastava and Singh]{srivastava-singh-2021-challenges}
Vivek Srivastava and Mayank Singh.
\newblock Challenges and limitations with the metrics measuring the complexity of code-mixed text.
\newblock In Thamar Solorio, Shuguang Chen, Alan~W. Black, Mona Diab, Sunayana Sitaram, Victor Soto, Emre Yilmaz, and Anirudh Srinivasan (eds.), \emph{Proceedings of the Fifth Workshop on Computational Approaches to Linguistic Code-Switching}, pp.\  6--14, Online, June 2021. Association for Computational Linguistics.
\newblock \doi{10.18653/v1/2021.calcs-1.2}.
\newblock URL \url{https://aclanthology.org/2021.calcs-1.2/}.

\bibitem[Su et~al.(2026)Su, Pan, Bai, Liu, Dong, Huang, Lv, Hu, Zhang, Gai, and Zhou]{su2026klearreasoneradvancingreasoningcapability}
Zhenpeng Su, Leiyu Pan, Xue Bai, Dening Liu, Guanting Dong, Jiaming Huang, Minxuan Lv, Wenping Hu, Fuzheng Zhang, Kun Gai, and Guorui Zhou.
\newblock Klear-reasoner: Advancing reasoning capability via gradient-preserving clipping policy optimization, 2026.
\newblock URL \url{https://arxiv.org/abs/2508.07629}.

\bibitem[Tam et~al.(2025{\natexlab{a}})Tam, Wu, Chiu, Lin, Chen, and yi~Lee]{Tam2025LanguageMH}
Zhi~Rui Tam, Cheng-Kuang Wu, Yu~Ying Chiu, Chieh-Yen Lin, Yun-Nung Chen, and Hung yi~Lee.
\newblock Language matters: How do multilingual input and reasoning paths affect large reasoning models?
\newblock \emph{arXiv preprint arXiv:2505.17407}, 2025{\natexlab{a}}.

\bibitem[Tam et~al.(2025{\natexlab{b}})Tam, Wu, Chiu, Lin, Chen, and yi~Lee]{tam2025languagemattersmultilingualinput}
Zhi~Rui Tam, Cheng-Kuang Wu, Yu~Ying Chiu, Chieh-Yen Lin, Yun-Nung Chen, and Hung yi~Lee.
\newblock Language matters: How do multilingual input and reasoning paths affect large reasoning models?, 2025{\natexlab{b}}.
\newblock URL \url{https://arxiv.org/abs/2505.17407}.

\bibitem[Team et~al.(2025)Team, Du, Gao, Xing, Jiang, Chen, Li, Xiao, Du, Liao, Tang, Wang, Zhang, Yuan, Lu, Tang, Sung, Wei, Lai, Guo, Zhu, Ding, Hu, Yang, Zhang, Yao, Zhao, Lu, Li, Yu, Gao, Zheng, Yuan, Chen, Guo, Su, Wang, Zhao, Zhang, Liu, Yan, Wu, Shi, Ye, Yu, Dong, Zhang, Ma, Pan, Gong, Liu, Ma, Wei, Cao, Huang, Jiang, Gao, Xiong, He, Huang, Xu, Wu, He, Wei, Jia, Wu, Xu, Zu, Zhou, Pan, Charles, Li, Hu, Liu, Chen, Wang, Liu, Qin, Liu, Yang, Bao, Du, Wu, Wang, Zhou, Wang, Li, Zhu, Zhang, Wang, Yang, Huang, Huang, Xu, Yang, and Lin]{kimiteam2025kimik15scalingreinforcement}
Kimi Team, Angang Du, Bofei Gao, Bowei Xing, Changjiu Jiang, Cheng Chen, Cheng Li, Chenjun Xiao, Chenzhuang Du, Chonghua Liao, Chuning Tang, Congcong Wang, Dehao Zhang, Enming Yuan, Enzhe Lu, Fengxiang Tang, Flood Sung, Guangda Wei, Guokun Lai, Haiqing Guo, Han Zhu, Hao Ding, Hao Hu, Hao Yang, Hao Zhang, Haotian Yao, Haotian Zhao, Haoyu Lu, Haoze Li, Haozhen Yu, Hongcheng Gao, Huabin Zheng, Huan Yuan, Jia Chen, Jianhang Guo, Jianlin Su, Jianzhou Wang, Jie Zhao, Jin Zhang, Jingyuan Liu, Junjie Yan, Junyan Wu, Lidong Shi, Ling Ye, Longhui Yu, Mengnan Dong, Neo Zhang, Ningchen Ma, Qiwei Pan, Qucheng Gong, Shaowei Liu, Shengling Ma, Shupeng Wei, Sihan Cao, Siying Huang, Tao Jiang, Weihao Gao, Weimin Xiong, Weiran He, Weixiao Huang, Weixin Xu, Wenhao Wu, Wenyang He, Xianghui Wei, Xianqing Jia, Xingzhe Wu, Xinran Xu, Xinxing Zu, Xinyu Zhou, Xuehai Pan, Y.~Charles, Yang Li, Yangyang Hu, Yangyang Liu, Yanru Chen, Yejie Wang, Yibo Liu, Yidao Qin, Yifeng Liu, Ying Yang, Yiping Bao, Yulun Du, Yuxin Wu, Yuzhi Wang, Zaida
  Zhou, Zhaoji Wang, Zhaowei Li, Zhen Zhu, Zheng Zhang, Zhexu Wang, Zhilin Yang, Zhiqi Huang, Zihao Huang, Ziyao Xu, Zonghan Yang, and Zongyu Lin.
\newblock Kimi k1.5: Scaling reinforcement learning with llms, 2025.
\newblock URL \url{https://arxiv.org/abs/2501.12599}.

\bibitem[Team(2025)]{qwq32b}
Qwen Team.
\newblock Qwq-32b: Embracing the power of reinforcement learning, March 2025.
\newblock URL \url{https://qwenlm.github.io/blog/qwq-32b/}.

\bibitem[Unicode(2025)]{unicode}
Unicode.
\newblock Languages and scripts, 2025.
\newblock URL \url{https://www.unicode.org/cldr/charts/47/supplemental/languages_and_scripts.html}.

\bibitem[Wang et~al.(2025{\natexlab{a}})Wang, Lange, Adel, Ma, Str{\"o}tgen, and Schuetze]{wang-etal-2025-language-mixing}
Mingyang Wang, Lukas Lange, Heike Adel, Yunpu Ma, Jannik Str{\"o}tgen, and Hinrich Schuetze.
\newblock Language mixing in reasoning language models: Patterns, impact, and internal causes.
\newblock In Christos Christodoulopoulos, Tanmoy Chakraborty, Carolyn Rose, and Violet Peng (eds.), \emph{Proceedings of the 2025 Conference on Empirical Methods in Natural Language Processing}, pp.\  2637--2665, Suzhou, China, November 2025{\natexlab{a}}. Association for Computational Linguistics.
\newblock ISBN 979-8-89176-332-6.
\newblock \doi{10.18653/v1/2025.emnlp-main.132}.
\newblock URL \url{https://aclanthology.org/2025.emnlp-main.132/}.

\bibitem[Wang et~al.(2025{\natexlab{b}})Wang, Lange, Adel, Ma, Strötgen, and Schütze]{wang2025languagemixingreasoninglanguage}
Mingyang Wang, Lukas Lange, Heike Adel, Yunpu Ma, Jannik Strötgen, and Hinrich Schütze.
\newblock Language mixing in reasoning language models: Patterns, impact, and internal causes, 2025{\natexlab{b}}.
\newblock URL \url{https://arxiv.org/abs/2505.14815}.

\bibitem[Wang(2025)]{Kimi-K1.5-Distill-data}
Rongsheng Wang.
\newblock The dataset distilled from kimi-k1.5.
\newblock \url{https://huggingface.co/datasets/wangrongsheng/Kimi-K1.5-Distill-data}, 2025.

\bibitem[Winata et~al.(2026)Winata, Anugraha, Irawan, Das, Yoo, Dashore, Kulkarni, Zhang, Sakajo, Hudi, Ovalle, Montariol, Gaschi, Anugraha, Puranik, Ahmed, Merin, and Chersoni]{winata2026largelanguagemodelsunderstand}
Genta~Indra Winata, David Anugraha, Patrick~Amadeus Irawan, Anirban Das, Haneul Yoo, Paresh Dashore, Shreyas Kulkarni, Ruochen Zhang, Haruki Sakajo, Frederikus Hudi, Anaelia Ovalle, Syrielle Montariol, Felix Gaschi, Michael Anugraha, Rutuj~Ravindra Puranik, Zawad~Hayat Ahmed, Adril~Putra Merin, and Emmanuele Chersoni.
\newblock Can large language models understand, reason about, and generate code-switched text?, 2026.
\newblock URL \url{https://arxiv.org/abs/2601.07153}.

\bibitem[Xie et~al.(2025)Xie, Liu, Bie, Chan, Song, Wang, CHEN, and Chen]{NEURIPS2025_4847447c}
Peng Xie, Xingyuan Liu, Yequan Bie, Tsz~Wai Chan, Yangqiu Song, Yang Wang, Hao CHEN, and Kani Chen.
\newblock Switchlingua: The first large-scale multilingual and multi-ethnic code-switching dataset.
\newblock In D.~Belgrave, C.~Zhang, H.~Lin, R.~Pascanu, P.~Koniusz, M.~Ghassemi, and N.~Chen (eds.), \emph{Advances in Neural Information Processing Systems}, volume 38, Main Conference. Curran Associates, Inc., 2025.
\newblock \doi{10.52202/085713-1683}.
\newblock URL \url{https://proceedings.neurips.cc/paper_files/paper/2025/file/4847447cae911dde321ca19afe7d8707-Paper-Datasets_and_Benchmarks_Track.pdf}.

\bibitem[Xu et~al.(2025)Xu, Peng, Awadalla, Chen, Chen, Gao, Kim, Li, Ren, Shen, Wang, Xu, Gao, and Chen]{xu2025phi4minireasoningexploringlimitssmall}
Haoran Xu, Baolin Peng, Hany Awadalla, Dongdong Chen, Yen-Chun Chen, Mei Gao, Young~Jin Kim, Yunsheng Li, Liliang Ren, Yelong Shen, Shuohang Wang, Weijian Xu, Jianfeng Gao, and Weizhu Chen.
\newblock Phi-4-mini-reasoning: Exploring the limits of small reasoning language models in math, 2025.
\newblock URL \url{https://arxiv.org/abs/2504.21233}.

\bibitem[Xuan et~al.(2025)Xuan, Yang, Qi, Zeng, Xiao, Feng, Liu, Xing, Wang, Gao, Lu, Jiang, Li, Li, Yu, Dong, Gu, Li, Xie, Juefei-Xu, Khomh, Yoshie, Chen, Teodoro, Liu, Goebel, Ma, Marrese-Taylor, Lu, Iwasawa, Matsuo, and Li]{xuan2025mmluproxmultilingualbenchmarkadvanced}
Weihao Xuan, Rui Yang, Heli Qi, Qingcheng Zeng, Yunze Xiao, Aosong Feng, Dairui Liu, Yun Xing, Junjue Wang, Fan Gao, Jinghui Lu, Yuang Jiang, Huitao Li, Xin Li, Kunyu Yu, Ruihai Dong, Shangding Gu, Yuekang Li, Xiaofei Xie, Felix Juefei-Xu, Foutse Khomh, Osamu Yoshie, Qingyu Chen, Douglas Teodoro, Nan Liu, Randy Goebel, Lei Ma, Edison Marrese-Taylor, Shijian Lu, Yusuke Iwasawa, Yutaka Matsuo, and Irene Li.
\newblock Mmlu-prox: A multilingual benchmark for advanced large language model evaluation, 2025.
\newblock URL \url{https://arxiv.org/abs/2503.10497}.

\bibitem[Yang et~al.(2024)Yang, Yang, Zhang, Hui, Zheng, Yu, Li, Liu, Huang, Wei, Lin, Yang, Tu, Zhang, Yang, Yang, Zhou, Lin, Dang, Lu, Bao, Yang, Yu, Li, Xue, Zhang, Zhu, Men, Lin, Li, Tang, Xia, Ren, Ren, Fan, Su, Zhang, Wan, Liu, Cui, Zhang, and Qiu]{qwen2.5}
An~Yang, Baosong Yang, Beichen Zhang, Binyuan Hui, Bo~Zheng, Bowen Yu, Chengyuan Li, Dayiheng Liu, Fei Huang, Haoran Wei, Huan Lin, Jian Yang, Jianhong Tu, Jianwei Zhang, Jianxin Yang, Jiaxi Yang, Jingren Zhou, Junyang Lin, Kai Dang, Keming Lu, Keqin Bao, Kexin Yang, Le~Yu, Mei Li, Mingfeng Xue, Pei Zhang, Qin Zhu, Rui Men, Runji Lin, Tianhao Li, Tianyi Tang, Tingyu Xia, Xingzhang Ren, Xuancheng Ren, Yang Fan, Yang Su, Yichang Zhang, Yu~Wan, Yuqiong Liu, Zeyu Cui, Zhenru Zhang, and Zihan Qiu.
\newblock Qwen2.5 technical report.
\newblock \emph{arXiv preprint arXiv:2412.15115}, 2024.

\bibitem[Yang et~al.(2025)Yang, Li, Yang, Zhang, Hui, Zheng, Yu, Gao, Huang, Lv, Zheng, Liu, Zhou, Huang, Hu, Ge, Wei, Lin, Tang, Yang, Tu, Zhang, Yang, Yang, Zhou, Zhou, Lin, Dang, Bao, Yang, Yu, Deng, Li, Xue, Li, Zhang, Wang, Zhu, Men, Gao, Liu, Luo, Li, Tang, Yin, Ren, Wang, Zhang, Ren, Fan, Su, Zhang, Zhang, Wan, Liu, Wang, Cui, Zhang, Zhou, and Qiu]{yang2025qwen3technicalreport}
An~Yang, Anfeng Li, Baosong Yang, Beichen Zhang, Binyuan Hui, Bo~Zheng, Bowen Yu, Chang Gao, Chengen Huang, Chenxu Lv, Chujie Zheng, Dayiheng Liu, Fan Zhou, Fei Huang, Feng Hu, Hao Ge, Haoran Wei, Huan Lin, Jialong Tang, Jian Yang, Jianhong Tu, Jianwei Zhang, Jianxin Yang, Jiaxi Yang, Jing Zhou, Jingren Zhou, Junyang Lin, Kai Dang, Keqin Bao, Kexin Yang, Le~Yu, Lianghao Deng, Mei Li, Mingfeng Xue, Mingze Li, Pei Zhang, Peng Wang, Qin Zhu, Rui Men, Ruize Gao, Shixuan Liu, Shuang Luo, Tianhao Li, Tianyi Tang, Wenbiao Yin, Xingzhang Ren, Xinyu Wang, Xinyu Zhang, Xuancheng Ren, Yang Fan, Yang Su, Yichang Zhang, Yinger Zhang, Yu~Wan, Yuqiong Liu, Zekun Wang, Zeyu Cui, Zhenru Zhang, Zhipeng Zhou, and Zihan Qiu.
\newblock Qwen3 technical report, 2025.
\newblock URL \url{https://arxiv.org/abs/2505.09388}.

\bibitem[Ye et~al.(2025)Ye, Huang, Xiao, Chern, Xia, and Liu]{ye2025limoreasoning}
Yixin Ye, Zhen Huang, Yang Xiao, Ethan Chern, Shijie Xia, and Pengfei Liu.
\newblock Limo: Less is more for reasoning, 2025.
\newblock URL \url{https://arxiv.org/abs/2502.03387}.

\bibitem[Yong et~al.(2025)Yong, Adilazuarda, Mansurov, Zhang, Muennighoff, Eickhoff, Winata, Kreutzer, Bach, and Aji]{yong2025crosslingualreasoningtesttimescaling}
Zheng-Xin Yong, M.~Farid Adilazuarda, Jonibek Mansurov, Ruochen Zhang, Niklas Muennighoff, Carsten Eickhoff, Genta~Indra Winata, Julia Kreutzer, Stephen~H. Bach, and Alham~Fikri Aji.
\newblock Crosslingual reasoning through test-time scaling, 2025.
\newblock URL \url{https://arxiv.org/abs/2505.05408}.

\bibitem[Yu et~al.(2025)Yu, Jeong, Pawar, Shin, Jin, Myung, Oh, and Augenstein]{yu2025entangled}
Haeun Yu, Seogyeong Jeong, Siddhesh Pawar, Jisu Shin, Jiho Jin, Junho Myung, Alice Oh, and Isabelle Augenstein.
\newblock Entangled in representations: Mechanistic investigation of cultural biases in large language models.
\newblock \emph{arXiv e-prints}, pp.\  arXiv--2508, 2025.

\bibitem[Zhao et~al.(2026)Zhao, Liu, Schuetze, and Hedderich]{zhao-etal-2026-comprehensive}
Raoyuan Zhao, Yihong Liu, Hinrich Schuetze, and Michael~A. Hedderich.
\newblock A comprehensive evaluation of multilingual chain-of-thought reasoning: Performance, consistency, and faithfulness across languages.
\newblock In Vera Demberg, Kentaro Inui, and Llu{\'i}s Marquez (eds.), \emph{Findings of the {A}ssociation for {C}omputational {L}inguistics: {EACL} 2026}, pp.\  5223--5247, Rabat, Morocco, March 2026. Association for Computational Linguistics.
\newblock ISBN 979-8-89176-386-9.
\newblock \doi{10.18653/v1/2026.findings-eacl.276}.
\newblock URL \url{https://aclanthology.org/2026.findings-eacl.276/}.

\bibitem[Zheng et~al.(2024)Zheng, Zhang, Zhang, Ye, Luo, Feng, and Ma]{zheng2024llamafactory}
Yaowei Zheng, Richong Zhang, Junhao Zhang, Yanhan Ye, Zheyan Luo, Zhangchi Feng, and Yongqiang Ma.
\newblock Llamafactory: Unified efficient fine-tuning of 100+ language models.
\newblock In \emph{Proceedings of the 62nd Annual Meeting of the Association for Computational Linguistics (Volume 3: System Demonstrations)}, Bangkok, Thailand, 2024. Association for Computational Linguistics.
\newblock URL \url{http://arxiv.org/abs/2403.13372}.

\bibitem[Zhou et~al.(2025)Zhou, Chen, Shi, Zhang, Lei, Feng, Xiong, Yan, Wang, Cao, Yin, Wang, Dai, Dong, Wang, and Huang]{zhou-etal-2025-socialeval}
Jinfeng Zhou, Yuxuan Chen, Yihan Shi, Xuanming Zhang, Leqi Lei, Yi~Feng, Zexuan Xiong, Miao Yan, Xunzhi Wang, Yaru Cao, Jianing Yin, Shuai Wang, Quanyu Dai, Zhenhua Dong, Hongning Wang, and Minlie Huang.
\newblock {S}ocial{E}val: Evaluating social intelligence of large language models.
\newblock In Wanxiang Che, Joyce Nabende, Ekaterina Shutova, and Mohammad~Taher Pilehvar (eds.), \emph{Proceedings of the 63rd Annual Meeting of the {A}ssociation for {C}omputational {L}inguistics (Volume 1: Long Papers)}, pp.\  30958--31012, Vienna, Austria, July 2025. Association for Computational Linguistics.
\newblock ISBN 979-8-89176-251-0.
\newblock \doi{10.18653/v1/2025.acl-long.1496}.
\newblock URL \url{https://aclanthology.org/2025.acl-long.1496/}.

\end{thebibliography}
\bibliographystyle{iclr2027_conference}

\appendix
\section{Appendix}
\subsection{Reasoning models used to develop code-switched reasoning behavior taxonomy}\label{sec:taxonomy-models}
\begin{longtable}{ll}
    
    \label{tab:models} \\
    \toprule
Model & Citation \\
\midrule
\endfirsthead
\toprule
Model & Citation \\
\midrule
\endhead
DeepSeek-R1 and its distilled variants & \cite{Guo_2025} \\
gemini-2.0-flash-thinking-exp-1219  & \cite{gemini, geminiteam2025geminifamilyhighlycapable}\\
Claude 3.7 Sonnet & \cite{claude2.7sonnet}\\
Llama-3.1-Nemotron-Nano-8B-v1 & \cite{bercovich2025llamanemotronefficientreasoningmodels}\\
gemma-3-4b-it & \cite{gemma-3-4b-it}\\
Phi-4-mini-reasoning & \cite{xu2025phi4minireasoningexploringlimitssmall}\\
QwQ-32B & \cite{qwen2.5, qwq32b}\\
Magistral Small 1.0 & \cite{mistralai2025magistral}\\
Kimi k1.5  & \cite{kimiteam2025kimik15scalingreinforcement}\\
Klear-Reasoner-8B and Klear-Reasoner-8B-SFT & \cite{su2026klearreasoneradvancingreasoningcapability}\\
\bottomrule
\end{longtable}
\subsection{Languages supported by models used to develop code-switched reasoning behavior taxonomy}\label{sec:taxonomy-langs}
\begin{longtable}{p{0.3\linewidth}p{0.3\linewidth}p{0.3\linewidth}}
    \label{tab:languages}\\
    \toprule
       Model & Officially supported languages prompted on in our study & Unsupported languages prompted on in our study\\
        \midrule
        \endfirsthead
        \toprule
         Model & Officially supported languages prompted on in our study & Unsupported languages prompted on in our study\\
        \midrule
        \endhead
        DeepSeek-R1-Distill-Qwen-1.5B & en, zh & es, ar, hi, de, fr, pt, ja, ko, bn, sw, it, id, yo, my, th, ru\\
        DeepSeek-R1-Distill-Qwen-7B & en, zh & es, ar, hi, de, fr, pt, ja, ko, bn, sw, it, id, yo, my, th, ru\\
        DeepSeek-R1-Distill-Llama-8B & en, zh & es, ar, hi, de, fr, pt, ja, ko, bn, sw, it, id, yo, my, th, ru\\
        DeepSeek-R1-Distill-Qwen-14B & en, zh & es, ar, hi, de, fr, pt, ja, ko, bn, sw, it, id, yo, my, th, ru\\
        DeepSeek-R1-Distill-Qwen-32B & en, zh & es, ar, hi, de, fr, pt, ja, ko, bn, sw, it, id, yo, my, th, ru \\
        DeepSeek-R1-Distill-Llama-70B & en, zh & es, ar, hi, de, fr, pt, ja, ko, bn, sw, it, id, yo, my, th, ru\\
        DeepSeek-R1 & en, zh & es, ar, hi, de, fr, pt, ja, ko, bn, sw, it, id, yo, my, th, ru\\
        Llama-3.1-Nemotron-Nano-8B-v1 & en, de, fr, it, pt, es, th, hi & zh, ar, ja, ko, bn, sw, id, yo, my, ru\\
        phi-4-mini-reasoning & en & zh, es, ar, hi, de, fr, pt, ja, ko, bn, sw, it, id, yo, my, th, ru\\
        gemma-3-4b-it & $140+$ languages\footnote{The Gemma 3 paper and website do not specify which languages are officially supported, but we assume that the over 140 languages covered by the model include those languages that we prompt the model on in this work \citep{gemma-3-4b-it, gemmateam2025gemma3technicalreport}.} & Unknown\\
        magistral-small-2506 & en, fr, de, el, hi, id, it, ja, ko, ms, ne, pl, pt, ro, ru, sr, es, tr, uk, vi, ar, bn, zh, fa & sw, yo, my, th\\
        gemini-2.0-flash-thinking-exp-1219 & ar, bn, bg, zh, hr, cs, da, nl, en, et, fi, fr, de, el, iw, hi, hu, id, it, ja, ko, lv, lt, no, pl, pt, ro, ru, sr, sk, sl, es, sw, sv, th, tr, uk, vi & None\footnote{The Gemini 2.0 paper and website do not specify which languages are officially supported, but we assume broad multilingual coverage by this (then-frontier) model \citep{geminiteam2025geminifamilyhighlycapable}.}\\
        Claude 3.7 Sonnet & en, es, pt, it, fr, id, de, ar, zh, ko, ja, hi, bn, sw, yo & None\footnote{Similar to the Gemma and Gemini models, we assume this strong closed model provides broad language coverage.}\\
        QwQ-32B & en, zh & es, ar, hi, de, fr, pt, ja, ko, bn, sw, it, id, yo, my, th, ru\\
        Kimi k1.5 & en, zh & es, ar, hi, de, fr, pt, ja, ko, bn, sw, it, id, yo, my, th, ru\\
        Klear-Reasoner-8B and Klear-Reasoner-8B-SFT & en & zh\\
    \bottomrule
\end{longtable}
\subsection{Datasets used to develop code-switched reasoning behavior taxonomy}\label{sec:taxonomy-data}
\begin{longtable}{p{0.35\linewidth}p{0.25\linewidth}p{0.3\linewidth}}
    \label{tab:dataset_citations} \\
    \toprule
    Dataset & Type & Citation \\
    \midrule
    \endfirsthead
    \toprule
    Dataset & Type & Citation \\
    \midrule
    \endhead
    s1K-1.1, s1k-magistral-small-2506, and s1K-claude-3-7-sonnet & Coding, math, puzzle, science, legal, logic, humanities & \cite{muennighoff2025s} \\
    Global MMLU & Math, science, humanities, medical, business, legal & \cite{singh2025globalmmluunderstandingaddressing}\\
    MMLU ProX & Math, science, humanities, medical, business, legal & \cite{xuan2025mmluproxmultilingualbenchmarkadvanced}\\
    BIG-Bench Extra Hard & Puzzle, logic, commonsense, linguistics, spatial & \cite{kazemi2025bigbenchextrahard}\\
    UniMoral & Cultural, social & \cite{kumar2025rulesmeantbrokenunderstanding}\\
    OpenMathReasoning & Math & \cite{moshkov2025aimo2winningsolutionbuilding}\\
    Kimi-K1.5-Distill-data & Math & \cite{Kimi-K1.5-Distill-data, ye2025limoreasoning}\\
    CulturalBench & Cultural & \cite{chiu2025culturalbenchrobustdiversechallenging, chiu-etal-2025-culturalbench}\\
    SocialEval & Social & \cite{zhou-etal-2025-socialeval}\\
    CultureScope & Cultural & \cite{yu2025entangled}\\
\bottomrule
\end{longtable}

\subsection{Configuration for generating reasoning examples to develop taxonomy of code-switched reasoning behaviors}\label{sec:taxonomy-config}
We generate reasoning examples on the BBEH, UniMoral, Global MMLU Lite, and MMLU ProX Lite datasets. All other data used in our experiments is drawn from preexisting reasoning example datasets.

\textbf{BBEH} Following the recommendations from the model creators,\footnote{\url{https://huggingface.co/deepseek-ai/DeepSeek-R1}} for prompting DeepSeek-R1-Distill-Qwen-7B and DeepSeek-R1-Distill-Llama-8B we set \texttt{temperature=0.6, top\_p=0.95}. We also set \texttt{max\_tokens=32768}, which is the maximum allowed by the DeepSeek distilled models.

\textbf{UniMoral} For both Phi-4-mini-reasoning and DeepSeek-R1-Distill-Llama-8B, we set \texttt{temperature=0.6}, \texttt{top\_p=0.95}, and \texttt{max\_tokens=32768}.

\textbf{Global MMLU Lite} For DeepSeek-R1-Distill-Qwen-1.5B, DeepSeek-R1-Distill-Qwen-7B, DeepSeek-R1-Distill-Llama-8B, and DeepSeek-R1-Distill-Qwen-14B, we set \texttt{temperature=0.6}, \texttt{top\_p=0.9}, and \texttt{max\_tokens=32768}, again following the recommendations from the model creators.\footnote{\url{https://huggingface.co/deepseek-ai/DeepSeek-R1}} For DeepSeek-R1-Distill-Qwen-32B and DeepSeek-R1-Distill-Llama-70B, we set \texttt{temperature=1.0}, \texttt{top\_p=1.0}, and \texttt{max\_tokens=32768}. For prompting DeepSeek-R1, we leave the API defaults as is and use the January 2025 model checkpoint.\footnote{\url{https://api-docs.deepseek.com/news/news250120}}

\textbf{MMLU ProX Lite} For Llama-3.1-Nemotron-Nano-8B-v1, we set \texttt{temperature=0.6}, \texttt{top\_p=0.95}, and \texttt{max\_tokens=32768}, following NVIDIA's recommendations.\footnote{\url{https://build.nvidia.com/nvidia/llama-3_1-nemotron-nano-8b-v1/modelcard}} We also turn reasoning mode on. For gemma-3-4b-it and DeepSeek-R1-Distill-Qwen-7B, we set \texttt{temperature=0} and \texttt{max\_tokens=2048}, as we initially followed the defaults in \texttt{lm-evaluation-harness}.\footnote{\url{https://github.com/EleutherAI/lm-evaluation-harness/blob/main/lm_eval/tasks/mmlu_prox/en/_en_template_yaml}} For DeepSeek-R1-Distill-Llama-8B, we set \texttt{temperature=0} and \texttt{max\_tokens=32768}.

\subsection{Prompt for brainstorming code-switched reasoning behavior taxonomy}\label{sec:brainstorm}
\begin{tcolorbox}[colback=gray!20, colframe=black!50, boxrule=0.5pt, arc=3mm, left=3mm, right=3mm, top=2mm, bottom=2mm]
Here is a problem and the reasoning process that a model generated when it tried to solve the problem.
    
Problem: (enclosed in double backticks)

\begin{verbatim}
``
{problem}
``
\end{verbatim}

Reasoning process: (enclosed in triple backticks, the reasoning process has been split into distinct reasoning steps in the format of \verb|<step_idx><reasoning_step_content></step_idx>|)
\begin{verbatim}
```
{reasoning}
```
\end{verbatim}

You are tasked with analyzing any code-switching strategies used in the above reasoning, where code-switching is defined as the use of multiple scripts or languages. Your goal is to extract and describe patterns based on various criteria that characterize how the model uses code-switching in its reasoning.

Please follow these guidelines:

1. Identify multiple *meaningful criteria* that differentiate code-switching strategies. Each criterion should have a clear and descriptive name that reflects a real aspect of the code-switching process. **Do not use generic placeholders like `Criterion 1'.**

2. For each criterion, provide a brief description.

3. Present your analysis in the following format, using <patterns> and </patterns> tags to enclose the list:

\begin{verbatim}
<patterns>
Descriptive Criterion Name (Description of the Criterion)
Descriptive Criterion Name (Description of the Criterion)
...
Descriptive Criterion Name (Description of the Criterion)
</patterns>
\end{verbatim}

4. Do not include any additional explanations or commentary within the <patterns> tags.

5. If no code-switching is observed in the reasoning, simply return "<patterns></patterns>".
\end{tcolorbox}
\subsection{BERTopic modeling details for code-switched reasoning behavior taxonomy}\label{sec:bertopic}
\begin{figure}[!htbp]
    \centering
    \includegraphics[trim={0 0 0 0},clip, width=0.8\textwidth]{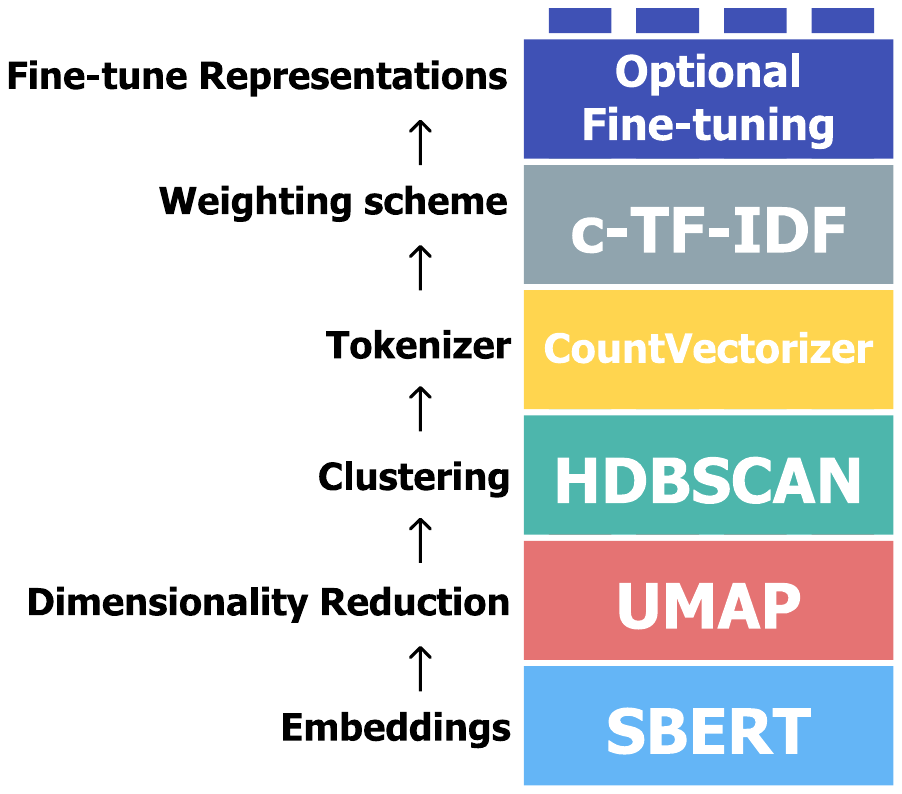} 
    \caption{Overview of the BERTopic pipeline from \citet{grootendorst2022bertopicneuraltopicmodeling}.}
    \label{fig:bertopic}
\end{figure}
All components of our topic modeling pipeline used in the development of the code-switched reasoning taxonomy are implemented in BERTopic (Figure \ref{fig:bertopic}). We use all-mpnet-base-v2 to embed the text of each criterion, UMAP for dimension reduction, HDBSCAN for clustering, and Gemini 2.5 Flash for generating topic representations. For UMAP dimension reduction, we set \lstinline{n_neighbors=15, n_components=5, min_dist=0.0, metric=`cosine'}. For HDBSCAN clustering, we set \lstinline{min_cluster_size=15, metric=`euclidean', cluster_selection_method=`eom'}, and \lstinline{prediction_data=True}. We use \lstinline{CountVectorizer} with \lstinline{stop_words=``english''} for topic tokenization and c-TF-IDF for extracting topic words. Finally, we use Gemini 2.5 Flash to generate more interpretable topic representations.

\subsection{Human evaluation of LLM annotations}\label{sec:human}
To validate the LLM-brainstormed criteria to differentiate code-switching strategies, we follow \citet{lee2025cotencyclopediaanalyzingpredicting} in asking our evaluator to answer the following yes-no question for each instance: ``Are the automatically generated, detailed criteria plausible?''

\subsection{Word-level language identification (LID)}\label{sec:word-lid}
For most languages we use whitespace to detect word boundaries. For Chinese, Japanese, Thai, and Burmese, which lack whitespace between words, we use jieba,\footnote{\url{https://github.com/fxsjy/jieba}} nagisa,\footnote{\url{https://github.com/taishi-i/nagisa}}, PyThaiNLP \citep{pythainlp}, and myanmar\_words\footnote{\url{https://github.com/linhtutkyawdev/myanmar_words}} to obtain word tokens. Jieba has been benchmarked\footnote{\url{https://github.com/howl-anderson/Chinese_tokenizer_benchmark}} for accuracy against other publicly available Chinese tokenizers, demonstrating comparable or better performance with most of them (F1 score of 0.815). PyThaiNLP is also competitive when benchmarked\footnote{\url{https://github.com/PyThaiNLP/pythainlp/blob/main/tokenization-benchmark.md}} against the (limited) other available Thai tokenizers (F1 score of 0.73). 

For word-level language identification, we primarily rely on the \textit{Lingua} library, which is computationally efficient, designed to work well on short texts, and covers most languages in our experiments, with the exception of Amharic, Burmese, and Igbo.\footnote{\url{https://github.com/pemistahl/lingua-py}} Lingua has been benchmarked against other standard open-source LID tools such as Google's Compact Language Detector v3 (CLD3) and achieves better performance in most languages.

For Amharic and Burmese, which are written in non-Latin scripts, we use GlotScript to separate Amharic and Burmese words from English words \citep{kargaran-etal-2024-glotscript}. We use GlotScript because it is specifically designed to cover low-resource languages. For Igbo, we use the FastText language identification model along with the English dictionary from \citet{marchisio-etal-2024-understanding} to catch English words that are missed by FastText.

To validate our LID approach on in-domain data, we assemble a validation set of 140 reasoning examples generated from prompting Qwen3-Next-80B-A3B-Thinking in Amharic, Hindi, Igbo, Indonesian, Malay, Swahili, and Yoruba on Global MMLU \citep{singh2025globalmmluunderstandingaddressing}. We achieve an average word-level LID accuracy of 0.933 on our validation set.

\subsection{Prompt for evaluating accuracy of code-switching}\label{sec:acc-eval}
\begin{tcolorbox}[colback=gray!20, colframe=black!50, boxrule=0.5pt, arc=3mm, left=3mm, right=3mm, top=2mm, bottom=2mm]
Your task is to score the accuracy of the code-switching in the following text on a scale from 1 to 3, where 1 is the lowest and 3 is the highest. Accuracy measures whether the code-switched terms are used correctly.\\

Text to score:\\

\begin{verbatim}

{reasoning}

\end{verbatim}

Use the following rubric to guide your scoring. The rubric is formatted as "\verb|<score>. <description>|":\\

1. Low Accuracy: Code-switched terms are incorrect or inappropriate.\\

2. Moderate Accuracy: Most code-switched terms are appropriate but with minor mistakes.\\

3. High Accuracy: Code-switched terms are accurate and appropriately used.\\

Respond with only your score in the following format: <score>
\end{tcolorbox}
\subsection{Prompt for evaluating fluency of code-switching}\label{sec:fl-eval}
\begin{tcolorbox}[colback=gray!20, colframe=black!50, boxrule=0.5pt, arc=3mm, left=3mm, right=3mm, top=2mm, bottom=2mm]
Your task is to score the fluency of the code-switching in the following text on a scale from 1 to 3, where 1 is the lowest and 3 is the highest. Fluency measures how natural and easy to understand the text is, considering grammar, syntax, and the smooth integration of code-switching.\\

Text to score:

\begin{verbatim}

{reasoning}

\end{verbatim}

Use the following rubric to guide your scoring. The rubric is formatted as ``\verb|<score>. <description>|'':\\

1. Low Fluency: The text is difficult to understand, awkward, or features poor grammar or syntax; or, the code-switching disrupts the flow of the text.\\

2. Moderate Fluency: The text is understandable but may have awkward or unnatural phrasing, grammar and syntax is acceptable, and code-switching is somewhat smooth but not perfectly integrated.\\

3. High Fluency: The text is natural and easy to understand; grammar and syntax are good; and code-switching is smooth and seamless, enhancing the sentence flow.\\

Respond with only your score in the following format: <score>
\end{tcolorbox}
\subsection{Reasoning-related subjects in Global MMLU}\label{sec:reasoning-subs}
\begin{itemize}
    \item abstract\_algebra
    \item business\_ethics
    \item college\_biology
    \item college\_chemistry
    \item college\_computer\_science
    \item college\_mathematics
    \item college\_medicine
    \item college\_physics
    \item computer\_security 
    \item conceptual\_physics
    \item econometrics
    \item electrical\_engineering
    \item elementary\_mathematics
    \item formal\_logic
    \item high\_school\_biology
    \item high\_school\_chemistry
    \item high\_school\_computer\_science
    \item high\_school\_european\_history
    \item high\_school\_mathematics
    \item high\_school\_physics
    \item high\_school\_statistics
    \item high\_school\_us\_history
    \item high\_school\_world\_history
    \item jurisprudence
    \item logical\_fallacies
    \item machine\_learning
    \item moral\_scenarios
    \item professional\_accounting
    \item professional\_law
    \item professional\_medicine
    \item professional\_psychology
\end{itemize}

\subsection{MT pipeline for SFT data synthesis}\label{sec:mt}
We use SeamlessM4T v2 for translation from English into all languages, except Malay \citep{communication2023seamlessmultilingualexpressivestreaming}. For Malay, we use the 3.3B variant of NLLB-200, since unlike Seamless M4T v2, this model supports translation from English into Standard Malay \citep{nllbteam2022languageleftbehindscaling}. We select these models because they are the strongest available open-source massively multilingual MT models.

To further improve the quality of the translations, we use \citet{chen-etal-2024-iterative}'s approach for translation refinement with the 70B Apertus model and filter the resulting translations for quality \citep{apertus2025apertusdemocratizingopencompliant}.

Given a translation from the 3.3B variant of the NLLB-200 model or from the Seamless M4T v2 model, we prompt the 70B Apertus model with the following prompt sourced from \citeauthor{chen-etal-2024-iterative} to improve the translation:

\begin{lstlisting}
    Source: {English source text}
    Bad translation: {Translation into target language}
    Please give me a better {target language} translation without any explanation.
\end{lstlisting}

We use the default recommended sampling parameters for the Apertus model.\footnote{\url{https://huggingface.co/swiss-ai/Apertus-70B-2509}}

Since Apertus does not always comply with the instructions not to provide additional explanation, we perform regex-based postprocessing to cleanup extraneous text. After postprocessing, we only keep the output from Apertus if the number of whitespace-separated tokens is less than twice as long as in the original translation. Otherwise, we retain the original machine translation from NLLB-200/Seamless M4T v2. We perform an additional quality filtering step to remove any translations that contain sequences of character or whitespace-separated token quadrigrams repeated more than 10 times, or sequences of the ``\#'' or ``*'' characters repeated more than 10 times.

\subsection{Instance-level language identification}\label{sec:instance-lid}
Using the prompt below with the LID validation set from \ref{sec:word-lid}, we obtain comparable accuracies of 0.892 for Qwen3-Next-80B-A3B-Instruct (supports 119 languages and dialects) and 0.942 for Qwen3.5-27B (supports 201 languages and dialects) \citep{yang2025qwen3technicalreport}.\footnote{\url{https://qwen.ai/blog?id=qwen3.5}}
\begin{tcolorbox}[colback=gray!20, colframe=black!50, boxrule=0.5pt, arc=3mm, left=3mm, right=3mm, top=2mm, bottom=2mm]
Identify the languages that appear in the following text:\\

\begin{verbatim}
{text}

\end{verbatim}

Use the following language codes. The language codes are provided in the format ``\verb|<language_name>: <language_code>|'':\\

\begin{verbatim}
{language_codes}

\end{verbatim}

If you identify a language that is not listed above, please use the appropriate two-letter ISO 639 language code for that language.\\

Return your answer as a list in the following format: ``\verb|<language_code1>, <language_code2>, ...|''\\

If you are unable to identify any of the languages in the provided text, return "None"
\end{tcolorbox}

\subsection{SFT Configuration}\label{sec:sft-config}
We use the following configuration with LLaMA Factory for SFT \citep{zheng2024llamafactory}. We set the \verb|cutoff_len| by computing the 95th percentile of the distribution of token lengths when considering the combined total length of input and target token sequences in each training dataset. \verb|cutoff_len| is specific to each model, language, and SFT task. For the native-language reasoning, English-language reasoning, strategically code-switched reasoning, and synthetically code-switched reasoning tasks, we set \verb|eval_steps| and \verb|save_steps| to \verb|10|. For SFT for machine translation and reasoning prompt translation into English, we set \verb|eval_steps| and \verb|save_steps| to \verb|100|. We set \verb|gradient_accumulation_steps| to ensure an effective batch size of \verb|16|.
\begin{tcolorbox}[colback=gray!20, colframe=black!50, boxrule=0.5pt, arc=3mm, left=3mm, right=3mm, top=2mm, bottom=2mm]
\begin{verbatim}
model_name_or_path: {deepseek-ai/DeepSeek-R1-Distill-Llama-8B, 
    microsoft/Phi-4-reasoning, Qwen/Qwen3-8B}
stage: sft
do_train: true
finetuning_type: lora
lora_target: all
dataset_dir: {path2dataset_dir}
dataset: {dataset_name}
eval_dataset: {eval_dataset_name}
template: {deepseekr1, phi4, qwen3}
cutoff_len: {cutoff_len}
output_dir: {output_dir}
run_name: {run_name}
learning_rate: 1.0e-6
eval_strategy: steps
eval_steps: {eval_steps}
save_steps: {save_steps}
gradient_accumulation_steps: {grad_accum_steps}
overwrite_cache: true
preprocessing_num_workers: 16
logging_steps: 10
plot_loss: true
overwrite_output_dir: true
per_device_train_batch_size: 1 
num_train_epochs: 3.0
lr_scheduler_type: cosine
warmup_ratio: 0.1
bf16: true
ddp_timeout: 180000000
per_device_eval_batch_size: 1
report_to: wandb
\end{verbatim}
\end{tcolorbox}

\subsection{Accuracy on \encore corpus SFT test set (curated from Global MMLU and Global PIQA)}\label{sec:acc}

\textbf{Language codes:} hi = Hindi, id = Indonesian, ms = Malay, am = Amharic, ig = Igbo, sw = Swahili, yo = Yoruba. 

\textbf{SFT conditions:} No-SFT baseline = reasoning model without any SFT intervention applied; English reasoning = fine-tune for reasoning in English; Machine translation = fine-tune to perform machine translation; Native reasoning = fine-tune to perform native-language reasoning; Strategic/Synthetic CSW = fine-tune to perform strategically/synthetically code-switched reasoning; Translated prompts = fine-tune to translate reasoning prompts into English. Yoruba has no native-reasoning or synthetic-CSW fine-tuning conditions due to insufficient synthesized data. 

\textbf{Evaluation configurations:} the \encore test set is split into Global PIQA and Global MMLU subsets; Global PIQA uses fully native-language prompts.

{\footnotesize
\setlength{\tabcolsep}{1.5pt}
\begin{longtable}{lccccccc}
\caption{Accuracy by condition, language, and model on \encore test set. Subscripts are Wilson 95\% confidence intervals.}
\label{tab:accuracy-appendix} \\
\toprule
Condition & hi & id & ms & am & ig & sw & yo \\
\midrule
\endfirsthead
\toprule
Condition & hi & id & ms & am & ig & sw & yo \\
\midrule
\endhead
\multicolumn{8}{l}{\textit{Global MMLU} --- DeepSeek-R1-Distill-Llama-8B ($n=500$ per cell)} \\
No-SFT baseline & $0.43_{[.39,.48]}$ & $0.54_{[.49,.58]}$ & $0.52_{[.47,.56]}$ & $0.23_{[.20,.27]}$ & $0.22_{[.19,.26]}$ & $0.14_{[.11,.17]}$ & $0.28_{[.25,.33]}$ \\
English reasoning & $0.44_{[.40,.48]}$ & $0.59_{[.54,.63]}$ & $0.52_{[.47,.56]}$ & $0.23_{[.20,.27]}$ & $0.21_{[.17,.24]}$ & $0.17_{[.14,.20]}$ & $0.30_{[.26,.34]}$ \\
Machine translation & $0.40_{[.36,.45]}$ & $0.56_{[.51,.60]}$ & $0.52_{[.47,.56]}$ & $0.24_{[.20,.28]}$ & $0.30_{[.26,.34]}$ & $0.21_{[.18,.25]}$ & $0.25_{[.22,.29]}$ \\
Native reasoning & $0.41_{[.37,.46]}$ & $0.55_{[.51,.59]}$ & $0.56_{[.51,.60]}$ & $0.21_{[.18,.25]}$ & $0.23_{[.20,.27]}$ & $0.21_{[.18,.25]}$ & -- \\
Strategic CSW & $0.42_{[.38,.46]}$ & $0.57_{[.53,.61]}$ & $0.54_{[.50,.59]}$ & $0.23_{[.20,.27]}$ & $0.21_{[.18,.25]}$ & $0.17_{[.14,.21]}$ & $0.28_{[.24,.32]}$ \\
Synthetic CSW & $0.43_{[.39,.47]}$ & $0.58_{[.53,.62]}$ & $0.55_{[.50,.59]}$ & $0.24_{[.20,.28]}$ & $0.22_{[.18,.26]}$ & $0.17_{[.14,.21]}$ & -- \\
Translated prompts & $0.42_{[.38,.47]}$ & $0.55_{[.51,.59]}$ & $0.55_{[.51,.60]}$ & $0.24_{[.20,.28]}$ & $0.28_{[.25,.33]}$ & $0.21_{[.17,.25]}$ & $0.24_{[.21,.28]}$ \\
\multicolumn{8}{l}{\textit{Global MMLU} --- Phi-4-reasoning ($n=500$ per cell)} \\
No-SFT baseline & $0.72_{[.68,.76]}$ & $0.76_{[.72,.80]}$ & $0.75_{[.71,.79]}$ & $0.25_{[.22,.29]}$ & $0.33_{[.29,.38]}$ & $0.65_{[.61,.69]}$ & $0.32_{[.28,.36]}$ \\
English reasoning & $0.72_{[.68,.76]}$ & $0.74_{[.70,.78]}$ & $0.73_{[.69,.77]}$ & $0.24_{[.20,.28]}$ & $0.30_{[.26,.34]}$ & $0.62_{[.58,.66]}$ & $0.28_{[.25,.33]}$ \\
Machine translation & $0.73_{[.69,.77]}$ & $0.73_{[.69,.77]}$ & $0.73_{[.69,.77]}$ & $0.24_{[.21,.28]}$ & $0.32_{[.28,.36]}$ & $0.63_{[.59,.68]}$ & $0.26_{[.23,.30]}$ \\
Native reasoning & $0.74_{[.70,.77]}$ & $0.76_{[.72,.80]}$ & $0.72_{[.67,.75]}$ & $0.25_{[.21,.29]}$ & $0.32_{[.28,.36]}$ & $0.63_{[.58,.67]}$ & -- \\
Strategic CSW & $0.73_{[.69,.77]}$ & $0.73_{[.69,.77]}$ & $0.72_{[.68,.76]}$ & $0.22_{[.19,.26]}$ & $0.31_{[.27,.35]}$ & $0.64_{[.59,.68]}$ & $0.27_{[.23,.31]}$ \\
Synthetic CSW & $0.75_{[.71,.79]}$ & $0.74_{[.70,.78]}$ & $0.73_{[.69,.77]}$ & $0.24_{[.21,.28]}$ & $0.32_{[.28,.36]}$ & $0.68_{[.64,.72]}$ & -- \\
Translated prompts & $0.71_{[.67,.75]}$ & $0.77_{[.73,.80]}$ & $0.73_{[.69,.77]}$ & $0.25_{[.21,.29]}$ & $0.31_{[.27,.35]}$ & $0.64_{[.60,.68]}$ & $0.26_{[.23,.30]}$ \\
\multicolumn{8}{l}{\textit{Global MMLU} --- Qwen3-8B ($n=500$ per cell)} \\
No-SFT baseline & $0.74_{[.70,.77]}$ & $0.79_{[.76,.83]}$ & $0.76_{[.72,.80]}$ & $0.30_{[.26,.34]}$ & $0.31_{[.27,.35]}$ & $0.41_{[.37,.45]}$ & $0.28_{[.24,.32]}$ \\
English reasoning & $0.73_{[.69,.77]}$ & $0.77_{[.73,.81]}$ & $0.78_{[.74,.81]}$ & $0.31_{[.27,.35]}$ & $0.35_{[.31,.39]}$ & $0.40_{[.36,.45]}$ & $0.26_{[.22,.30]}$ \\
Machine translation & $0.74_{[.70,.77]}$ & $0.77_{[.73,.80]}$ & $0.73_{[.69,.77]}$ & $0.39_{[.35,.44]}$ & $0.38_{[.34,.43]}$ & $0.47_{[.42,.51]}$ & $0.30_{[.26,.34]}$ \\
Native reasoning & $0.75_{[.71,.79]}$ & $0.78_{[.74,.82]}$ & $0.74_{[.70,.78]}$ & $0.30_{[.26,.34]}$ & $0.32_{[.28,.37]}$ & $0.44_{[.40,.49]}$ & -- \\
Strategic CSW & $0.74_{[.70,.78]}$ & $0.78_{[.74,.81]}$ & $0.76_{[.72,.80]}$ & $0.31_{[.27,.35]}$ & $0.33_{[.29,.37]}$ & $0.44_{[.40,.48]}$ & $0.31_{[.27,.35]}$ \\
Synthetic CSW & $0.74_{[.70,.77]}$ & $0.77_{[.74,.81]}$ & $0.76_{[.72,.80]}$ & $0.30_{[.26,.34]}$ & $0.31_{[.27,.35]}$ & $0.42_{[.38,.47]}$ & -- \\
Translated prompts & $0.73_{[.69,.77]}$ & $0.76_{[.72,.79]}$ & $0.77_{[.73,.80]}$ & $0.32_{[.28,.36]}$ & $0.30_{[.27,.35]}$ & $0.37_{[.33,.41]}$ & $0.27_{[.24,.31]}$ \\
\multicolumn{8}{l}{\textit{Global PIQA} --- DeepSeek-R1-Distill-Llama-8B ($n=100$ per cell)} \\
No-SFT baseline & $0.80_{[.71,.87]}$ & $0.84_{[.76,.90]}$ & $0.73_{[.64,.81]}$ & $0.50_{[.40,.60]}$ & $0.16_{[.10,.24]}$ & $0.49_{[.39,.59]}$ & $0.50_{[.40,.60]}$ \\
English reasoning & $0.71_{[.61,.79]}$ & $0.80_{[.71,.87]}$ & $0.70_{[.60,.78]}$ & $0.40_{[.31,.50]}$ & $0.24_{[.17,.33]}$ & $0.44_{[.35,.54]}$ & $0.48_{[.38,.58]}$ \\
Machine translation & $0.54_{[.44,.63]}$ & $0.78_{[.69,.85]}$ & $0.73_{[.64,.81]}$ & $0.42_{[.33,.52]}$ & $0.46_{[.37,.56]}$ & $0.43_{[.34,.53]}$ & $0.36_{[.27,.46]}$ \\
Native reasoning & $0.68_{[.58,.76]}$ & $0.80_{[.71,.87]}$ & $0.72_{[.63,.80]}$ & $0.43_{[.34,.53]}$ & $0.23_{[.16,.32]}$ & $0.47_{[.38,.57]}$ & -- \\
Strategic CSW & $0.79_{[.70,.86]}$ & $0.77_{[.68,.84]}$ & $0.76_{[.67,.83]}$ & $0.39_{[.30,.49]}$ & $0.24_{[.17,.33]}$ & $0.52_{[.42,.62]}$ & $0.46_{[.37,.56]}$ \\
Synthetic CSW & $0.75_{[.66,.82]}$ & $0.80_{[.71,.87]}$ & $0.76_{[.67,.83]}$ & $0.44_{[.35,.54]}$ & $0.25_{[.18,.34]}$ & $0.51_{[.41,.61]}$ & -- \\
Translated prompts & $0.77_{[.68,.84]}$ & $0.82_{[.73,.88]}$ & $0.70_{[.60,.78]}$ & $0.35_{[.26,.45]}$ & $0.31_{[.23,.41]}$ & $0.43_{[.34,.53]}$ & $0.42_{[.33,.52]}$ \\
\multicolumn{8}{l}{\textit{Global PIQA} --- Phi-4-reasoning ($n=100$ per cell)} \\
No-SFT baseline & $0.94_{[.88,.97]}$ & $0.95_{[.89,.98]}$ & $0.83_{[.74,.89]}$ & $0.52_{[.42,.62]}$ & $0.52_{[.42,.62]}$ & $0.81_{[.72,.87]}$ & $0.50_{[.40,.60]}$ \\
English reasoning & $0.93_{[.86,.97]}$ & $0.95_{[.89,.98]}$ & $0.86_{[.78,.91]}$ & $0.54_{[.44,.63]}$ & $0.53_{[.43,.62]}$ & $0.83_{[.74,.89]}$ & $0.49_{[.39,.59]}$ \\
Machine translation & $0.94_{[.88,.97]}$ & $0.95_{[.89,.98]}$ & $0.84_{[.76,.90]}$ & $0.54_{[.44,.63]}$ & $0.68_{[.58,.76]}$ & $0.83_{[.74,.89]}$ & $0.54_{[.44,.63]}$ \\
Native reasoning & $0.96_{[.90,.98]}$ & $0.93_{[.86,.97]}$ & $0.90_{[.83,.94]}$ & $0.58_{[.48,.67]}$ & $0.63_{[.53,.72]}$ & $0.84_{[.76,.90]}$ & -- \\
Strategic CSW & $0.95_{[.89,.98]}$ & $0.94_{[.88,.97]}$ & $0.84_{[.76,.90]}$ & $0.58_{[.48,.67]}$ & $0.63_{[.53,.72]}$ & $0.83_{[.74,.89]}$ & $0.59_{[.49,.68]}$ \\
Synthetic CSW & $0.94_{[.88,.97]}$ & $0.95_{[.89,.98]}$ & $0.89_{[.81,.94]}$ & $0.46_{[.37,.56]}$ & $0.53_{[.43,.62]}$ & $0.84_{[.76,.90]}$ & -- \\
Translated prompts & $0.94_{[.88,.97]}$ & $0.96_{[.90,.98]}$ & $0.81_{[.72,.87]}$ & $0.54_{[.44,.63]}$ & $0.59_{[.49,.68]}$ & $0.82_{[.73,.88]}$ & $0.53_{[.43,.62]}$ \\
\multicolumn{8}{l}{\textit{Global PIQA} --- Qwen3-8B ($n=100$ per cell)} \\
No-SFT baseline & $0.85_{[.77,.91]}$ & $0.90_{[.83,.94]}$ & $0.84_{[.76,.90]}$ & $0.63_{[.53,.72]}$ & $0.56_{[.46,.65]}$ & $0.73_{[.64,.81]}$ & $0.48_{[.38,.58]}$ \\
English reasoning & $0.86_{[.78,.91]}$ & $0.89_{[.81,.94]}$ & $0.80_{[.71,.87]}$ & $0.58_{[.48,.67]}$ & $0.56_{[.46,.65]}$ & $0.66_{[.56,.75]}$ & $0.53_{[.43,.62]}$ \\
Machine translation & $0.70_{[.60,.78]}$ & $0.89_{[.81,.94]}$ & $0.83_{[.74,.89]}$ & $0.58_{[.48,.67]}$ & $0.51_{[.41,.61]}$ & $0.67_{[.57,.75]}$ & $0.32_{[.24,.42]}$ \\
Native reasoning & $0.87_{[.79,.92]}$ & $0.89_{[.81,.94]}$ & $0.79_{[.70,.86]}$ & $0.60_{[.50,.69]}$ & $0.49_{[.39,.59]}$ & $0.72_{[.63,.80]}$ & -- \\
Strategic CSW & $0.87_{[.79,.92]}$ & $0.89_{[.81,.94]}$ & $0.82_{[.73,.88]}$ & $0.59_{[.49,.68]}$ & $0.56_{[.46,.65]}$ & $0.71_{[.61,.79]}$ & $0.53_{[.43,.62]}$ \\
Synthetic CSW & $0.86_{[.78,.91]}$ & $0.90_{[.83,.94]}$ & $0.85_{[.77,.91]}$ & $0.60_{[.50,.69]}$ & $0.53_{[.43,.62]}$ & $0.75_{[.66,.82]}$ & -- \\
Translated prompts & $0.89_{[.81,.94]}$ & $0.89_{[.81,.94]}$ & $0.79_{[.70,.86]}$ & $0.62_{[.52,.71]}$ & $0.50_{[.40,.60]}$ & $0.74_{[.65,.82]}$ & $0.45_{[.36,.55]}$ \\
\bottomrule
\end{longtable}
}

\subsection{Accuracy on English subsets of Global PIQA and \encore corpus SFT test set (curated from Global MMLU)}
English scores for the No-SFT baseline (no fine-tuned condition has been evaluated on English). 

{\footnotesize
\begin{table}[h]
\centering
\begin{tabular}{lcc}
\toprule
Model & \encore Global MMLU test set & Global PIQA \\
\midrule
DeepSeek-R1-Distill-Llama-8B & $0.70_{[.65,.73]}$ & $0.76_{[.67,.83]}$ \\
Phi-4-reasoning & $0.83_{[.79,.86]}$ & $0.89_{[.81,.94]}$ \\
Qwen3-8B & $0.83_{[.79,.86]}$ & $0.89_{[.81,.94]}$ \\
\bottomrule
\end{tabular}
\caption{Accuracy for English, no-SFT baseline only. Subscripts are Wilson 95\% confidence intervals.}
\label{tab:english-accuracy-appendix}
\end{table}
}

\subsection{Supervised fine-tuning task dataset synthesis details}\label{sec:synthesis}
\emph{Native-language reasoning.} Given a prompt $p_l$ in language $l$, the model is fine-tuned to produce its reasoning $r_l$ and final answer $a_l$ in the same language $l$, yielding $(p_l, r_l, a_l)$. Given some $p_l$, we synthesize $r_l$ and $a_l$ by prompting Qwen3-Next-80B-A3B-Thinking with the equivalent English prompt $p_e$ from Global MMLU (see section \ref{sec:datasets}). We filter for instances with correct answers using the reference answers in Global MMLU. To ensure efficient reasoning, we then filter correct reasoning demonstrations by length (in number of output tokens), using a model- and language-specific cutoff.\footnote{We compute the 95th percentile of output lengths (in tokens) for correct reasoning instances from each baseline (non\textendash fine-tuned) reasoning model on the validation set in each language.} Finally, we translate the English reasoning $r_e$ and answers $a_e$ into the target language to yield $r_l$ and $a_l$. We source the prompt $p_l$ directly from Global MMLU.

\emph{Machine translation into English.} We fine-tune our models to perform translation from each of the languages in section \ref{sec:datasets} into English, sourcing examples from NLLB. The fine-tuning instances for this task consist of a prompt $p_l$ instructing the model to translate an example non-English sentence into English, followed by empty reasoning $r$ and the English translation of the sentence as the final answer $a_e$: $(p_l, r, a_e)$.

\emph{Reasoning prompt translation into English.} We use non-English prompts and their English translations from Global MMLU (see section \ref{sec:datasets}) as examples for fine-tuning. More formally, the fine-tuning instances for this task consist of a prompt $p_l$ instructing the model to translate an example non-English reasoning prompt into English, followed by empty reasoning $r$ and the English translation of the prompt as the final answer $a_e$: $(p_l, r, a_e)$.

\emph{English-language reasoning.} Given a prompt $p_l$ in language $l$, the model is fine-tuned to generate English-language reasoning $r_e$ before returning its final answer $a_l$ in the same language $l$ as the prompt: $(p_l, r_e, a_l)$.

\emph{Strategically code-switched reasoning.} This condition fine-tunes the student reasoning models on examples of code-switched reasoning from Qwen3-Next-80B-A3B-Thinking. Given prompt $p_l$ in language $l$, we generate candidate code-switched reasoning demonstrations $r_{csw}$ and final answers $a_l$ in language $l$ by prompting Qwen3-Next-80B-A3B-Thinking with $p_l$. Next, we filter for instances where multiple languages are present in $r_{csw}$. We detect code-switched reasoning instances using Qwen3-Next-80B-A3B-Instruct and the prompt from appendix \ref{sec:instance-lid} \citep{yang2025qwen3technicalreport}. This yields fine-tuning instance $(p_l, r_{csw}, a_l)$.

\emph{Synthetically code-switched reasoning.} Given prompt $p_l$ in language $l$, this task also fine-tunes the student reasoning models on examples of code-switched reasoning $r_{csw}$ to produce final answers $a_l$ in language $l$. For $p_l$ in language $l$, we generate reasoning and answers in both English and $l$ ($r_e, a_e, r_l, a_l$) using the same approach as for the \emph{native-language reasoning} task. We use newlines to segment the reasoning into steps as done in \citet{chen2025sealsteerablereasoningcalibration}, then randomly select half of these steps to be in English, with the other half remaining in language $l$. We splice together the final code-switched reasoning trace $r_{csw}$ from $r_e$ and $r_l$, yielding fine-tuning instance $(p_l, r_{csw}, a_l)$.

\subsection{Full forest plots (pooled, Global MMLU, and Global PIQA) for association between code-switching and final answer accuracy}\label{sec:appendix-3panel}
\begin{figure}[htbp]
    \centering
        \includegraphics[width=\linewidth]{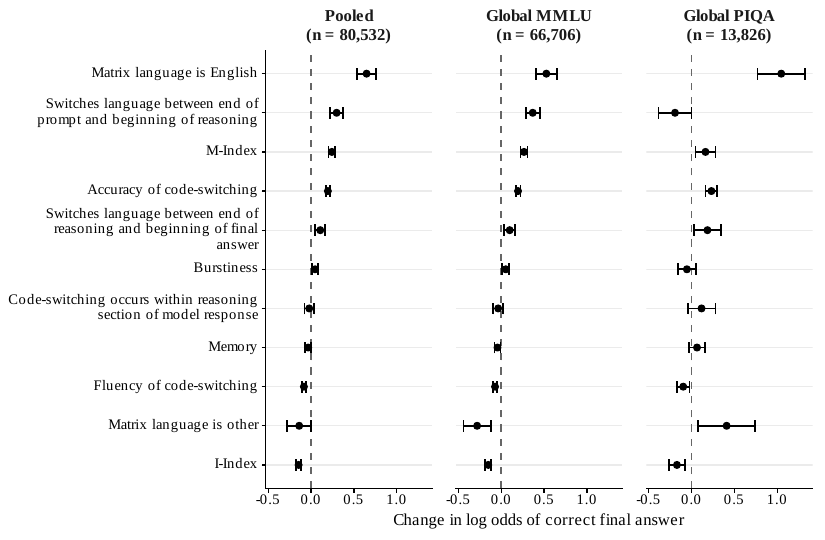}
    
     \caption{Across all evaluated models, languages, and datasets, certain code-switching behaviors in reasoning are associated with higher odds of a correct final answer. In particular, featuring English as the matrix language, finishing the reasoning in a different language before switching back to the target language for the final answer, higher accuracy of code-switching, and higher M-Index are all significantly associated with better reasoning performance.}\label{fig:3panel}
\end{figure}

Figure \ref{fig:3panel} shows the relationship between code-switching behaviors and final answer correctness for evaluations on Global MMLU, Global PIQA, and both datasets pooled.

\subsection{Full heatmaps (pooled, Global MMLU, and Global PIQA)}
\begin{figure}[t]
    \centering
        \includegraphics[width=\linewidth]{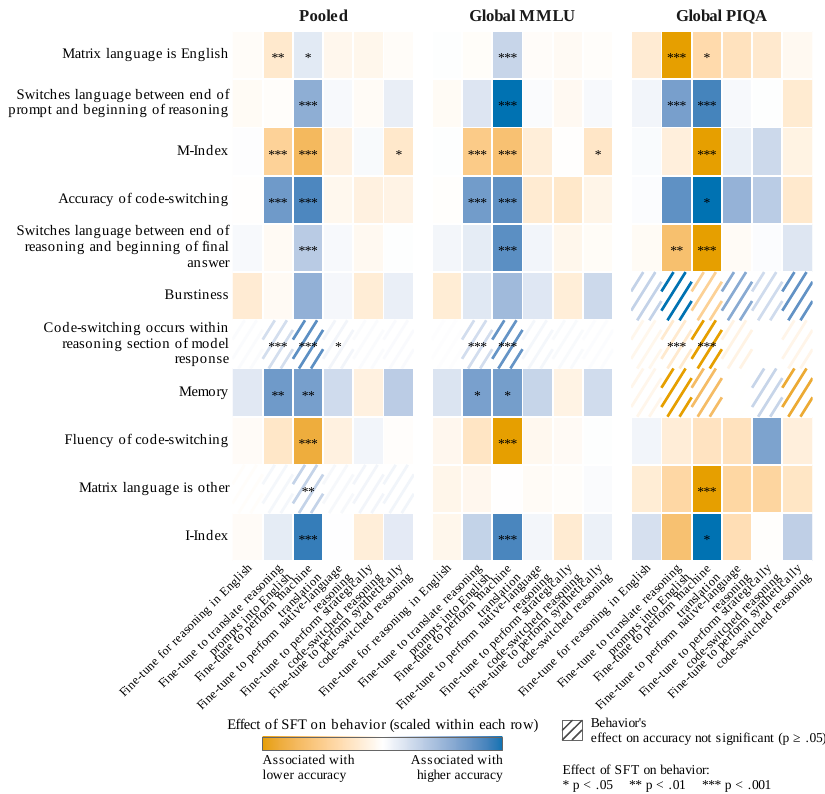}
     \caption{Across all models, languages, and datasets evaluated, our translation-based SFT intervention tasks (translating reasoning prompts into English and performing open-domain machine translation into English) emerged as useful approaches for promoting code-switched reasoning behaviors associated with higher final answer accuracy.}
    \label{fig:3panel-heatmap}
\end{figure}

Figure \ref{fig:3panel-heatmap} shows the effect of our SFT interventions on the Global MMLU and Global PIQA datasets separately, in addition to the pooled view shown in the main paper text.

\subsection{Gemma teacher generalization experiments}\label{sec:gemma}
\begin{figure}[th]
    \centering
        \includegraphics[width=\linewidth]{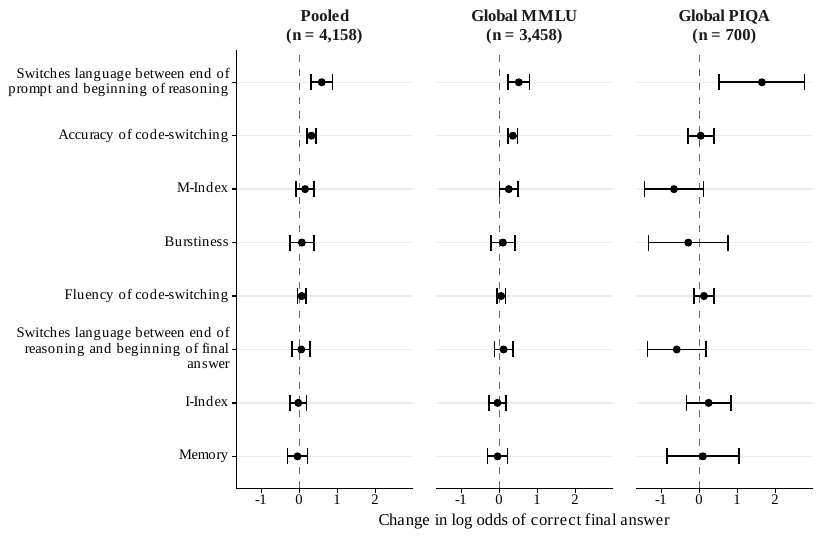}
    \caption{In addition to evaluating models fine-tuned on reasoning demonstrations from Qwen3-Next-80B-A3B-Thinking, our results also show that our approach generalizes to other teacher models (here, gemma-4-26B-A4B-it). We see that in the reasoning of models fine-tuned on gemma-4-26B-A4B-it reasoning demonstrations, certain key code-switched reasoning behaviors (e.g., accuracy of code-switching) remain associated with better performance, in line with our results from the Qwen3-Next-80B-A3B-Thinking teacher experiments. }
    \label{fig:3panel-gemma}
\end{figure}

\begin{figure}[th]
    \centering
        \includegraphics[width=\linewidth]{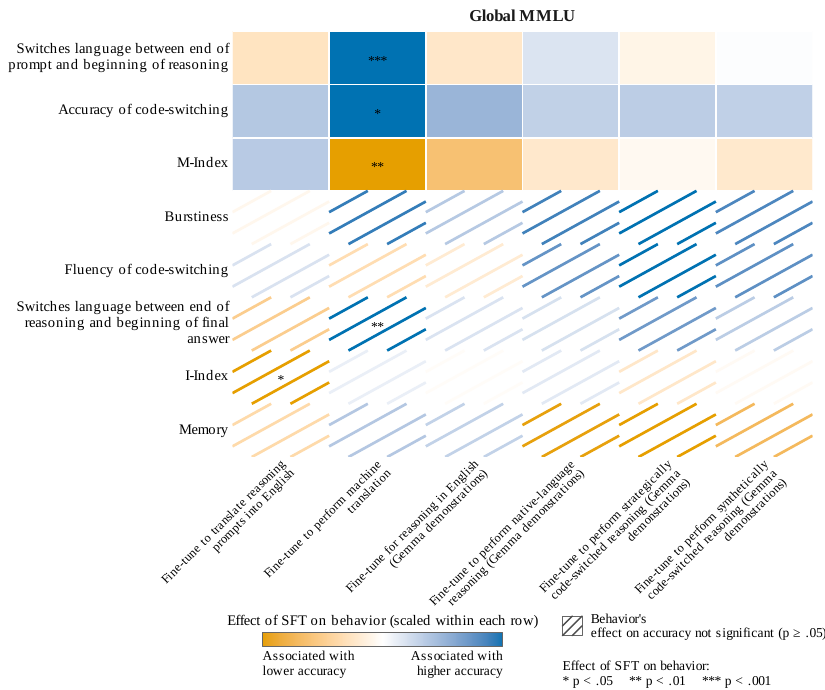}
    \caption{When using gemma-4-26B-A4B-it as the teacher model instead of Qwen3-Next-80B-A3B-Thinking, we see that our SFT interventions still significantly boost certain helpful code-switched reasoning behaviors.}
    \label{fig:3panel-heatmap-gemma}
\end{figure}

Figures \ref{fig:3panel-gemma} and \ref{fig:3panel-heatmap-gemma} show that our framework generalizes across teacher models. 

\subsection{Modeling reasoning task response correctness for computing average marginal effects}\label{sec:ame}
Let $y_{i} \in \{0,1\}$ indicate whether response $i$,
given by base model $m$ under condition $c$ to prompt $p$ in language $\ell$ from
dataset $d$, is correct. We fit $\operatorname{logit} \Pr(y_{i} = 1) = \beta_0 + \delta_{d} + \mu_{m} + \gamma_{c} + \lambda_{\ell} + (\gamma\mu)_{cm} + (\gamma\lambda)_{c\ell} + u_{q}$, where $u_{q} \sim \mathcal{N}(0, \sigma_u^2)$. We include fixed effects for dataset, base model, condition (reference level is the baseline reasoning model without fine-tuning), and prompt language; condition-by-model and
condition-by-language interactions; and a random intercept $u_q$ for each question.

\subsection{Fluency judge validation}\label{sec:switch-lingua}

We validate our code-switching fluency judge against scores on a random sample from
SwitchLingua\footnote{\url{https://huggingface.co/datasets/Shelton1013/SwitchLingua_text}}.
We compute Spearman's correlation and Kendall's correlation between our fluency scores and those from SwitchLingua (Table \ref{tab:fl-validation-correlations}).

We use the SwitchLingua language pairs that overlap with our study: Arabic, Chinese, French, German, Hindi, Italian, Japanese, Korean, Russian, and Spanish, each code-switched with English. For each instance, SwitchLingua provides a \texttt{fluency\_score} on a 0--10 scale.
For each language we randomly sample 500 instances with a fixed seed for reproducibility, stratifying to cover the full distribution of possible fluency scores. This yields 5{,}000 text instances.

Each sampled text is scored once by our fluency judge, with the same configuration that produces the fluency scores in our main experiments. We use \texttt{Qwen/Qwen3.5-27B} in non-reasoning mode and prompt with a three-level rubric (see appendix \ref{sec:fl-eval}). Decoding uses temperature 0.7, top-$p$ 0.8, top-$k$ 20, and a presence
penalty of 1.5.

For each language, and for all languages pooled, we compute Spearman's $\rho$
and Kendall's $\tau_b$ between our judged score and the SwitchLingua reference score. We use rank correlations because the two scales differ and $\tau_b$ to adjust for fluency score ties. We report 95\% confidence
intervals, obtained by computing the correlation on $1,000$ samples ($n=5,000$ for the full SwitchLingua subset used here, or $n=500$ for per-language estimates) of (judge score, reference score) pairs with replacement. We then take the 2.5th and 97.5th percentiles of the resampled statistic as our confidence interval bounds. We also report two-sided $p$-values
against the null hypothesis of no association between our scores and SwitchLingua.

\begin{table}[t]                          
\centering                                                                        \small
\setlength{\tabcolsep}{4pt}
\begin{tabular}{@{}lrcrcrrr@{}}
\toprule
Language & $n$ & Spearman $\rho$ [95\% CI] & $p$ & Kendall $\tau_b$ [95\% CI] & $p$ &
\shortstack[r]{Our judge\\mean score\\min score=1\\max score=3} &
\shortstack[r]{SwitchLingua\\reference\\mean score\\min score=0\\max score=10} \\
 \midrule
 Arabic   & 500 & 0.430 [0.348, 0.506] & $<$0.001 & 0.381 [0.311, 0.448] & $<$0.001 & 2.700 & 8.612 \\
 Chinese  & 500 & 0.369 [0.278, 0.444] & $<$0.001 & 0.325 [0.245, 0.391] & $<$0.001 & 2.718 & 8.811 \\
 French   & 500 & 0.472 [0.393, 0.540] & $<$0.001 & 0.417 [0.348, 0.477] & $<$0.001 & 2.684 & 8.874 \\
 German   & 500 & 0.542 [0.472, 0.611] & $<$0.001 & 0.476 [0.416, 0.537] & $<$0.001 & 2.522 & 8.680 \\
 Hindi    & 500 & 0.332 [0.253, 0.408] & $<$0.001 & 0.301 [0.229, 0.370] & $<$0.001 & 2.738 & 8.485 \\
 Italian  & 500 & 0.510 [0.436, 0.584] & $<$0.001 & 0.447 [0.384, 0.511] & $<$0.001 & 2.616 & 8.709 \\
 Japanese & 500 & 0.504 [0.432, 0.568] & $<$0.001 & 0.440 [0.378, 0.498] & $<$0.001 & 2.664 & 8.825 \\
 Korean   & 500 & 0.365 [0.284, 0.444] & $<$0.001 & 0.323 [0.251, 0.393] & $<$0.001 & 2.804 & 8.918 \\
 Russian  & 500 & 0.464 [0.383, 0.540] & $<$0.001 & 0.404 [0.334, 0.472] & $<$0.001 & 2.458 & 8.488 \\
 Spanish  & 500 & 0.402 [0.320, 0.484] & $<$0.001 & 0.355 [0.283, 0.426] & $<$0.001 & 2.700 & 8.850 \\
 \midrule
 All      & 5{,}000 & 0.445 [0.421, 0.469] & $<$0.001 & 0.391 [0.370, 0.413] & $<$0.001 & 2.660 & 8.725 \\
 \bottomrule
 \end{tabular}
 \caption{Rank correlation between our fluency judge's scores (1--3) and the
 SwitchLingua reference fluency scores (0--10), per language and pooled. Correlations are reported with 95\% confidence intervals. $p$-values are
 two-sided.}
 \label{tab:fl-validation-correlations}
 \end{table}

\subsection{Exploratory data analysis on BERTopic representations for taxonomy design}\label{appendix:eda}
\begin{figure}[htbp]
    \centering
    \includegraphics[width=\textwidth, trim=0 2000px 0 0, clip]{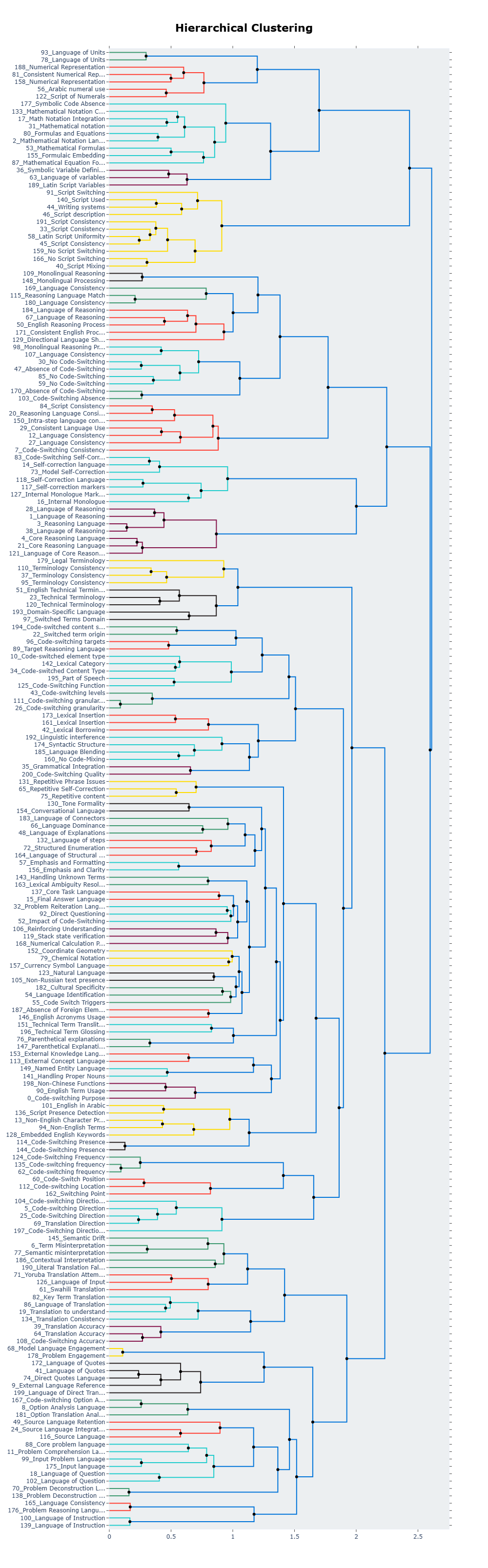}
    \caption{Part 1 of dendrogram plot of topic representations from BERTopic pipeline (see appendix \ref{sec:bertopic} and section \ref{sec:taxonomy-approach}).}\label{fig:dendro1}
\end{figure}

Figures \ref{fig:dendro1} \textendash \ref{fig:dendro3} show a dendrogram representation of the topics used to build the code-switched reasoning behavior taxonomy.

\begin{figure}[htbp]
    \centering
    \includegraphics[width=\textwidth, trim=0 1000px 0 1215px, clip]{newplot.png}
    \caption{Part 2 of dendrogram plot of topic representations from BERTopic pipeline (see appendix \ref{sec:bertopic} and section \ref{sec:taxonomy-approach}).}
\end{figure}

\begin{figure}[htbp]
    \centering
    \includegraphics[width=\textwidth, trim=0 0 0 2215px, clip]{newplot.png}
    \caption{Part 3 of dendrogram plot of topic representations from BERTopic pipeline (see appendix \ref{sec:bertopic} and section \ref{sec:taxonomy-approach}).}\label{fig:dendro3}
\end{figure}

\subsection{Full taxonomy of code-switched reasoning behaviors}\label{sec:full_taxonomy}
The taxonomy is shown in Table \ref{tab:code_switch_schema}.
\begin{table}[htbp]
\centering
\footnotesize
\caption{Taxonomy of code-switched reasoning.}
\label{tab:code_switch_schema}
\resizebox{\textwidth}{!}{%
\begin{tabular}{|p{2cm}|p{8.5cm}|p{3cm}|}
\hline
\textbf{Dimension} & \textbf{Name:} Description & \textbf{Granularity} \\
\hline

\multirow{4}{*}{Function} 
& \textbf{Translation and Interpretation}: Translates or interprets foreign language phrases into the primary language of reasoning to understand meaning. & Step-level\\
& \quad \textbf{Lexical Ambiguity Resolution}: Identifies unclear terms in one language and clarifies meaning in another. & Step-level  \\
& \quad \textbf{Parenthetical Translation}: Provides a direct translation of a single word or short phrase in parentheses, often immediately after the term appears, reinforcing the precise meaning of a concept for clarity and accuracy.  & Step-level  \\
\cline{2-3}
& \textbf{Lexical Retention}: Retains words/phrases from the original language of the prompt, without translating them to the primary language of reasoning. & Step-level  \\
\hline

\multirow{15}{*}{Form} 
& \textbf{Directionality of Code-switching}: Whether the code-switching is from the primary language of reasoning to a second language, or vice-versa.  & Switch-level  \\
\cline{2-3}
& \textbf{Scope of Code-switching}: Whether the code-switching occurs at the level of individual words, phrases, sentences, or entire reasoning steps. & Word-/Phrase-/Sentence-/Step-/Instance-level  \\
\cline{2-3}
& \textbf{Language-Role Mapping}: Which language is used for different aspects of the reasoning process. & Instance-level \\
& \quad \textbf{Language of Core Reasoning}: The primary language used for the logical progression and problem-solving steps of the reasoning process. The ``matrix language'' of the reasoning. & Instance-level  \\
& \quad \textbf{Language of Internal Monologue}: The language in which the model uses conversational phrases and self-correction markers common in human thought processes, such as ``Okay, so...'', ``Let's see.'', ``Wait,'', ``Hmm.'', ``No.'', ``Alright,'' to structure its reasoning and signal its thought progression. & Instance-level  \\
& \quad \textbf{Language of Problem Deconstruction}: The language used when breaking down or rephrasing parts of the original problem statement or options. & Instance-level  \\
& \quad \textbf{Language of Self-Correction}: The language used by the model during self-correction or re-evaluation of its understanding. & Instance-level  \\
& \quad \textbf{Language of External Knowledge Retrieval}: The language used when recalling or integrating external information or general knowledge relevant to the problem. & Instance-level  \\
\cline{2-3}
& \textbf{Language-Content Mapping}: Maps how different languages are used for different content types. & Step-level \\
& \quad \textbf{Language of Direct Quotes}: Language used when quoting original problem text. & Step-level  \\
& \quad \textbf{Language of Technical Terminology}: Language for domain-specific terms and concepts. & Step-level  \\
& \quad \textbf{Language of Final Answer}: Language used to state the final answer or conclusion. & Step-level \\
\cline{2-3}
& \textbf{Interleaving Natural and Non-natural Language Elements}: Mixing elements such as code and math into natural language. & Step-level \\
& \quad \textbf{Interleaving Math and Natural Language}: Integration of mathematical symbols, equations, or expressions directly within natural language sentences or phrases. & Step-level  \\
& \quad \textbf{Interleaving Code and Natural Language}: Integration of elements such as code snippets, programming language constructs, or pseudocode directly within natural language sentences or phrases. & Step-level  \\
\hline

\multirow{2}{*}{Coherence} 
& \textbf{Fluency of Code-Switching}: Assesses how natural and understandable code-switching is, considering aspects such as grammar and syntax. & Instance-level  \\
\cline{2-3}
& \textbf{Accuracy of Code-Switching}: Assesses whether code-switching uses semantically appropriate terms for the context. & Instance-level  \\
\cline{2-3}
& \textbf{Consistency of Terminology}: Whether specific terms, once introduced in one language, are consistently maintained in that language or switch back and forth.  & Instance-level  \\
\hline
\end{tabular}%
}
\end{table}

\subsection{Example topic representations}
Table \ref{tab:topic-examples} shows examples of topic representations from the BERTopic pipeline, their refined labels generated by Gemini 2.5 Flash, and the corresponding taxonomy dimensions/categories informed by that topic.
\begin{table}[h]
\caption{Example topics from the BERTopic pipeline. For each topic we show its size ($n$, number of criteria), its top c-TF-IDF keywords, the label generated by Gemini 2.5 Flash, a representative criterion, and the taxonomy element it informed.}
\label{tab:topic-examples}
\centering
\small
\begin{tabular}{@{}r p{1.5cm} p{2.5cm} p{4cm} p{3cm}@{}}
\toprule
Topic ID \# & c-TF-IDF keywords & Gemini label & Text of representative document (``document'' here means code-switching criterion) & Corresponding taxonomy element \\
\midrule
5 & direction, occurs, vice, versa & Code-switching Direction
    & Direction of Code-Switching (Describes the direction of the switch, e.g., from reasoning language to problem language, or vice-versa.)
    & \textbf{Form:} Directionality of Code-switching \\
\addlinespace
26 & granularity, extent, size, length, segment
    & Code-switching Granularity
    & Granularity of Code-Switching (The size or extent of the switched segment, e.g., single words, phrases, full sentences.)
    & \textbf{Form:} Scope of Code-switching \\
\addlinespace
19 & translates, translation, meaning, aid, understand & Translation to Understand & Direct Translation and Interpretation (The model translates Korean phrases into English to understand their meaning in the context of the question.) & \textbf{Function:} Translation and Interpretation\\
\addlinespace
9 & external, quoting, referring, quoted & External Language Reference &  Language of External Problem Reference (The language used when directly quoting or referring to parts of the original problem statement.) & \textbf{Function:} Lexical Retention \\
\addlinespace
24 & retention, retains, preservation, lexical, original & Source Language Retention & Lexical Retention (Instances where the model retains specific words or phrases from the original language of the problem within its English reasoning, without translating them.) &  \textbf{Function:} Lexical Retention\\
35 & integrated, grammatically, integration
    & Grammatical Integration
    & Grammatical Integration (Assesses how well the switched language elements are integrated grammatically into the surrounding text.)
    & \textbf{Coherence:} Fluency of Code-Switching \\
\addlinespace
39 & accuracy, accurately, translation
    & Translation Accuracy
    & Accuracy of Translation/Interpretation (This assesses how accurately the model translates or interprets the Swahili text into the language of its core reasoning.)
    & \textbf{Coherence:} Accuracy of Code-Switching \\
\bottomrule
\end{tabular}
\end{table}

\subsection{Details of statistical modeling from section \ref{sec:taxonomy-approach}}\label{appendix:regression}
For reasoning trace $t$ answering prompt $p$ in language $l$, $\text{logit}\,P(\text{correct}_t = 1) = \beta_0 + \beta_{\text{dataset}} + \beta_{\text{model}} + \sum_k \beta_k\,\text{boundary}_{k,t} + \sum_m \gamma_m\, c_t\, z(\text{metric}_{m,t}) + \beta_{\text{matrix\_lang}} + u_\ell + v_p$, where $u_\ell \sim \mathcal{N}(0,\sigma^2_\ell)$ and $v_p \sim \mathcal{N}(0,\sigma^2_p)$, and $c_t$ indicates whether $t$ features code-switching within the reasoning trace. We drop the CMI because it is collinear with the M-Index.

\subsection{Details of statistical modeling for relationship between code-switching and performance in section \ref{sec:sft-results} (Figure \ref{fig:1panel})}\label{appendix:section4-csw-perf}
For reasoning trace $t$, prompt $p$, and language $\ell$, $\text{logit}\,P(\text{correct}_t = 1) = \beta_0 + \beta_{\text{dataset}} + \beta_{\text{model}} + \beta_{\text{train}}\, z(\text{train\_exs}_t) + \sum_k \beta_k\,\text{boundary}_{k,t} + \sum_m \gamma_m\, c_t\, z(\text{metric}_{m,t}) + \beta_{\text{mt}}\, z(\text{mt}_\ell) + \beta_{\text{matrix\_lang}} + u_\ell + v_p$, where $u_\ell \sim \mathcal{N}(0,\sigma^2_\ell)$ and $v_p \sim \mathcal{N}(0,\sigma^2_p)$. $c_t$ indicates whether $t$ features code-switching within the reasoning trace. $\text{train\_exs}_t$ is the number of SFT examples used. $\text{mt}_\ell$ is the MT quality for $\ell$ reported by \citet{communication2023seamlessmultilingualexpressivestreaming} and \citet{nllbteam2022languageleftbehindscaling}. We drop CMI from the regression because it is collinear with M-Index based on the adjusted GVIF.

\subsection{Mediation analysis}\label{appendix:mediation}

For the mediation analysis, we fit regressions for the relationships $X \rightarrow Y$, $X \rightarrow M$, and $X + M \rightarrow Y$. The treatment $X$ is the task a given model has been fine-tuned on, the mediator $M$ is the set of code-switching behaviors in an instance of reasoning from the model, and the outcome $Y$ is whether the final answer generated from that reasoning is correct.

\textbf{Data.} The analysis uses the full \encore test set (a parallel set of 500 questions from Global MMLU translated into each target language and 100 non-parallel questions from Global PIQA per language). All SFT tasks and reasoning models from the section \ref{sec:sft} experiments (DeepSeek-R1-Distill-Llama-8B, Phi-4-reasoning, Qwen3-8B) are evaluated. We also evaluate the baseline non-fine-tuned models. This gives 80{,}532 reasoning instances.

\textbf{Statistical models.} We model the relationships $X \rightarrow Y$ and $X + M \rightarrow Y$ using logistic mixed-effects regressions with crossed random intercepts for language and question. The regression for $X \rightarrow Y$ predicts final answer correctness using the following predictors: the \encore subset (Global MMLU or Global PIQA), the model, and the SFT condition. 

The regression for $X + M \rightarrow Y$ adds the code-switching behaviors $M$ as predictors: indicator variables for a code-switch at the prompt-reasoning or reasoning\textendash final answer boundary, CMI, M-Index, I-Index, burstiness, memory, code-switching fluency, code-switching accuracy, and matrix language. We also control for the language-specific quality of English-to-target machine translation used in data synthesis.

\textbf{Decomposition of total effect into direct and mediated effects.} Let $\hat{p}_q(c, m)$ be the probability that a model fine-tuned in SFT task condition $c$ answers question $q$ correctly when its reasoning includes code-switching behaviors $m$ (given by the regression predicting $X + M \rightarrow Y$). Then let $S_c$ be the set of test responses $i = (q, \ell, k)$ (question, language,
base model) for which condition $c$ was evaluated, and let $m_i(c)$ denote the
code-switching behaviors observed in response $i$ under condition $c$. Then
\begin{align*}
p_{\text{ref}}(c)    &= \frac{1}{|S_c|}\sum_{i \in S_c} \hat p_i\big(0,\, m_i(0)\big), \\
p_{\text{direct}}(c) &= \frac{1}{|S_c|}\sum_{i \in S_c} \hat p_i\big(c,\, m_i(0)\big), \\
p_{\text{focal}}(c)  &= \frac{1}{|S_c|}\sum_{i \in S_c} \hat p_i\big(c,\, m_i(c)\big).
\end{align*}
Here $p_{\text{direct}}$ is the reasoning task accuracy predicted by our regression if the SFT task condition were applied while holding fixed the code-switching behaviors observed from the baseline non-fine-tuned model.

The effects are
\begin{align*}
\text{Total effect} &= p_{\text{focal}} - p_{\text{ref}}, \\
\text{Average Direct Effects (ADE)} &= p_{\text{direct}} - p_{\text{ref}}, \\
\text{Average Causal Mediation Effects (ACME)} &= p_{\text{focal}} - p_{\text{direct}},
\end{align*}
so that the total effect is exactly the sum of the direct and mediated effects.  

We estimate the joint mediation effect of the code-switching behaviors for each SFT task condition. Estimating single mediation effects would require considering impossible outcomes. For example, an I-Index of $0$ indicates an absence of any code-switching, in which case code-switching fluency and code-switching accuracy would be undefined.

\textbf{Uncertainty quantification.} We report 95\% bootstrap confidence intervals computed by cluster bootstrapping with 500 resampling steps. Each sample draws 1,200 test set questions with replacement, keeps all responses to the sampled questions, refits the regression $\hat{p}_q(c, m)$, and recomputes the total effect, ADE, and ACME. The interval bounds are the 2.5th and 97.5th percentiles of these 500 results.

\begin{table}[!htbp]
\centering
\tiny
\begin{tabular}{lrrr}
\toprule
SFT task demonstrations & Total effect & ADE [95\% CI] & ACME [95\% CI] \\
\midrule
Reasoning in English & $-$0.35 [$-$1.18, 0.53] & $-$0.39 [$-$1.21, 0.47] & 0.04 [$-$0.05, 0.14] \\
Translating reasoning prompts into English & $-$0.14 [$-$1.07, 0.74] & $-$0.25 [$-$1.15, 0.62] & 0.10 [$-$0.00, 0.22] \\
Open-domain machine translation into English & $-$0.07 [$-$1.03, 0.79] & $-$0.52 [$-$1.50, 0.34] & \textbf{0.46 [0.24, 0.70]} \\
Native-language reasoning & 0.07 [$-$0.75, 1.04] & 0.02 [$-$0.80, 0.95] & 0.05 [$-$0.06, 0.16] \\
Strategically code-switched reasoning & 0.16 [$-$0.68, 1.17] & 0.17 [$-$0.65, 1.13] & $-$0.01 [$-$0.10, 0.08] \\
Synthetically code-switched reasoning & 0.25 [$-$0.68, 1.13] & 0.30 [$-$0.63, 1.17] & $-$0.04 [$-$0.13, 0.05] \\
\bottomrule
\end{tabular}
\caption{Effect of each SFT intervention on reasoning task accuracy relative to the baseline non\textendash fine-tuned reasoning model, mediated by code-switching behaviors (measured in percentage points).}
\label{tab:mediation}
\end{table}

\textbf{Results of mediation analysis.} Table~\ref{tab:mediation} gives the estimated total effect, ADE, and ACME for each SFT task. Fine-tuning on demonstrations of open-domain machine translation into English has positive mediated effects on reasoning task accuracy.

\begin{table}[t]
\centering
\tiny
\setlength{\tabcolsep}{5pt}
\begin{tabular}{lrrr}
\toprule
 & Reason in English & Translate reasoning prompts to English & Open-domain MT to English\\
\midrule
\multicolumn{4}{l}{\textit{Languages}} \\
\cellcolor{resHigh}\quad Hindi & \textbf{0.25 [0.01, 0.49]} & 0.18 [$-$0.08, 0.43] & \textbf{$-$0.78 [$-$1.14, $-$0.40]} \\
\cellcolor{resHigh}\quad Indonesian & 0.20 [$-$0.03, 0.45] & 0.18 [$-$0.04, 0.42] & $-$0.02 [$-$0.31, 0.28] \\
\cellcolor{resHigh}\quad Malay & $-$0.09 [$-$0.30, 0.15] & 0.06 [$-$0.16, 0.31] & \textbf{0.41 [0.13, 0.72]} \\
\cellcolor{resLow}\quad Amharic & $-$0.12 [$-$0.33, 0.11] & \textbf{$-$0.30 [$-$0.60, $-$0.01]} & \textbf{1.12 [0.57, 1.61]} \\
\cellcolor{resLow}\quad Igbo & $-$0.07 [$-$0.30, 0.13] & \textbf{0.32 [0.06, 0.63]} & \textbf{0.62 [0.18, 1.07]} \\
\cellcolor{resLow}\quad Swahili & 0.07 [$-$0.22, 0.40] & $-$0.01 [$-$0.34, 0.29] & \textbf{0.81 [0.38, 1.21]} \\
\cellcolor{resLow}\quad Yoruba & 0.03 [$-$0.24, 0.30] & 0.28 [$-$0.11, 0.69] & \textbf{1.14 [0.60, 1.68]} \\
\addlinespace
\multicolumn{4}{l}{\textit{Models}} \\
\cellcolor{multiA}\quad Phi-4-reasoning & $-$0.05 [$-$0.20, 0.10] & $-$0.09 [$-$0.24, 0.06] & \textbf{$-$0.18 [$-$0.35, $-$0.01]} \\
\cellcolor{multiB}\quad DeepSeek-R1-Distill-Llama-8B & 0.08 [$-$0.14, 0.33] & \textbf{0.40 [0.09, 0.71]} & \textbf{2.14 [1.68, 2.72]} \\
\cellcolor{multiC}\quad Qwen3-8B & 0.10 [$-$0.03, 0.22] & 0.05 [$-$0.08, 0.19] & \textbf{$-$0.36 [$-$0.68, $-$0.03]} \\
\midrule\midrule
 & Native reasoning & Strategic code-switching & Synthetic code-switching \\
\midrule
\multicolumn{4}{l}{\textit{Languages}} \\
\cellcolor{resHigh}\quad Hindi & 0.17 [$-$0.06, 0.41] & 0.14 [$-$0.10, 0.38] & 0.12 [$-$0.11, 0.35] \\
\cellcolor{resHigh}\quad Indonesian & 0.10 [$-$0.13, 0.37] & 0.16 [$-$0.08, 0.41] & 0.16 [$-$0.07, 0.39] \\
\cellcolor{resHigh}\quad Malay & $-$0.03 [$-$0.26, 0.19] & $-$0.08 [$-$0.29, 0.16] & $-$0.13 [$-$0.35, 0.10] \\
\cellcolor{resLow}\quad Amharic & $-$0.20 [$-$0.45, 0.00] & $-$0.15 [$-$0.39, 0.10] & \textbf{$-$0.34 [$-$0.57, $-$0.09]} \\
\cellcolor{resLow}\quad Igbo & \textbf{0.31 [0.07, 0.56]} & 0.00 [$-$0.24, 0.21] & 0.04 [$-$0.20, 0.26] \\
\cellcolor{resLow}\quad Swahili & $-$0.03 [$-$0.32, 0.29] & $-$0.24 [$-$0.55, 0.04] & $-$0.11 [$-$0.43, 0.17] \\
\cellcolor{resLow}\quad Yoruba & -- & 0.07 [$-$0.18, 0.31] & -- \\
\addlinespace
\multicolumn{4}{l}{\textit{Models}} \\
\cellcolor{multiA}\quad Phi-4-reasoning & 0.01 [$-$0.13, 0.16] & $-$0.05 [$-$0.20, 0.09] & $-$0.06 [$-$0.20, 0.07] \\
\cellcolor{multiB}\quad DeepSeek-R1-Distill-Llama-8B & 0.08 [$-$0.19, 0.37] & 0.04 [$-$0.18, 0.27] & $-$0.07 [$-$0.31, 0.18] \\
\cellcolor{multiC}\quad Qwen3-8B & 0.06 [$-$0.06, 0.19] & $-$0.01 [$-$0.13, 0.11] & 0.00 [$-$0.12, 0.12] \\
\bottomrule
\end{tabular}
\caption{ACME by SFT intervention, language, and model in percentage points, with 95\% confidence intervals. Resource level groups:
\swatch{resHigh}~high/mid, \swatch{resLow}~low. Models, ordered from least to most baseline (non\textendash fine-tuned) languages officially supported (based on official model docs):
\swatch{multiA}~Phi, \swatch{multiB}~DeepSeek, \swatch{multiC}~Qwen. The native-language and synthetically code-switched reasoning SFT interventions were not applied to Yoruba (as explained in section \ref{sec:sft}).}\label{tab:mediation-subgroups}
\end{table}
Table~\ref{tab:mediation-subgroups} breaks down ACME by language and model. We see that SFT for open-domain MT into English has consistently positive ACME in the low-resource languages and for DeepSeek-R1-Distill-Llama-8B.

\end{document}